\documentclass[10pt,twocolumn]{next++}
\usepackage{array,colortbl,tabularx,float,pifont,bm,wrapfig,enumitem}

\definecolor{oursrow}{RGB}{217,217,255}
\definecolor{tablehead}{RGB}{248,248,248}
\definecolor{bThirtyTwoGrid}{RGB}{0,0,0}

\newlength{\cmpmethodwidth}
\newlength{\cmpnumwidth}

\newcommand{\method}{SPARC}

\newcommand{\appquestion}[1]{\par\addvspace{6pt}\noindent\textbf{\textit{#1}}\par\nopagebreak[4]\vspace{3pt}\noindent\ignorespaces}

\title{SPARC: Semantic Calibration with Latent Multimodal Knowledge for Robust Image-Text Retrieval}
\author[1,5]{Zhongkuan Mao}
\author[1]{Wenzhuo Zhao}
\author[2]{Haidong Chen}
\author[3]{Shangze Li}
\author[4]{Yidong Wang}
\author[5]{Jinhe Bi}
\author[6]{Yunpu Ma}
\author[5]{Junfeng Fang}
\author[5,\dagger]{Fei Shen}
\author[5]{Tat-Seng Chua}

\affiliation[1]{Sichuan University}
\affiliation[2]{North China University of Technology}
\affiliation[3]{Nanjing University of Science and Technology}
\affiliation[4]{Peking University}
\affiliation[5]{National University of Singapore}
\affiliation[6]{Ludwig-Maximilians-Universit\textrm{ä}t M\textrm{ü}nchen}
\abstract{Robust image–text retrieval aims to maintain accurate cross-modal matching when image or text queries are incomplete, supporting reliable information access in real-world applications.
However, existing methods typically focus on adapting to distribution shifts, without adequately addressing candidate discrimination for incomplete queries and ranking instability.
To this end, we propose semantic projection and alignment with relational constraints (\textbf{SPARC}), which elicits latent multimodal knowledge from frozen encoders to improve candidate discrimination and ranking stability.
Specifically, we first devise a semantic calibration module on a product of spheres that refines incomplete visual and textual features through tangent-space updates, maintaining the unit-sphere structure used for cosine similarity.
Second, we introduce a robust semantic guidance module that combines latent multimodal knowledge encoded in complete image--text features with cross-modal semantic anchors to promote semantic alignment of the calibrated features.
Finally, we introduce a matching relation constraint module that fuses original query features with their calibrated counterparts and applies a soft ranking constraint to penalize reductions in initially positive matching margins, thereby limiting ranking perturbations.
Extensive experiments on Flickr30K and COCO demonstrate that SPARC outperforms competing methods under various image occlusion and word deletion settings, while achieving consistent gains across different pretrained backbones.
}

\checkdata[Resources]{
  \href{https://github.com/MttJelly/SPARC}{\raisebox{-1ex}{\includegraphics[height=4.2ex]{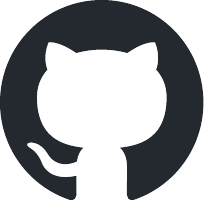}}~Code}
}

\begin{document}
\maketitle
\addtocontents{toc}{\protect\setcounter{tocdepth}{0}}

\begingroup
  \renewcommand{\thefootnote}{\fnsymbol{footnote}}
  \footnotetext[2]{
    Corresponding Author: Fei Shen
    (\href{mailto:shenfei29@nus.edu.sg}{shenfei29@nus.edu.sg}).
  }
\endgroup

\section{Introduction}

\begin{figure}[t]
    \centering
    \includegraphics[width=0.98\linewidth]{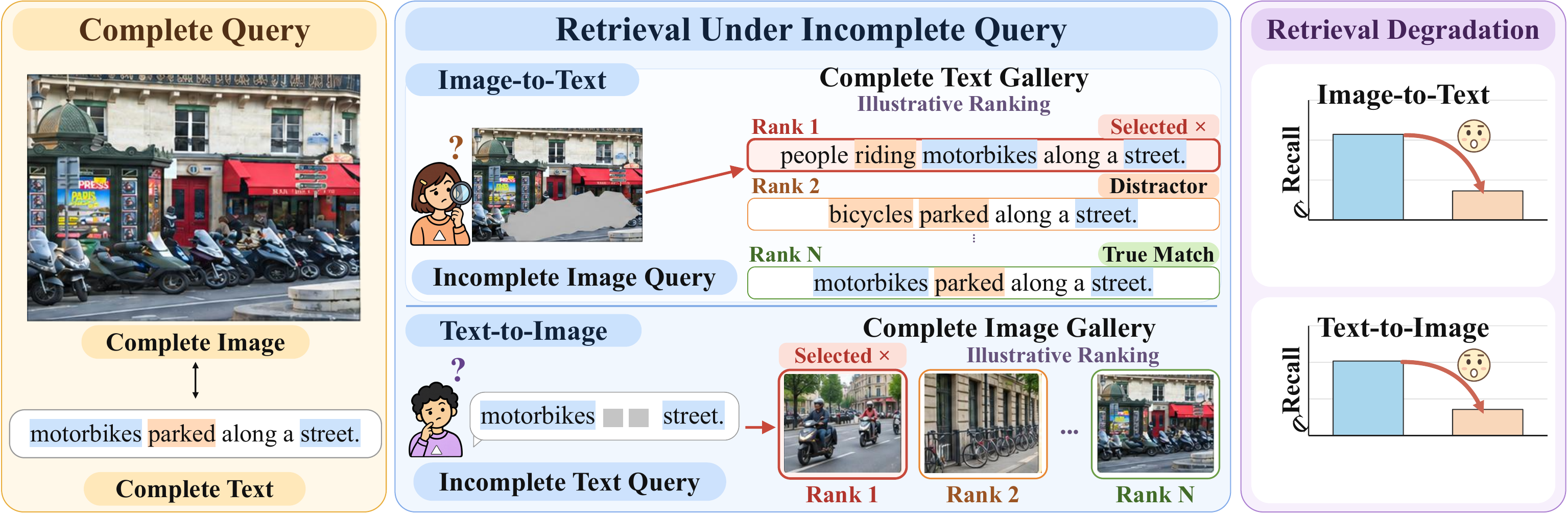}
    \vspace{-0.3cm}
    \caption{\textbf{Cross-Modal Retrieval under Incomplete Queries}: incomplete queries can rank distractors above true matches.}
    \label{fig:intro_accessibility}
    \vspace{-0.3cm}
\end{figure}

Robust image--text retrieval~\cite{kim2023rvse,li2025testtime,qiu2022robustness,chen2023benchmarking} aims to maintain accurate cross-modal matching with incomplete queries, supporting the reliable deployment of vision--language models (VLMs) in real-world applications.
In e-commerce search, users may submit partially occluded product photos or incomplete descriptions to retrieve matching products.
Models such as CLIP~\cite{radford2021clip}, ALIGN~\cite{jia2021align}, and SigLIP~\cite{zhai2023siglip} acquire rich multimodal knowledge through large-scale image--text pretraining, providing transferable semantic representations for this task.
However, as illustrated in Figure~\ref{fig:intro_accessibility}, image occlusion or word deletion can weaken fine-grained attribute and relational cues, hindering the use of this knowledge to distinguish correct candidates from semantically similar distractors and maintain stable retrieval rankings.

Most image--text retrieval studies~\cite{radford2021clip,lee2018stacked,faghri2017vse++,li2019visual,diao2021similarity} focus on conventional settings with relatively complete queries, achieving substantial progress on standard benchmarks through global feature alignment, local semantic interactions, and contrastive learning.
Recent work has extended these efforts to retrieval with incomplete queries, improving robustness through data augmentation and adaptation to distribution shifts.
For example, RVSE~\cite{kim2023rvse} uses semantic-preserving augmentation to learn features robust to image and text perturbations, while TCR~\cite{li2025testtime} addresses query shift through test-time adaptation, improving query feature uniformity and reducing the query--gallery modality gap.
However, retrieval gains alone do not reveal how models use latent multimodal knowledge to distinguish correct candidates from semantically similar distractors with incomplete queries.
Feature adjustment may also disturb useful matching relations, motivating joint consideration of candidate discrimination and ranking stability.

As the parable of the blind men and the elephant illustrates, local tactile information and prior experience can provide cues about an object's shape and category even without visual input.
Similarly, the remaining image regions or words in incomplete queries can be linked to object, attribute, and scene associations encoded in pretrained models.
Local cues thus convey information about visible content while also providing a basis for eliciting latent multimodal knowledge, allowing learned semantic associations to support matching decisions from limited evidence.
Therefore, retrieval performance with incomplete queries depends not only on how much feature information is retained but also on how effectively latent multimodal knowledge supports candidate discrimination and ranking stability.

For that, we propose semantic projection and alignment with relational constraints (SPARC) to improve candidate discrimination and ranking stability using latent multimodal knowledge from frozen pretrained encoders.
Specifically, we first devise a semantic calibration (SC) module on a product of spheres that refines visual and textual features from incomplete inputs, keeping the calibrated features on the unit spheres used for cosine matching.
Second, we present a robust semantic guidance (RSG) module that combines latent multimodal knowledge from complete image--text features and cross-modal semantic anchors with contrastive learning for semantic alignment and candidate discrimination.
Finally, we introduce a matching relation constraint (MRC) module that fuses original and calibrated query features and uses a soft ranking constraint to penalize reductions in initially positive matching margins.
Complete image--text references are used only for training supervision; at inference, only the incomplete query is calibrated for retrieval against a fixed, complete gallery.
Extensive experiments on Flickr30K~\cite{young2014flickr} and COCO~\cite{lin2014coco} demonstrate that SPARC outperforms competing methods under various image occlusion and word deletion settings, while achieving consistent gains across different pretrained backbones.
The main contributions are summarized as follows:

\begin{itemize}
    \item We propose SPARC for robust image--text retrieval, using latent multimodal knowledge from frozen encoders to guide semantic calibration of incomplete queries, while constraining matching relations with gallery candidates to improve candidate discrimination and ranking stability.

    \item We devise SC, RSG, and MRC modules, combining tangent-space updates on a product of spheres, training supervision from complete image--text features and cross-modal semantic anchors, and soft ranking penalties for reductions in initially positive matching margins.

    \item Across four pre-trained backbones, SPARC achieves higher retrieval performance than competing methods on Flickr30K and COCO under image occlusion and word deletion, while maintaining consistent gains across different incomplete types and severity levels.
\end{itemize}

\section{Related Work}

\noindent{\textbf{Vision--Language Models.}}
Vision--language foundation models~\cite{yao2022filip,Zhai_2022_CVPR,mu2022slip,singh2022flava,xu2024demystifying,wu2026himo} acquire transferable semantic representations through large-scale image--text pretraining.
For example, CLIP and ALIGN learn visual--textual associations through dual-encoder contrastive learning~\cite{radford2021clip,jia2021align}, while SigLIP adopts a pairwise sigmoid objective~\cite{zhai2023siglip}.
Building on these representations, CLIP-Adapter introduces lightweight residual feature adaptation~\cite{gao2021adapter}, while CLIP-Refine and AlignCLIP~\cite{yamaguchi2025refine,eslami2025alignclip} improve cross-modal alignment through post-pretraining and modified pretraining, respectively.
Prompt learning and feature adaptation~\cite{zhou2022learning,zhou2022conditional,zhang2022tip,khattak2023maple} further support efficient transfer to downstream tasks.
However, these methods primarily improve representation quality and cross-modal alignment, leaving the loss of fine-grained discrimination under image occlusion or word deletion insufficiently addressed.

\noindent{\textbf{Latent Knowledge.}}
Research on latent knowledge~\cite{li2026latent,shi2026culture,li2026forensic,dou2026beyond,guo2026hub,shi2026correct,kan2026v} investigates how semantic information encoded in pretrained models can be identified, elicited, and utilized across tasks and modalities.
For example, neuron-level analyses~\cite{schwettmann2023multimodal,huo2024mmneuron,goh2021multimodal,oikarinen2023clipdissect} identify concept-sensitive units in pretrained networks, including multimodal neurons in CLIP.
Sparse autoencoders (SAEs)~\cite{zaigrajew2025interpreting,bricken2023towards,gao2024sae} decompose dense activations into sparse features, providing another approach to interpreting encoded concepts.
For example, PatchSAE~\cite{lim2025patchsae} reveals interpretable visual concepts in CLIP and links prompt-adaptation gains to concepts learned during pretraining, while DNA~\cite{dou2026dna} identifies task-sensitive layers and forgery-discriminative units to uncover latent forgery knowledge.
However, these studies leave insufficiently explored how corrupted queries can elicit retained knowledge to distinguish semantically similar candidates while preserving useful matching relations.

\begin{figure*}[t]
    \centering
    \includegraphics[width=0.98\linewidth]{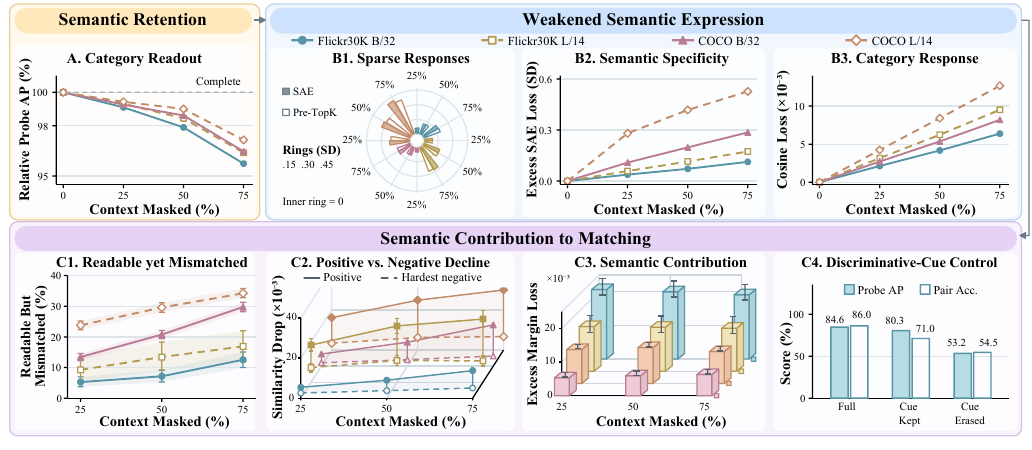}
    \vspace{-0.3cm}
    \caption{
    \textbf{Semantic Retention, Expression, and Matching under Incomplete Input.}
    \textbf{(A)} Local probe AP normalized to complete inputs.
    \textbf{(B1--B3)} SAE/pre-TopK attenuation, excess SAE loss over matched controls, and category cosine loss.
    \textbf{(C1)} Readable But Mismatched rates from global features in initially readable, correctly retrieved cohorts.
    \textbf{(C2)} Similarity drops for positives and hardest negatives.
    \textbf{(C3)} Excess match-margin loss for semantic versus matched random removal.
    \textbf{(C4)} Probe AP and pair accuracy for full, cue-kept, and cue-erased inputs.
    }
    \label{fig:partial_semantic_evidence}
    \vspace{-0.3cm}
\end{figure*}

\section{Methodology}
\label{sec:method}
\begin{figure*}[t]
    \centering
    \includegraphics[width=0.98\linewidth]{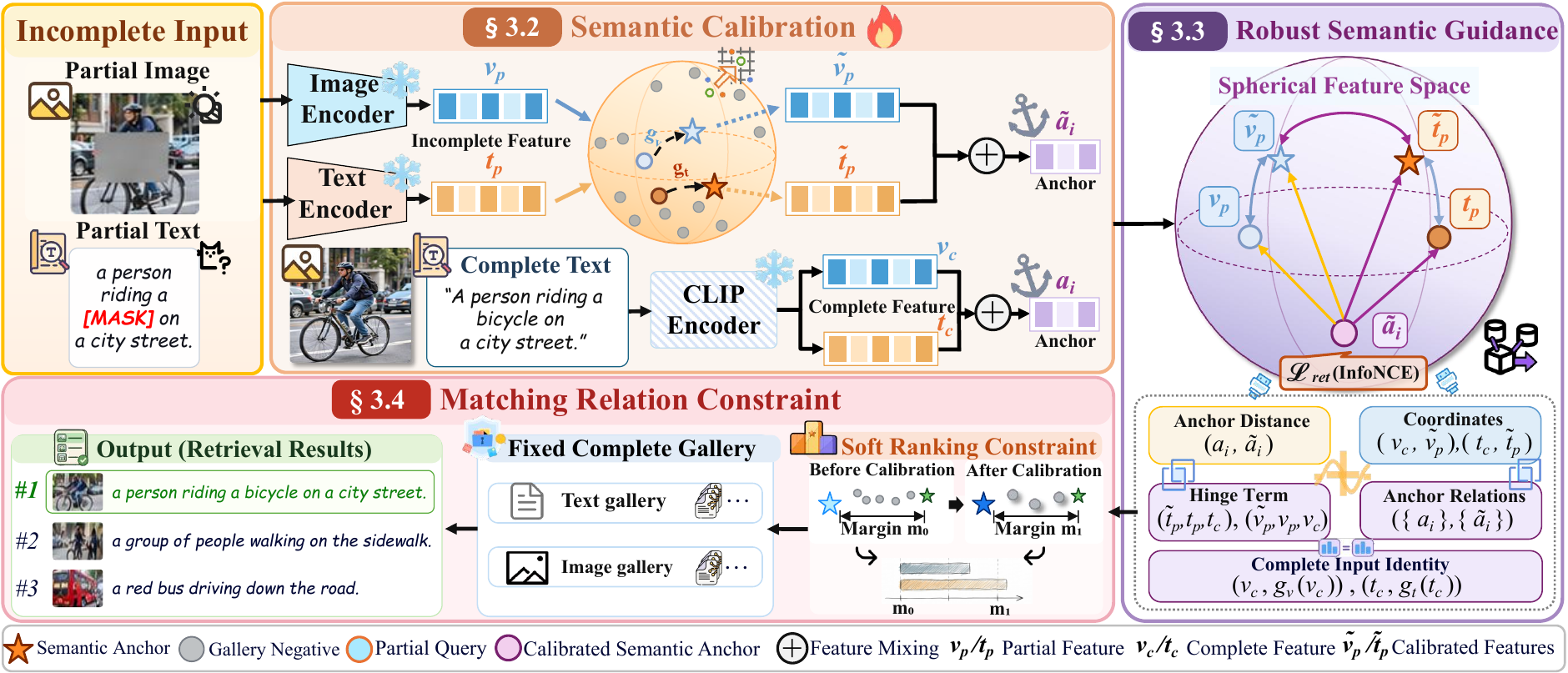}
    \vspace{-0.3cm}
    \caption{
    \textbf{Overview of SPARC.}
    Semantic calibration maps $g_v$ and $g_t$ refine frozen CLIP features
    and construct calibrated anchors $\tilde a_i$.
    Complete features and anchors $a_i$ provide fixed training references
    for robust semantic guidance.
    Feature mixing and soft ranking constraint constrain query--gallery
    matching relations.
    Only the observed query and its modality-specific map are required
    for calibration at inference.
    }
    \label{fig:framework}
    \vspace{-0.3cm}
\end{figure*}

SPARC elicits latent multimodal knowledge from frozen pretrained encoders to improve candidate discrimination and ranking stability for incomplete queries.
As shown in Figure~\ref{fig:framework}, it comprises three complementary modules.
Semantic Calibration (SC) refines query features through modality-specific tangent-space updates on a product of spheres.
Robust Semantic Guidance (RSG) combines complete image--text features and cross-modal semantic anchors with contrastive learning to supervise semantic alignment and candidate discrimination.
Matching Relation Constraint (MRC) combines feature mixing with a soft ranking constraint to limit disruption to useful matching relations.
The calibration maps are jointly optimized, with complete semantic references used only during training.

\subsection{Problem Setup and Motivation}
\label{sec:problem_formulation}

\noindent\textbf{Retrieval Setting.}
Let $(I_i,\{T_{ik}\}_{k=1}^{5})$ denote an image and its five associated captions.
Frozen pretrained encoders produce complete visual and textual features $v_{c,i}$ and $t_{c,ik}$, respectively.
During training, we sample one caption per image and generate incomplete image and text views, yielding features $v_{p,i}$ and $t_{p,i}$.
The corresponding complete caption feature is denoted by $t_{c,i}$, while all five complete caption features remain available for semantic anchor construction.
Subscripts $c$ and $p$ denote complete and incomplete inputs, respectively, and all encoder features are $\ell_2$-normalized.
We learn modality-specific calibration maps $g_v$ and $g_t$, producing $\tilde v_{p,i}=g_v(v_{p,i})$ and $\tilde t_{p,i}=g_t(t_{p,i})$.
Image-to-text retrieval matches an incomplete image query against complete captions, while text-to-image retrieval matches an incomplete caption query against complete images.
Gallery features are obtained from the original frozen encoders and remain fixed.
At inference, only the available query is calibrated and mixed with its original feature for retrieval.
Complete query references are used only during training.

\noindent\textbf{Motivating Observations.}
\label{sec:partial_knowledge}
We examine retained semantics on Flickr30K~\cite{young2014flickr} and COCO~\cite{lin2014coco} using frozen CLIP ViT-B/32 and ViT-L/14~\cite{radford2021clip}, progressively masking context while preserving annotated target regions.
Diagnostic probes and SAEs are fitted to frozen training-split features, with configurations selected on the validation split.
Readouts, SAE parameters, and semantic-direction selection rules remain fixed during test analysis.
Figure~\ref{fig:partial_semantic_evidence}(A) shows local category-probe AP retains 95.7--97.2\% of its complete-input value under 75\% context masking.
Panels (B1--B3) show weakened category responses and alignment, indicating that retained semantics can coexist with diminished expression.
Initially readable and correctly retrieved queries can retain readable global category information while losing correct top-1 matches (C1).
Correct-match similarities decline more than hardest-negative similarities, reducing matching margins (C2).
Semantic-component removal causes greater margin losses than matched random controls (C3), and cue-preservation and cue-erasure comparisons demonstrate the contribution of discriminative cues to matching (C4).
Repeating the global-readout analysis under 50\% center masking and word deletion with frozen CLIP ViT-B/32 yields readable-but-mismatched rates of 22.14--31.15\% among initially readable, correctly retrieved queries across datasets.
The appendix provides these extended results, feature layers, annotation sources, panel-specific sample sizes, and uncertainty estimation.
Thus, incomplete queries can retain useful semantic evidence with insufficient separation between correct candidates and similar distractors.

\subsection{Semantic Calibration}
\label{sec:calibration}

\noindent\textbf{Modality-Specific Maps.}
Normalized visual and textual features lie on unit hyperspheres, defining the joint space $\mathcal M=\mathbb S_v^{d-1}\times\mathbb S_t^{d-1}$, where $d$ is the shared feature dimension.
SC learns separate visual and textual calibration maps, $g_v$ and $g_t$, on their respective factors. Each map receives only its corresponding query feature, allowing the two retrieval directions to operate independently at inference.

\noindent\textbf{Tangent-Space Updates.}
For modality $m\in\{v,t\}$ and unit query feature $z$, a two-layer network $h_m$ predicts a residual.
We project this residual onto the tangent space at $z$, add it to the original feature, and normalize the result as
\begin{equation}
\label{eq:tangent_calibration}
\begin{aligned}
r_m^\perp(z)
&=h_m(z)-\langle h_m(z),z\rangle z,\\
g_m(z)
&=\operatorname n\!\left(z+r_m^\perp(z)\right),
\end{aligned}
\end{equation}
where $\operatorname{n}(x)=x/\|x\|_2$.
The tangent projection ensures that
$\langle z,r_m^\perp(z)\rangle=0$.
For unit-normalized $z$, the angular displacement satisfies $\angle(z,g_m(z))=\arctan\|r_m^\perp(z)\|_2$,
so the residual norm controls the angular calibration step. Normalization keeps the feature on the unit sphere for cosine matching. We initialize the output layer of each $h_m$ to zero, so that
$g_m(z)=z$ at initialization. Each query undergoes a single calibration update. Training optimizes the calibration networks while keeping the pretrained encoders frozen.

\subsection{Robust Semantic Guidance}
\label{sec:semantic_guidance}

RSG guides semantic calibration through reference consistency
and structural regularization, with
$\mathcal L_{\mathrm{RSG}}=\mathbb E_i[\varphi_i+\omega_i]$,
where $\mathbb E_i$ averages over the training batch.
$\varphi_i$ and $\omega_i$ are computed
for each training pair.
We additionally apply the symmetric InfoNCE loss
$\mathcal L_{\mathrm{ret}}$~\cite{radford2021clip}
to calibrated image--text pairs for cross-modal candidate discrimination.
The similarities are
$s_{ij}=\tilde v_{p,i}^{\mathsf T}\tilde t_{p,j}/\tau$,
where $\tau$ is the temperature.

\begin{table*}[!t]
\centering
\caption{
\textbf{Evaluation Using CLIP ViT-B/32.}
For \textbf{image-to-text retrieval (I$\to$T)}, CM and RM denote
center masking and random masking of image queries.
For \textbf{text-to-image retrieval (T$\to$I)}, WD denotes
word deletion from text queries.
Best results are \textbf{bold}.
}
\label{tab:comparison_b32}

\begingroup
\renewcommand{\textbf}[1]{{\bfseries #1}}
\fontsize{8}{9.2}\selectfont
\setlength{\tabcolsep}{3pt}
\setlength{\cmpmethodwidth}{0.23\linewidth}
\renewcommand{\arraystretch}{1.02}
\setlength{\extrarowheight}{0pt}
\setlength{\aboverulesep}{0.15ex}
\setlength{\belowrulesep}{0.15ex}
\setlength{\cmpnumwidth}{\dimexpr(\linewidth-\cmpmethodwidth-18\tabcolsep)/9\relax}
\begin{tabular}{@{}>{\raggedright\arraybackslash}p{\cmpmethodwidth}*{9}{>{\raggedleft\arraybackslash}p{\cmpnumwidth}}@{}}
\toprule
\multicolumn{1}{l}{\textbf{(a) R@1 (\%)}} & \multicolumn{3}{c}{\textbf{25\%}} & \multicolumn{3}{c}{\textbf{50\%}} & \multicolumn{3}{c}{\textbf{75\%}} \\
\cmidrule(lr){2-4}\cmidrule(lr){5-7}\cmidrule(lr){8-10}
Method & CM & RM & WD & CM & RM & WD & CM & RM & WD \\
\midrule
\rowcolor{tablehead}
\multicolumn{10}{l}{\textbf{Flickr30K~\cite{young2014flickr}}} \\
AlignCLIP~\cite{eslami2025alignclip} & 51.00 & 50.30 & 41.54 & 30.40 & 33.30 & 24.38 & 14.00 & 15.60 & 9.12 \\
HiMo-CLIP~\cite{wu2026himo} & 51.00 & 54.90 & 49.22 & 34.20 & 35.50 & 28.64 & 12.20 & 14.40 & 10.96 \\
CalibCLIP~\cite{kang2025calibclip} & 53.90 & 59.20 & 37.32 & 35.60 & 39.70 & 21.18 & 16.90 & 17.60 & 8.14 \\
Frozen CLIP~\cite{radford2021clip} & 54.10 & 59.20 & 41.24 & 36.70 & 39.80 & 23.90 & 17.50 & 17.50 & 9.26 \\
RVSE~\cite{kim2023rvse} & 55.10 & 59.10 & 47.76 & 30.30 & 32.50 & 28.80 & 10.90 & 11.20 & 11.44 \\
CLIP-Refine~\cite{yamaguchi2025refine} & 56.50 & 61.90 & 43.02 & 37.20 & 40.30 & 25.42 & 18.00 & 17.80 & 9.72 \\
CLIP-Adapter~\cite{gao2021adapter} & 56.70 & 60.00 & 40.32 & 36.70 & 40.10 & 23.46 & 18.00 & 18.10 & 9.20 \\
TCR~\cite{li2025testtime} & 62.20 & 63.70 & 44.58 & 43.00 & 44.10 & 26.98 & 22.70 & 21.60 & 10.14 \\
R-Erasing~\cite{zhong2020random} & 63.70 & 68.70 & 47.80 & 46.70 & 48.60 & 28.84 & 27.40 & 27.60 & 11.30 \\
Fine-tuned CLIP & 66.00 & 70.80 & 48.28 & 52.40 & 51.60 & 32.06 & 29.00 & 28.40 & 12.72 \\
\rowcolor{oursrow}
\textbf{SPARC (Ours)} & \textbf{70.90} & \textbf{71.90} & \textbf{51.78} & \textbf{52.50} & \textbf{53.80} & \textbf{32.72} & \textbf{30.20} & \textbf{29.80} & \textbf{12.94} \\
\addlinespace[1.5pt]
\rowcolor{tablehead}
\multicolumn{10}{l}{\textbf{COCO~\cite{lin2014coco}}} \\
AlignCLIP~\cite{eslami2025alignclip} & 17.50 & 19.36 & 14.50 & 9.76 & 11.00 & 6.96 & 3.72 & 4.00 & 1.95 \\
HiMo-CLIP~\cite{wu2026himo} & 26.96 & 29.62 & 25.92 & 14.56 & 15.80 & 12.37 & 2.98 & 3.72 & 3.37 \\
CalibCLIP~\cite{kang2025calibclip} & 27.26 & 30.84 & 17.09 & 15.62 & 16.94 & 8.22 & 4.04 & 4.94 & 2.41 \\
Frozen CLIP~\cite{radford2021clip} & 27.66 & 31.04 & 19.58 & 15.90 & 17.62 & 9.60 & 4.36 & 5.12 & 2.82 \\
CLIP-Adapter~\cite{gao2021adapter} & 29.04 & 31.40 & 15.47 & 17.80 & 19.30 & 7.65 & 6.78 & 7.42 & 2.38 \\
CLIP-Refine~\cite{yamaguchi2025refine} & 30.62 & 33.78 & 20.35 & 17.38 & 19.66 & 10.29 & 5.42 & 6.10 & 2.93 \\
RVSE~\cite{kim2023rvse} & 33.12 & 35.34 & 25.40 & 19.20 & 19.98 & 12.09 & 5.66 & 6.02 & 3.27 \\
TCR~\cite{li2025testtime} & 37.64 & 39.76 & 22.74 & 24.72 & 25.94 & 10.79 & 11.08 & 10.78 & 2.99 \\
R-Erasing~\cite{zhong2020random} & 40.94 & 41.82 & 25.18 & 27.88 & 29.70 & 11.99 & 14.68 & 14.52 & 3.32 \\
Fine-tuned CLIP & 41.38 & 43.58 & 25.55 & 29.92 & 29.68 & 13.11 & 14.00 & 14.02 & 3.27 \\
\rowcolor{oursrow}
\textbf{SPARC (Ours)} & \textbf{44.58} & \textbf{46.16} & \textbf{27.62} & \textbf{31.52} & \textbf{31.84} & \textbf{14.06} & \textbf{14.96} & \textbf{15.48} & \textbf{4.27} \\
\bottomrule
\end{tabular}
\par\vspace{4pt}
\setlength{\cmpnumwidth}{\dimexpr(\linewidth-\cmpmethodwidth-18\tabcolsep)/9\relax}
\begin{tabular}{@{}>{\raggedright\arraybackslash}p{\cmpmethodwidth}*{9}{>{\raggedleft\arraybackslash}p{\cmpnumwidth}}@{}}
\toprule
\multicolumn{1}{l}{\textbf{(b) RSum}} & \multicolumn{3}{c}{\textbf{25\%}} & \multicolumn{3}{c}{\textbf{50\%}} & \multicolumn{3}{c}{\textbf{75\%}} \\
\cmidrule(lr){2-4}\cmidrule(lr){5-7}\cmidrule(lr){8-10}
Method & CM & RM & WD & CM & RM & WD & CM & RM & WD \\
\midrule
\rowcolor{tablehead}
\multicolumn{10}{l}{\textbf{Flickr30K~\cite{young2014flickr}}} \\
AlignCLIP~\cite{eslami2025alignclip} & 199.60 & 208.20 & 188.38 & 146.80 & 150.60 & 131.26 & 79.60 & 84.70 & 63.06 \\
HiMo-CLIP~\cite{wu2026himo} & 206.80 & 219.60 & 206.82 & 151.90 & 161.80 & 144.82 & 76.30 & 82.40 & 69.92 \\
CalibCLIP~\cite{kang2025calibclip} & 215.10 & 226.80 & 173.64 & 163.70 & 174.60 & 115.56 & 97.80 & 98.70 & 53.94 \\
Frozen CLIP~\cite{radford2021clip} & 215.60 & 226.60 & 186.48 & 165.40 & 175.20 & 126.76 & 98.80 & 99.70 & 60.48 \\
RVSE~\cite{kim2023rvse} & 214.40 & 226.60 & 203.44 & 143.50 & 151.70 & 145.76 & 63.10 & 63.60 & 73.00 \\
CLIP-Refine~\cite{yamaguchi2025refine} & 220.40 & 232.40 & 192.84 & 167.30 & 177.00 & 132.98 & 99.50 & 101.40 & 63.98 \\
CLIP-Adapter~\cite{gao2021adapter} & 221.20 & 229.20 & 182.66 & 171.50 & 177.50 & 125.84 & 101.70 & 105.90 & 61.94 \\
TCR~\cite{li2025testtime} & 235.50 & 238.50 & 195.02 & 185.00 & 191.10 & 137.48 & 119.10 & 119.00 & 67.16 \\
R-Erasing~\cite{zhong2020random} & 238.40 & 248.10 & 205.00 & 194.00 & 201.60 & 146.82 & 135.70 & 135.40 & 73.36 \\
Fine-tuned CLIP & 244.40 & 252.50 & 207.02 & 211.50 & 205.80 & 156.60 & 137.00 & 133.60 & 82.06 \\
\rowcolor{oursrow}
\textbf{SPARC (Ours)} & \textbf{255.10} & \textbf{256.00} & \textbf{216.18} & \textbf{216.30} & \textbf{216.50} & \textbf{158.44} & \textbf{145.40} & \textbf{149.80} & \textbf{82.18} \\
\addlinespace[1.5pt]
\rowcolor{tablehead}
\multicolumn{10}{l}{\textbf{COCO~\cite{lin2014coco}}} \\
AlignCLIP~\cite{eslami2025alignclip} & 102.22 & 108.90 & 93.23 & 66.52 & 70.20 & 51.14 & 29.58 & 30.26 & 18.39 \\
HiMo-CLIP~\cite{wu2026himo} & 136.24 & 144.62 & 136.34 & 83.94 & 90.06 & 77.87 & 27.74 & 31.62 & 28.08 \\
CalibCLIP~\cite{kang2025calibclip} & 139.58 & 151.98 & 99.97 & 91.02 & 98.22 & 55.49 & 34.38 & 38.42 & 20.63 \\
Frozen CLIP~\cite{radford2021clip} & 141.16 & 152.82 & 111.41 & 92.02 & 100.14 & 62.47 & 35.46 & 39.70 & 23.33 \\
CLIP-Adapter~\cite{gao2021adapter} & 144.90 & 151.66 & 92.39 & 100.40 & 105.80 & 53.03 & 47.34 & 51.68 & 20.55 \\
CLIP-Refine~\cite{yamaguchi2025refine} & 151.30 & 161.76 & 118.30 & 99.44 & 107.28 & 67.11 & 40.44 & 44.36 & 24.59 \\
RVSE~\cite{kim2023rvse} & 159.62 & 166.68 & 135.12 & 106.78 & 108.52 & 77.80 & 40.36 & 41.36 & 27.67 \\
TCR~\cite{li2025testtime} & 173.16 & 179.26 & 123.34 & 128.26 & 132.30 & 70.00 & 70.92 & 71.50 & 25.58 \\
R-Erasing~\cite{zhong2020random} & 182.96 & 187.48 & 134.53 & 142.38 & 145.94 & 77.64 & 85.52 & 85.16 & 27.94 \\
Fine-tuned CLIP & 184.72 & 189.24 & 136.16 & 149.58 & 146.34 & 87.57 & 79.78 & 82.74 & 31.16 \\
\rowcolor{oursrow}
\textbf{SPARC (Ours)} & \textbf{194.20} & \textbf{196.90} & \textbf{143.62} & \textbf{152.96} & \textbf{153.96} & \textbf{88.68} & \textbf{88.70} & \textbf{91.80} & \textbf{34.84} \\
\bottomrule
\end{tabular}
\endgroup
\end{table*}

\noindent\textbf{Semantic References.}
RSG uses latent multimodal knowledge encoded in complete image--text features to supervise the calibration maps.
Complete features provide modality-specific references, while cross-modal semantic anchors integrate visual and textual information.
For each image, we aggregate its five complete caption features and combine their normalized mean with the complete visual feature as
\begin{equation}
\label{eq:anchor}
\bar t_{c,i}
=\operatorname n\!\left(\frac{1}{5}\sum_{k=1}^{5}t_{c,ik}\right),
\quad
a_i=\operatorname n(v_{c,i}+\bar t_{c,i}).
\end{equation}
The corresponding calibrated anchor is
$\tilde a_i=\operatorname n(\tilde v_{p,i}+\tilde t_{p,i})$.
The sampled complete caption feature $t_{c,i}$ serves as the textual coordinate reference, while $\bar t_{c,i}$ is used for anchor construction.
Complete references are detached from gradient computation and used only during training.
Gradients through $\tilde a_i$ update both calibration maps.

\noindent\textbf{Reference Consistency.}
For unit features, let $d(x,y)=1-x^\mathsf{T}y$.
We use $z_{p,v,i}=v_{p,i}$, $z_{p,t,i}=t_{p,i}$,
$z_{c,v,i}=v_{c,i}$, $z_{c,t,i}=t_{c,i}$, and
$\tilde z_{p,m,i}=g_m(z_{p,m,i})$ for $m\in\{v,t\}$.
The reference discrepancies before and after calibration are
$D_{m,i}=d(z_{p,m,i},z_{c,m,i})$ and
$\tilde D_{m,i}=d(\tilde z_{p,m,i},z_{c,m,i})$,
and the anchor discrepancy is $D_{a,i}=d(\tilde a_i,a_i)$.
We compute the reference-consistency quantity as
\begin{equation}
\label{eq:coordinate_loss}
\varphi_i
=D_{a,i}+\sum_m
\Bigl(\tfrac12\tilde D_{m,i}
+[\tilde D_{m,i}-\rho D_{m,i}]_+\Bigr).
\end{equation}
where $[x]_+=\max(x,0)$ and $\rho$ specifies the target fraction
of the original reference discrepancy.
The anchor and coordinate distances guide calibrated features toward
fixed complete references, while the hinge term encourages
improvement relative to the original incomplete features.

\noindent\textbf{Structure and Identity Regularization.}
Relationships among complete anchors supervise the semantic
organization of calibrated features.
For $j\ne i$, we define
\begin{equation}
\label{eq:relation_distribution}
p_{ij}
=\frac{\exp(a_i^\mathsf{T}a_j/\tau)}
{\sum_{\ell\ne i}\exp(a_i^\mathsf{T}a_\ell/\tau)}.
\end{equation}
The calibrated distribution $\tilde p_{ij}$ is computed by substituting
calibrated anchors for complete anchors.
Writing $p_i=(p_{ij})_{j\ne i}$ and
$\tilde p_i=(\tilde p_{ij})_{j\ne i}$, we compute
\begin{equation}
\label{eq:relation_loss}
\omega_i=\operatorname{KL}(p_i\|\tilde p_i)
+\sum_{m\in\{v,t\}}
\|g_m(z_{c,m,i})-z_{c,m,i}\|_2^2.
\end{equation}
The KL divergence encourages consistency between anchor neighborhoods.
The squared distances discourage unnecessary adjustments to
complete-input features.

\subsection{Matching Relation Constraint}
\label{sec:matching_constraint}

MRC combines feature mixing with a soft ranking constraint loss
$\mathcal L_{\mathrm{MRC}}$.
The mixed feature defines the query state used to evaluate this
loss and perform retrieval.

\noindent\textbf{Feature Mixing.}
To retain original query evidence while incorporating calibrated semantics, MRC combines the original and calibrated features.
For modality $m\in\{v,t\}$, let $z_{p,m}$ denote an incomplete query feature and $\tilde z_{p,m}=g_m(z_{p,m})$ its calibrated counterpart.
Their mixed feature is
\begin{equation}
\label{eq:mixing}
z_m^\beta
=\operatorname n\!\left(
(1-\beta)z_{p,m}+\beta\tilde z_{p,m}
\right),
\end{equation}
where $\beta$ controls the calibrated feature's contribution.
The mixed feature is used for the training-time ranking constraint and retrieval against the fixed complete gallery at inference.

\noindent\textbf{Soft Ranking Constraint.}
For a query feature $q$, consider a complete training-batch gallery $\{y_j\}$ and let $\mathcal P(q)$ denote its positive candidate indices.
We define the best-positive versus hardest-negative matching margin as
\begin{equation}
\label{eq:query_margin}
m(q)
=\max_{j\in\mathcal P(q)}q^\mathsf{T}y_j
-\max_{j\notin\mathcal P(q)}q^\mathsf{T}y_j.
\end{equation}
The galleries provide five complete captions as positives for each image query and one complete image as positive for each text query.
With gallery features and positive indices fixed, we compare the original query $q_m=z_{p,m}$ and its mixed counterpart $q_m^\beta=z_m^\beta$ using their margins $m_0=m(q_m)$ and $m_1=m(q_m^\beta)$.
A positive initial margin means a correct candidate outranks all negatives in the training-batch gallery.
We penalize margin reductions through

\begin{equation}
\label{eq:retention}
\mathcal L_{\mathrm{MRC}}
=\frac{1}{2}\sum_{m\in\{v,t\}}
\mathbb E_{q_m}\left[
\frac{\mathbf 1_{\{m_0>0\}}[m_0-m_1]_+}{\tau}
\right].
\end{equation}

The expectation includes all queries, with zero contribution when $m_0\le0$.
The RSG objective remains active for these queries.
The MRC loss discourages decreases in useful matching margins for the mixed features used at inference.

\begin{table*}[!t]
\centering
\caption{
\textbf{Complete Input and Corruption Robustness Results Using CLIP ViT-B/32.}
Robustness results average image or text corruption conditions for
I$\to$T or T$\to$I, respectively.
RSum sums R@1, R@5, and R@10.
Best results are \textbf{bold}.
}
\label{tab:complete_robustness}
\label{tab:complete_inputs}
\label{tab:robustness}

\begingroup
\renewcommand{\textbf}[1]{{\bfseries #1}}
\fontsize{8}{9.2}\selectfont
\setlength{\tabcolsep}{3pt}
\setlength{\cmpmethodwidth}{0.23\linewidth}
\renewcommand{\arraystretch}{1.02}
\setlength{\extrarowheight}{0pt}
\setlength{\aboverulesep}{0.15ex}
\setlength{\belowrulesep}{0.15ex}
\setlength{\cmpnumwidth}{\dimexpr(\linewidth-\cmpmethodwidth-16\tabcolsep)/8\relax}
\begin{tabular}{@{}>{\raggedright\arraybackslash}p{\cmpmethodwidth}*{8}{>{\raggedleft\arraybackslash}p{\cmpnumwidth}}@{}}
\toprule
\multicolumn{1}{l}{\textbf{(a) Complete inputs}} & \multicolumn{4}{c}{\textbf{Flickr30K~\cite{young2014flickr}}} & \multicolumn{4}{c}{\textbf{COCO~\cite{lin2014coco}}} \\
\cmidrule(lr){2-5}\cmidrule(lr){6-9}
& \multicolumn{2}{c}{R@1 (\%)} & \multicolumn{2}{c}{RSum} & \multicolumn{2}{c}{R@1 (\%)} & \multicolumn{2}{c}{RSum} \\
\cmidrule(lr){2-3}\cmidrule(lr){4-5}\cmidrule(lr){6-7}\cmidrule(lr){8-9}
Method & I$\to$T & T$\to$I & I$\to$T & T$\to$I & I$\to$T & T$\to$I & I$\to$T & T$\to$I \\
\midrule
CLIP-Adapter~\cite{gao2021adapter} & 74.40 & 56.56 & 265.40 & 226.92 & 46.40 & 24.13 & 196.12 & 127.38 \\
AlignCLIP~\cite{eslami2025alignclip} & 74.70 & 58.28 & 260.00 & 232.10 & 34.28 & 22.86 & 164.02 & 128.40 \\
Fine-tuned CLIP & 78.00 & 62.54 & 269.80 & 241.44 & 51.30 & 34.64 & 210.18 & 168.12 \\
CalibCLIP~\cite{kang2025calibclip} & 78.20 & 54.60 & 272.30 & 221.46 & 50.16 & 26.84 & 208.48 & 140.80 \\
HiMo-CLIP~\cite{wu2026himo} & 78.40 & 66.26 & 270.90 & 245.78 & 51.32 & 37.26 & 210.56 & 178.55 \\
CLIP-Refine~\cite{yamaguchi2025refine} & 78.60 & 61.02 & 267.00 & 237.22 & 53.94 & 32.60 & 217.76 & 160.65 \\
RVSE~\cite{kim2023rvse} & 78.80 & 64.08 & 270.90 & 243.54 & 51.72 & 36.72 & 213.06 & 174.73 \\
TCR~\cite{li2025testtime} & 79.10 & 61.70 & 272.10 & 238.20 & 49.60 & 34.26 & 207.50 & 164.22 \\
Frozen CLIP~\cite{radford2021clip} & 79.30 & 58.88 & 272.40 & 232.38 & 50.10 & 30.52 & 208.50 & 153.35 \\
R-Erasing~\cite{zhong2020random} & 79.40 & 63.72 & 271.00 & 243.22 & 51.28 & 36.81 & 212.34 & 174.50 \\
\rowcolor{oursrow}
\textbf{SPARC (Ours)} & \textbf{80.40} & \textbf{67.38} & \textbf{273.40} & \textbf{251.52} & \textbf{54.58} & \textbf{38.83} & \textbf{218.92} & \textbf{180.88} \\
\bottomrule
\end{tabular}
\par\vspace{4pt}
\setlength{\cmpnumwidth}{\dimexpr(\linewidth-\cmpmethodwidth-16\tabcolsep)/8\relax}
\begin{tabular}{@{}>{\raggedright\arraybackslash}p{\cmpmethodwidth}*{8}{>{\raggedleft\arraybackslash}p{\cmpnumwidth}}@{}}
\toprule
\multicolumn{1}{l}{\textbf{(b) Corruption robustness}} & \multicolumn{4}{c}{\textbf{Flickr30K-C~\cite{qiu2022robustness}}} & \multicolumn{4}{c}{\textbf{COCO-C~\cite{qiu2022robustness}}} \\
\cmidrule(lr){2-5}\cmidrule(lr){6-9}
& \multicolumn{2}{c}{R@1 (\%)} & \multicolumn{2}{c}{RSum} & \multicolumn{2}{c}{R@1 (\%)} & \multicolumn{2}{c}{RSum} \\
\cmidrule(lr){2-3}\cmidrule(lr){4-5}\cmidrule(lr){6-7}\cmidrule(lr){8-9}
Method & I$\to$T & T$\to$I & I$\to$T & T$\to$I & I$\to$T & T$\to$I & I$\to$T & T$\to$I \\
\midrule
CLIP-Adapter~\cite{gao2021adapter} & 56.04 & 39.50 & 221.69 & 172.20 & 31.32 & 15.81 & 149.91 & 90.34 \\
AlignCLIP~\cite{eslami2025alignclip} & 55.90 & 40.60 & 218.55 & 176.35 & 23.06 & 14.81 & 123.18 & 89.90 \\
Fine-tuned CLIP & 59.56 & 45.43 & 227.88 & 189.89 & 35.40 & 24.32 & 164.53 & 126.97 \\
CalibCLIP~\cite{kang2025calibclip} & 57.75 & 37.01 & 226.88 & 163.67 & 33.07 & 17.50 & 157.28 & 98.39 \\
HiMo-CLIP~\cite{wu2026himo} & 57.18 & 46.11 & 223.49 & 188.84 & 33.02 & 25.83 & 156.53 & 129.59 \\
CLIP-Refine~\cite{yamaguchi2025refine} & 59.73 & 42.70 & 230.73 & 180.83 & 35.44 & 21.90 & 165.54 & 115.91 \\
RVSE~\cite{kim2023rvse} & 67.97 & 47.21 & 246.72 & 193.52 & 42.15 & 25.84 & 187.88 & 132.07 \\
TCR~\cite{li2025testtime} & 63.13 & 43.54 & 238.87 & 182.66 & 39.92 & 23.15 & 178.62 & 119.52 \\
Frozen CLIP~\cite{radford2021clip} & 57.81 & 40.73 & 226.84 & 174.88 & 33.33 & 20.15 & 158.02 & 108.71 \\
R-Erasing~\cite{zhong2020random} & 60.79 & 46.63 & 228.66 & 192.07 & 35.17 & 25.48 & 163.98 & 130.10 \\
\rowcolor{oursrow}
\textbf{SPARC (Ours)} & \textbf{68.08} & \textbf{50.33} & \textbf{248.26} & \textbf{202.73} & \textbf{45.13} & \textbf{28.41} & \textbf{195.09} & \textbf{141.55} \\
\bottomrule
\end{tabular}
\endgroup
\end{table*}

\subsection{Training and Inference}

\noindent\textbf{Joint Optimization.}
We jointly optimize the calibration networks with
$\mathcal L_{\mathrm{final}}=\mathcal L_{\mathrm{RSG}}
+\mathcal L_{\mathrm{ret}}+\mathcal L_{\mathrm{MRC}}$.
Training updates parameters of $h_v$ and $h_t$,
while the pretrained encoders remain frozen. The retrieval loss encourages discrimination between calibrated incomplete image–text pairs. Reference consistency anchors the calibrated representations to fixed complete-input features, while the structural and identity terms regularize their organization and complete-input behavior. The ranking constraint loss is evaluated using mixed queries and a fixed complete training gallery, using the same query construction and frozen gallery encoding as inference.

\noindent\textbf{Query-Only Inference.}
Given an incomplete query of modality $m$, the frozen encoder produces a unit-normalized feature $z_{p,m}$.
We apply the corresponding calibration map $g_m$ and mix its output with the original feature using Eq.~\eqref{eq:mixing}.
Candidates in the fixed complete gallery are ranked by cosine similarity $(z_m^\beta)^\mathsf{T}y_j$, where $y_j$ denotes a complete feature from the opposite modality.
Image queries retrieve complete captions, while text queries retrieve complete images.
Complete query references, cross-modal semantic anchors, and training losses are required only during training.

\section{Experiments and Analysis}
\label{sec:experiments}

\noindent\textbf{Datasets.}
We conduct experiments on Flickr30K~\cite{young2014flickr} and COCO~\cite{lin2014coco}, using fixed training, validation, and test splits following the Karpathy evaluation protocol~\cite{karpathy2015}.
Image-to-text retrieval uses incomplete image queries against complete captions, while text-to-image retrieval uses incomplete caption queries against complete images.
Image queries undergo center or random masking, and text queries undergo random word deletion.
Missingness rates of 25\%, 50\%, and 75\% specify the proportion of image area masked or words deleted.
Gallery images and captions remain complete.
We additionally evaluate common corruption robustness on Flickr30K-C~\cite{qiu2022robustness} and COCO-C~\cite{qiu2022robustness}.

\noindent\textbf{Metrics.}
We evaluate retrieval using R@$K$, the percentage of queries with at least one valid match among the top-$K$ candidates.
Each image query has five positive captions, while each caption query has one positive image.
RSum sums R@1, R@5, and R@10 within the same retrieval direction and evaluation condition.
Each backbone is evaluated under 18 incomplete-query conditions
(2 datasets $\times$ 3 corruption types $\times$ 3 missingness rates),
with 72 conditions in total across the four backbones.

\noindent\textbf{Hyperparameters.}
For SPARC, the pretrained encoders remain frozen throughout training and evaluation.
We use $224\times224$ images for CLIP ViT-B/32, ViT-B/16, and ViT-L/14, and $256\times256$ images for SigLIP2 ViT-B/32.
The calibration networks have a hidden width of 512 and are trained using AdamW~\cite{loshchilov2018decoupled} for 20 epochs, with learning rate and weight decay both set to $10^{-4}$ and a batch size of 512.
Training uses 50\% center masking for images and balanced sampling across text word-deletion rates of 25\%, 50\%, and 75\%.
All training runs use an NVIDIA A100 GPU. More details of all experiments and analyses are in the appendix.

\subsection{Main Results}
We compare SPARC with R-Erasing~\cite{zhong2020random},
frozen and fine-tuned CLIP~\cite{radford2021clip},
CLIP-Adapter~\cite{gao2021adapter},
CLIP-Refine~\cite{yamaguchi2025refine},
AlignCLIP~\cite{eslami2025alignclip},
CalibCLIP~\cite{kang2025calibclip},
RVSE~\cite{kim2023rvse}, TCR~\cite{li2025testtime} and HiMo-CLIP~\cite{wu2026himo}.
All trainable methods use the same in-domain training splits,
with 50\% center masking and the word-deletion rates defined above. All experiments are repeated with three seeds, with per-seed results provided in the appendix. Fine-tuned CLIP updates all model parameters.

\begin{table*}[!tp]
\centering
\caption{
\textbf{Results with CLIP ViT-B/16, CLIP ViT-L/14, and SigLIP2 ViT-B/32.}
CM and RM denote center and random masking for image-to-text
(I$\to$T) retrieval.
WD denotes word deletion for text-to-image (T$\to$I) retrieval.
Best results are \textbf{bold}.
}
\label{tab:backbone_l14}

\begingroup
\renewcommand{\textbf}[1]{{\bfseries #1}}
\fontsize{8}{9.2}\selectfont
\setlength{\tabcolsep}{3pt}
\setlength{\cmpmethodwidth}{0.23\linewidth}
\renewcommand{\arraystretch}{1.02}
\setlength{\extrarowheight}{0pt}
\setlength{\aboverulesep}{0.15ex}
\setlength{\belowrulesep}{0.15ex}
\setlength{\cmpnumwidth}{\dimexpr(\linewidth-\cmpmethodwidth-18\tabcolsep)/9\relax}
\begin{tabular}{@{}>{\raggedright\arraybackslash}p{\cmpmethodwidth}*{9}{>{\raggedleft\arraybackslash}p{\cmpnumwidth}}@{}}
\toprule
\multicolumn{1}{l}{\textbf{(a) R@1 (\%)}} & \multicolumn{3}{c}{\textbf{25\%}} & \multicolumn{3}{c}{\textbf{50\%}} & \multicolumn{3}{c}{\textbf{75\%}} \\
\cmidrule(lr){2-4}\cmidrule(lr){5-7}\cmidrule(lr){8-10}
Method & CM & RM & WD & CM & RM & WD & CM & RM & WD \\
\midrule
\rowcolor{tablehead}
\multicolumn{10}{l}{\textbf{Flickr30K~\cite{young2014flickr}\enspace--\enspace CLIP ViT-B/16}} \\
Frozen CLIP & 59.30 & 65.50 & 44.70 & 42.80 & 44.10 & 26.66 & 24.60 & 25.70 & 10.02 \\
Fine-tuned CLIP & 75.40 & 76.80 & 53.20 & 60.70 & 60.60 & 34.60 & 39.00 & 40.00 & 13.12 \\
\rowcolor{oursrow}
\textbf{SPARC (Ours)} & \textbf{77.40} & \textbf{78.20} & \textbf{55.52} & \textbf{60.80} & \textbf{61.40} & \textbf{35.18} & \textbf{42.40} & \textbf{41.70} & \textbf{13.92} \\
\addlinespace[1.5pt]
\rowcolor{tablehead}
\multicolumn{10}{l}{\textbf{Flickr30K~\cite{young2014flickr}\enspace--\enspace CLIP ViT-L/14}} \\
Frozen CLIP & 64.90 & 67.10 & 47.00 & 46.30 & 49.70 & 28.36 & 33.20 & 32.40 & 10.84 \\
Fine-tuned CLIP & 80.80 & 81.00 & 54.44 & 63.00 & 65.90 & 35.60 & 42.00 & 44.80 & 14.30 \\
\rowcolor{oursrow}
\textbf{SPARC (Ours)} & \textbf{81.20} & \textbf{81.60} & \textbf{59.08} & \textbf{68.60} & \textbf{70.10} & \textbf{38.08} & \textbf{51.50} & \textbf{49.60} & \textbf{15.02} \\
\addlinespace[1.5pt]
\rowcolor{tablehead}
\multicolumn{10}{l}{\textbf{Flickr30K~\cite{young2014flickr}\enspace--\enspace SigLIP2 ViT-B/32}} \\
Frozen SigLIP2 & 56.90 & 65.80 & 55.08 & 30.40 & 32.70 & 32.82 & 9.80 & 11.60 & 12.36 \\
Fine-tuned SigLIP2 & 67.50 & 71.50 & 54.32 & 44.40 & 47.00 & 35.84 & 18.80 & 23.60 & 14.54 \\
\rowcolor{oursrow}
\textbf{SPARC (Ours)} & \textbf{68.70} & \textbf{73.30} & \textbf{59.82} & \textbf{46.40} & \textbf{47.90} & \textbf{37.06} & \textbf{20.80} & \textbf{24.20} & \textbf{14.64} \\
\addlinespace[1.5pt]
\rowcolor{tablehead}
\multicolumn{10}{l}{\textbf{COCO~\cite{lin2014coco}\enspace--\enspace CLIP ViT-B/16}} \\
Frozen CLIP & 28.40 & 33.66 & 21.89 & 17.80 & 19.68 & 10.49 & 7.50 & 7.20 & 3.00 \\
Fine-tuned CLIP & 47.82 & 49.10 & 27.61 & 36.08 & 36.48 & 15.48 & 20.34 & 20.42 & 3.72 \\
\rowcolor{oursrow}
\textbf{SPARC (Ours)} & \textbf{48.92} & \textbf{50.98} & \textbf{29.75} & \textbf{37.00} & \textbf{38.32} & \textbf{15.51} & \textbf{21.98} & \textbf{21.24} & \textbf{4.67} \\
\addlinespace[1.5pt]
\rowcolor{tablehead}
\multicolumn{10}{l}{\textbf{COCO~\cite{lin2014coco}\enspace--\enspace CLIP ViT-L/14}} \\
Frozen CLIP & 32.12 & 33.32 & 23.92 & 17.66 & 21.40 & 11.79 & 8.96 & 10.40 & 3.37 \\
Fine-tuned CLIP & 49.70 & 51.04 & 29.40 & 36.40 & 37.76 & 16.35 & 21.16 & 20.54 & 5.01 \\
\rowcolor{oursrow}
\textbf{SPARC (Ours)} & \textbf{53.80} & \textbf{55.26} & \textbf{32.61} & \textbf{44.32} & \textbf{44.54} & \textbf{17.50} & \textbf{29.78} & \textbf{28.16} & \textbf{5.31} \\
\addlinespace[1.5pt]
\rowcolor{tablehead}
\multicolumn{10}{l}{\textbf{COCO~\cite{lin2014coco}\enspace--\enspace SigLIP2 ViT-B/32}} \\
Frozen SigLIP2 & 39.22 & 42.22 & 33.13 & 16.52 & 21.48 & 16.17 & 3.88 & 5.40 & 4.52 \\
Fine-tuned SigLIP2 & 46.28 & 49.02 & 32.91 & 26.84 & 32.02 & 18.46 & 10.70 & 12.14 & 5.55 \\
\rowcolor{oursrow}
\textbf{SPARC (Ours)} & \textbf{48.28} & \textbf{52.06} & \textbf{36.75} & \textbf{28.90} & \textbf{32.52} & \textbf{19.47} & \textbf{11.78} & \textbf{12.54} & \textbf{5.67} \\
\bottomrule
\end{tabular}
\par\vspace{4pt}
\setlength{\cmpnumwidth}{\dimexpr(\linewidth-\cmpmethodwidth-18\tabcolsep)/9\relax}
\begin{tabular}{@{}>{\raggedright\arraybackslash}p{\cmpmethodwidth}*{9}{>{\raggedleft\arraybackslash}p{\cmpnumwidth}}@{}}
\toprule
\multicolumn{1}{l}{\textbf{(b) RSum}} & \multicolumn{3}{c}{\textbf{25\%}} & \multicolumn{3}{c}{\textbf{50\%}} & \multicolumn{3}{c}{\textbf{75\%}} \\
\cmidrule(lr){2-4}\cmidrule(lr){5-7}\cmidrule(lr){8-10}
Method & CM & RM & WD & CM & RM & WD & CM & RM & WD \\
\midrule
\rowcolor{tablehead}
\multicolumn{10}{l}{\textbf{Flickr30K~\cite{young2014flickr}\enspace--\enspace CLIP ViT-B/16}} \\
Frozen CLIP & 229.40 & 243.80 & 194.72 & 188.80 & 192.80 & 135.44 & 127.00 & 130.70 & 64.28 \\
Fine-tuned CLIP & 263.60 & 265.40 & 218.24 & 234.60 & 234.00 & 164.46 & 172.50 & 173.50 & 83.82 \\
\rowcolor{oursrow}
\textbf{SPARC (Ours)} & \textbf{267.40} & \textbf{267.10} & \textbf{224.16} & \textbf{235.40} & \textbf{236.30} & \textbf{165.52} & \textbf{181.90} & \textbf{180.80} & \textbf{85.06} \\
\addlinespace[1.5pt]
\rowcolor{tablehead}
\multicolumn{10}{l}{\textbf{Flickr30K~\cite{young2014flickr}\enspace--\enspace CLIP ViT-L/14}} \\
Frozen CLIP & 246.20 & 250.30 & 199.20 & 200.50 & 208.80 & 138.70 & 154.50 & 156.30 & 66.72 \\
Fine-tuned CLIP & 271.00 & 274.50 & 221.14 & 236.70 & 242.40 & 165.72 & 183.20 & 187.80 & 85.84 \\
\rowcolor{oursrow}
\textbf{SPARC (Ours)} & \textbf{274.10} & \textbf{275.70} & \textbf{231.20} & \textbf{249.60} & \textbf{252.40} & \textbf{173.68} & \textbf{207.60} & \textbf{208.90} & \textbf{89.36} \\
\addlinespace[1.5pt]
\rowcolor{tablehead}
\multicolumn{10}{l}{\textbf{Flickr30K~\cite{young2014flickr}\enspace--\enspace SigLIP2 ViT-B/32}} \\
Frozen SigLIP2 & 223.30 & 242.40 & 220.46 & 144.70 & 158.00 & 155.62 & 64.60 & 76.10 & 73.72 \\
Fine-tuned SigLIP2 & 244.70 & 253.60 & 220.26 & 190.00 & 193.50 & 166.56 & 112.60 & 119.80 & 86.76 \\
\rowcolor{oursrow}
\textbf{SPARC (Ours)} & \textbf{250.30} & \textbf{260.30} & \textbf{231.92} & \textbf{195.00} & \textbf{200.00} & \textbf{171.56} & \textbf{118.90} & \textbf{125.80} & \textbf{87.40} \\
\addlinespace[1.5pt]
\rowcolor{tablehead}
\multicolumn{10}{l}{\textbf{COCO~\cite{lin2014coco}\enspace--\enspace CLIP ViT-B/16}} \\
Frozen CLIP & 141.82 & 158.88 & 119.29 & 101.52 & 109.48 & 67.64 & 55.24 & 54.40 & 24.64 \\
Fine-tuned CLIP & 200.62 & 205.12 & 143.72 & 169.26 & 167.26 & 93.54 & 109.30 & 109.86 & 36.29 \\
\rowcolor{oursrow}
\textbf{SPARC (Ours)} & \textbf{204.86} & \textbf{210.28} & \textbf{151.34} & \textbf{172.24} & \textbf{174.98} & \textbf{94.57} & \textbf{119.68} & \textbf{117.68} & \textbf{36.58} \\
\addlinespace[1.5pt]
\rowcolor{tablehead}
\multicolumn{10}{l}{\textbf{COCO~\cite{lin2014coco}\enspace--\enspace CLIP ViT-L/14}} \\
Frozen CLIP & 154.36 & 157.54 & 125.74 & 102.08 & 115.08 & 71.80 & 62.44 & 68.18 & 25.70 \\
Fine-tuned CLIP & 206.04 & 208.72 & 148.55 & 168.48 & 171.92 & 96.62 & 115.10 & 112.90 & 38.21 \\
\rowcolor{oursrow}
\textbf{SPARC (Ours)} & \textbf{217.54} & \textbf{221.70} & \textbf{159.80} & \textbf{189.46} & \textbf{192.00} & \textbf{101.50} & \textbf{146.34} & \textbf{142.30} & \textbf{39.56} \\
\addlinespace[1.5pt]
\rowcolor{tablehead}
\multicolumn{10}{l}{\textbf{COCO~\cite{lin2014coco}\enspace--\enspace SigLIP2 ViT-B/32}} \\
Frozen SigLIP2 & 172.48 & 182.02 & 158.72 & 91.88 & 108.80 & 92.74 & 30.08 & 36.78 & 32.85 \\
Fine-tuned SigLIP2 & 195.60 & 200.84 & 161.23 & 138.80 & 148.38 & 105.61 & 67.98 & 72.14 & 41.59 \\
\rowcolor{oursrow}
\textbf{SPARC (Ours)} & \textbf{203.38} & \textbf{211.26} & \textbf{172.44} & \textbf{143.22} & \textbf{154.54} & \textbf{109.64} & \textbf{75.16} & \textbf{77.46} & \textbf{42.03} \\
\bottomrule
\end{tabular}
\endgroup
\end{table*}

\begin{table}[t]
\centering
\caption{
\textbf{Module Ablation of SPARC.}
mR@1 (\%) and mRSum average nine incomplete-query conditions with equal weights. FD denotes feature distillation.
Best results within each backbone are \textbf{bold}.
}
\vspace{-0.3cm}
\label{tab:ablation}
\begingroup
\definecolor{tablehead}{RGB}{248,248,248}
\definecolor{modelrow}{RGB}{245,245,245}
\definecolor{oursrow}{RGB}{217,217,255}
\definecolor{inactive}{RGB}{170,170,170}
\newcommand{\ablationcheck}{\ding{51}}
\newcommand{\ablationcross}{\textcolor{inactive}{\ding{55}}}
\renewcommand{\textbf}[1]{{\bfseries #1}}
\scriptsize
\setlength{\tabcolsep}{1.15pt}
\renewcommand{\arraystretch}{1.05}
\setlength{\extrarowheight}{0.5pt}
\setlength{\aboverulesep}{0pt}
\setlength{\belowrulesep}{0pt}
\setlength{\arrayrulewidth}{0.3pt}
\newlength{\ablsettingwidth}
\newlength{\ablscwidth}
\newlength{\ablmodulewidth}
\newlength{\ablmetricwidth}
\setlength{\ablsettingwidth}{0.18\columnwidth}
\setlength{\ablscwidth}{0.055\columnwidth}
\setlength{\ablmodulewidth}{0.075\columnwidth}
\setlength{\ablmetricwidth}{%
  \dimexpr(\columnwidth-\ablsettingwidth-\ablscwidth
  -2\ablmodulewidth-14\tabcolsep-3\arrayrulewidth)/4\relax}
\begin{tabular}{@{}
>{\raggedright\arraybackslash}p{\ablsettingwidth}|
>{\centering\arraybackslash}p{\ablscwidth}
*{2}{>{\centering\arraybackslash}p{\ablmodulewidth}}|
*{2}{>{\raggedleft\arraybackslash}p{\ablmetricwidth}}|
*{2}{>{\raggedleft\arraybackslash}p{\ablmetricwidth}}@{}
}
\toprule
\rowcolor{tablehead}
& \multicolumn{3}{c|}{\textbf{Modules}}
& \multicolumn{2}{c|}{\textbf{Flickr30K}}
& \multicolumn{2}{c}{\textbf{COCO}} \\
\cmidrule[\arrayrulewidth]{2-8}
\rowcolor{tablehead}
Setting & SC & RSG & MRC
& mR@1 & mRSum & mR@1 & mRSum \\
\midrule

\rowcolor{modelrow}
\multicolumn{8}{l}{\textcolor{black}{\textbf{CLIP ViT-B/32}}} \\
Baseline
& \ablationcross & \ablationcross & \ablationcross
& 33.24 & 150.56 & 14.86 & 84.28 \\
FD
& \ablationcross & \ablationcross & \ablationcross
& 42.03 & 179.03 & 21.18 & 120.07 \\
Joint FD
& \ablationcheck & \ablationcross & \ablationcheck
& 44.74 & 187.12 & 25.07 & 126.34 \\
A1
& \ablationcheck & \ablationcross & \ablationcross
& 40.24 & 172.40 & 20.91 & 108.77 \\
A2
& \ablationcheck & \ablationcheck & \ablationcross
& 44.33 & 186.24 & 25.10 & 125.87 \\
A3
& \ablationcheck & \ablationcross & \ablationcheck
& 44.30 & 185.89 & 25.16 & 126.22 \\
\rowcolor{oursrow}
\textbf{Ours}
& \ablationcheck & \ablationcheck & \ablationcheck
& \textbf{45.17} & \textbf{188.43}
& \textbf{25.61} & \textbf{127.30} \\
\midrule

\rowcolor{modelrow}
\multicolumn{8}{l}{\textcolor{black}{\textbf{CLIP ViT-L/14}}} \\
Baseline
& \ablationcross & \ablationcross & \ablationcross
& 42.20 & 180.14 & 18.10 & 98.10 \\
FD
& \ablationcross & \ablationcross & \ablationcross
& 52.78 & 210.40 & 28.10 & 147.18 \\
Joint FD
& \ablationcheck & \ablationcross & \ablationcheck
& 56.80 & 216.77 & 34.02 & 154.98 \\
A1
& \ablationcheck & \ablationcross & \ablationcross
& 51.04 & 203.16 & 26.81 & 129.92 \\
A2
& \ablationcheck & \ablationcheck & \ablationcross
& 55.92 & 214.93 & 33.60 & 153.49 \\
A3
& \ablationcheck & \ablationcross & \ablationcheck
& 56.49 & 216.98 & 33.84 & 155.04 \\
\rowcolor{oursrow}
\textbf{Ours}
& \ablationcheck & \ablationcheck & \ablationcheck
& \textbf{57.20} & \textbf{218.06}
& \textbf{34.59} & \textbf{156.69} \\
\bottomrule
\end{tabular}
\endgroup
\vspace{-0.2cm}
\end{table}

\noindent\textbf{Main Comparison.}
SPARC leads in R@1 and RSum across 18 partial-observation conditions on both datasets (Table~\ref{tab:comparison_b32}). Equal-weight mean gains over frozen CLIP are 11.34\% and 40.45\%. Under 25\% center masking on Flickr30K, image-to-text R@1 reaches 70.90\%, exceeding frozen CLIP by 16.80\% and the strongest comparator, fine-tuned CLIP, by 4.90\%. Under 25\% word deletion on COCO, text-to-image R@1 reaches 27.62\%, exceeding frozen CLIP by 8.04\% and the strongest comparator, HiMo-CLIP, by 1.70\%. With frozen encoders and gallery features, these gains are consistent with more effective activation and use of pretrained latent multimodal knowledge through query-side calibration.

\noindent\textbf{Complete Queries.}
We directly evaluate each method's checkpoint from incomplete-query training on complete queries (Table~\ref{tab:complete_robustness}(a)). SPARC achieves the highest R@1 and RSum on both datasets. Image-to-text and text-to-image R@1 gains over frozen CLIP are 1.10\% and 8.50\% on Flickr30K, and 4.48\% and 8.31\% on COCO. These results indicate that calibration learned from incomplete-query training also benefits complete-query retrieval.

\noindent\textbf{Common Corruptions.}
Table~\ref{tab:complete_robustness}(b) evaluates broader image and text corruptions on Flickr30K-C and COCO-C. SPARC achieves the highest R@1 and RSum in both retrieval directions and benchmarks, outperforming robustness-oriented baselines such as RVSE and R-Erasing. On COCO-C, I$\to$T and T$\to$I R@1 reach 45.13\% and 28.41\%, exceeding the strongest comparator, RVSE, by 2.98\% and 2.57\%. The corresponding RSum gains are 7.21\% and 9.48\%. These results support effectiveness of SPARC's query-side calibration under a broader range of visual and textual perturbations.

\noindent\textbf{Backbone Generality.}
Table~\ref{tab:backbone_l14} evaluates SPARC with CLIP ViT-B/16, CLIP ViT-L/14, and SigLIP2 ViT-B/32. SPARC outperforms the corresponding frozen and fine-tuned baselines in R@1 and RSum across all 18 conditions for each backbone. The average R@1/RSum gains over frozen models are 13.44/43.67, 15.74/48.26, and 8.38/31.39 points. Compared with fine-tuned models, the gains are 1.36/4.49, 4.27/13.22, and 1.63/6.13 points. Calibration modules are trained independently for each dataset and backbone. These results support SPARC's applicability across different patch sizes, model scales, and pretrained vision--language models.

\begin{figure}[t]
    \centering
    \includegraphics[width=0.94\linewidth]
    {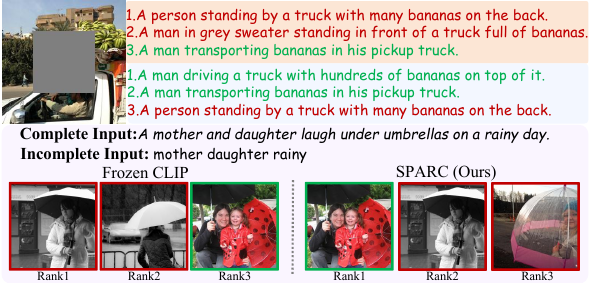}
    \vspace{-0.3cm}
    \caption{
    \textbf{Qualitative Analysis.}
    Orange and blue backgrounds denote Frozen CLIP and ours, respectively.
    Green and red text or image borders indicate correct matches and distractors.
    }
    \label{fig:qualitative_retrieval}
    \vspace{-0.2cm}
\end{figure}

\noindent\textbf{Qualitative Results.}
Figure~\ref{fig:qualitative_retrieval} presents COCO examples using CLIP ViT-B/32. For image-to-text retrieval, frozen CLIP prioritizes captions describing a man standing near a banana-loaded truck, whereas SPARC ranks the correct description of a man driving the truck first. For text-to-image retrieval, the incomplete query leads frozen CLIP to prioritize other umbrella scenes, while SPARC retrieves the paired image of a mother and daughter together. Both correct matches move from rank 3 to rank 1, illustrating improved discrimination among semantically similar candidates.

\subsection{Ablation Studies}
\label{sec:ablation}
Table~\ref{tab:ablation} compares module combinations on both datasets using CLIP ViT-B/32 and ViT-L/14. A1 uses SC, A2 adds RSG, and A3 adds MRC to SC. Full SPARC achieves the highest mean R@1 and RSum across all settings.

\noindent\textbf{Feature Distillation.}
Joint FD distills to RSG's frozen references
$v_{c,i}$, $t_{c,i}$, and $a_i$, with five complete captions used to construct $a_i$. It retains SC, feature mixing, $\mathcal L_{\mathrm{ret}}$, and $\mathcal L_{\mathrm{MRC}}$ under the same training budget. Across all settings, SPARC outperforms both baselines, exceeding Joint FD by 0.40\%--0.57\% in mR@1. Comparison with Joint FD supports the benefit of RSG when complete references, calibration structure, matching constraints, and training budget are held fixed.

\noindent\textbf{Calibration Structure.}
A1 trains SC solely with $\mathcal{L}_{\mathrm{ret}}$,
improving mean R@1 and mean RSum over frozen CLIP
across all settings. SC improves both mean R@1 and mean RSum on all settings.
On COCO with CLIP ViT-L/14, mean R@1 increases from 18.10\% to 26.81\%, while mean RSum rises from 98.10 to 129.92.
These results support the effectiveness of SC in calibrating incomplete-query representations.

\noindent\textbf{Semantic Guidance.}
Adding RSG to SC (A2 vs.\ A1) improves metrics across all settings.
On Flickr30K with CLIP ViT-B/32, mean R@1 increases from 40.24\% to 44.33\%, and mean RSum rises from 172.40 to 186.24.
With MRC included, full SPARC also outperforms A3 across all settings.
These comparisons support the contribution of semantic guidance under both module configurations. The appendix separately evaluates reference consistency,
anchor-relation KL, and complete-input identity regularization,
with SC, MRC, and training settings fixed.

\begin{figure}[t]
    \centering
    \includegraphics[width=0.94\linewidth]{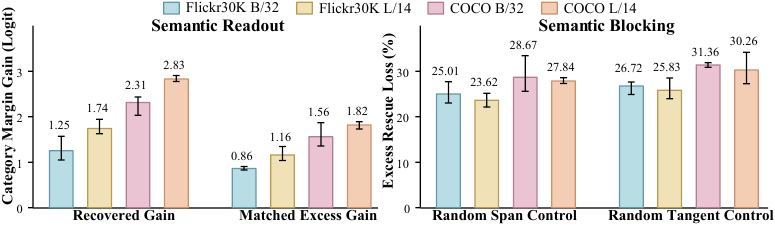}
    \vspace{-0.3cm}
    \caption{\textbf{Semantic Recovery and Blocking.} Left: category margin gains and matched gains (logits). Right: excess top-1 loss versus random span and tangent controls.}
    \label{fig:semantic_recovery}
\end{figure}

\begin{figure}[t]
    \centering
    \includegraphics[width=0.94\linewidth]{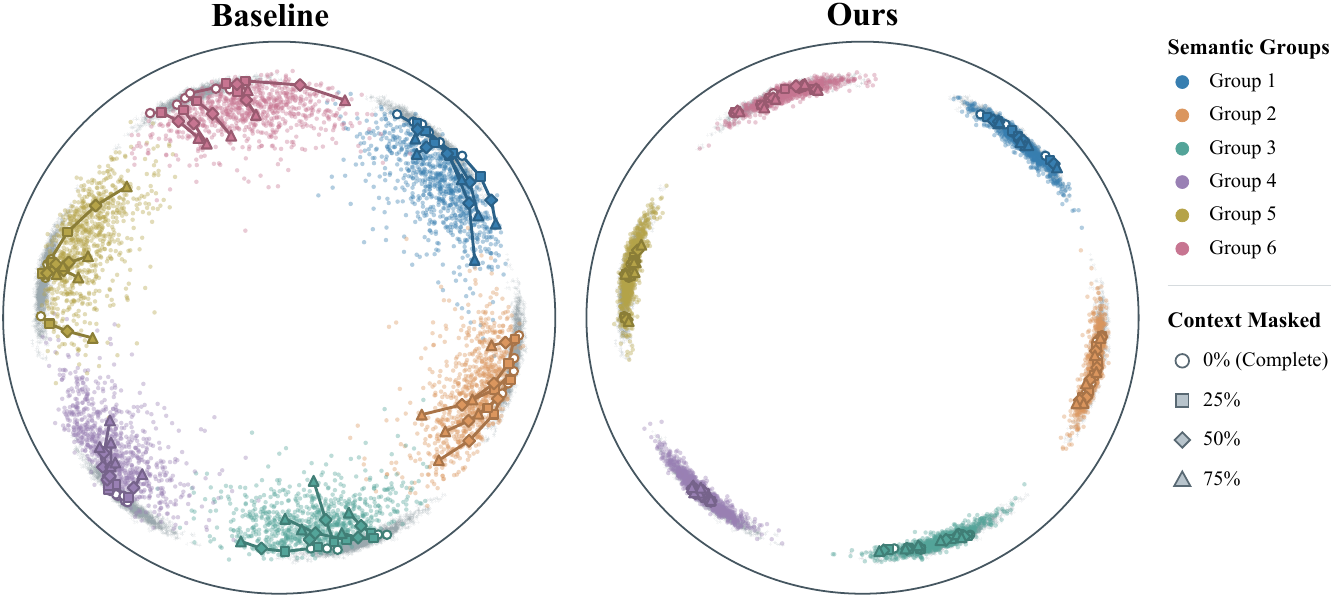}
    \vspace{-0.3cm}
    \caption{
    \textbf{Embedding Analysis on COCO Using CLIP ViT-B/32.}
    Baseline features (left) and SPARC features (right).
    }
    \vspace{-0.3cm}
    \label{fig:embedding_analysis}
\end{figure}

\noindent\textbf{Feature Mixing and Ranking Constraint.}
Adding MRC to SC (A3 vs.\ A1) improves retrieval across all four settings.
On COCO with CLIP ViT-L/14, mean R@1 increases from 26.81\% to 33.84\%, while mean RSum rises from 129.92 to 155.04.
MRC also benefits retrieval when RSG is present, as full SPARC consistently outperforms A2.
These module-level comparisons support the joint contribution of feature mixing and soft ranking constraint within MRC.

\subsection{Further Analysis}

\noindent\textbf{Semantic Recovery.}
Figure~\ref{fig:semantic_recovery} (left) compares initially incorrect frozen-CLIP queries corrected by SPARC with those that remain incorrect.
Within each corruption condition and modality, we match the groups on frozen-feature retrieval and category margins.
Fixed semantic readouts show category-margin gains of 1.25--2.83 logits for recovered queries, exceeding matched-group gains by 0.86--1.82 logits.
These comparisons associate retrieval recovery with stronger expression of retained semantics.

\noindent\textbf{Controlled Blocking.}
For a fixed cohort of recovered queries, we remove the calibration-update component in the semantic subspace defined by the readout directions.
Random-span and random-tangent controls match perturbation magnitudes, with random spans also matching subspace dimensionality.
All queries use the same fixed complete gallery.
Figure~\ref{fig:semantic_recovery} (right) shows positive excess losses of recovered matches across all four settings.
On COCO with ViT-B/32, these reach 28.67\% and 31.36\% relative to random-span and random-tangent controls, respectively.
These interventions support a functional contribution of retained semantic components to retrieval recovery.

\begin{figure}[t]
    \centering
    \includegraphics[width=0.92\linewidth]{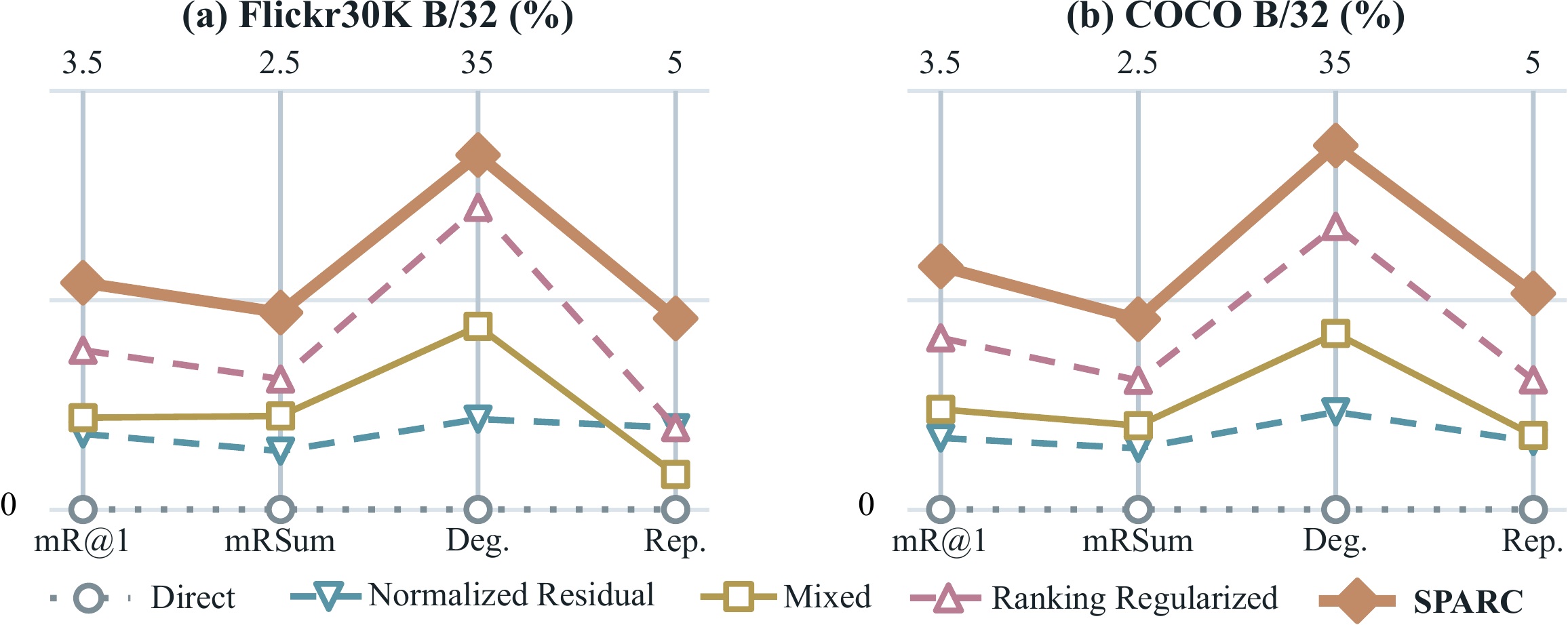}
    \vspace{-0.3cm}
    \caption{
    \textbf{Calibration Geometry and Match Preservation.}
    Relative improvements (\%) over Direct from equal-weight averages across nine conditions. Deg. and Rep. denote correct-to-incorrect rate reductions and incorrect-to-correct rate gains on frozen-query cohorts. Axes are normalized by their annotated limits.
    }
    \label{fig:calibration_preservation}
    \vspace{-0.3cm}
\end{figure}

\noindent\textbf{Neighborhood Preservation.}
Figure~\ref{fig:embedding_analysis} examines neighborhood preservation.
Each complete-input image feature's 15 nearest image neighbors define its reference neighborhood, excluding the query itself.
We measure mean neighborhood overlap using cosine similarity in normalized feature space before projection.
Under 75\% context masking, retention increases from 10.2\% with frozen CLIP to 56.7\% with SPARC.
The visualization shows more compact groups and shorter trajectories after calibration.
With encoders and gallery features frozen, these results support improved preservation of local semantic relations through query-side calibration.

\noindent\textbf{Calibration Geometry.}
Figure~\ref{fig:calibration_preservation} compares retrieval gains and top-1 match preservation on both datasets using CLIP ViT-B/32. Normalized Residual replaces Eq. (1) with $(g_m(z)=n(z+h_m(z)))$, keeping all other settings identical to SPARC. Mixed retains feature mixing and disables ranking constraint loss. SPARC outperforms Normalized Residual, with lower degradation and higher repair rates on both datasets, supporting the empirical benefit of tangent-space calibration. More results and details are in the appendix.

\section{Conclusion}
We propose SPARC, a query-side calibration that framework uses complete image-text representations as training-time semantic references. Modality-specific tangent-space updates refine incomplete-query features, while cross-modal anchors supervise semantic alignment. Feature mixing and a ranking-retention objective regularize the mixed query representations used for retrieval. The encoders and gallery features remain frozen, and inference requires only the observed query. On two datasets with four backbones, SPARC improves RSum in all 72 evaluated incomplete-query conditions, with an average gain of 40.94\% over the frozen models. Average R@1 gains over the fine-tuned models range from 1.36\% to 4.27\% across backbones.

\FloatBarrier
\bibliographystyle{assets/plainnat}
\bibliography{main}

\clearpage
\onecolumn
\beginappendix
\renewcommand{\thetable}{S\arabic{table}}
\renewcommand{\thefigure}{S\arabic{figure}}
\renewcommand{\theequation}{S\arabic{equation}}
\renewcommand{\theHtable}{S.\arabic{table}}
\renewcommand{\theHfigure}{S.\arabic{figure}}
\renewcommand{\theHequation}{S.\arabic{equation}}
\setcounter{table}{0}
\setcounter{figure}{0}
\setcounter{equation}{0}
\raggedbottom
\arrayrulecolor{black}
\setcounter{topnumber}{3}
\setcounter{bottomnumber}{2}
\setcounter{totalnumber}{5}
\renewcommand{\topfraction}{0.95}
\renewcommand{\bottomfraction}{0.90}
\renewcommand{\textfraction}{0.05}
\renewcommand{\floatpagefraction}{0.75}
\setlength{\floatsep}{6pt plus 1pt minus 1pt}
\setlength{\textfloatsep}{7pt plus 1pt minus 1pt}
\setlength{\intextsep}{6pt plus 1pt minus 1pt}
\setlength{\abovecaptionskip}{5pt}
\setlength{\belowcaptionskip}{2pt}
\setlength{\abovedisplayskip}{6pt}
\setlength{\belowdisplayskip}{6pt}
\setlength{\abovedisplayshortskip}{3pt}
\setlength{\belowdisplayshortskip}{3pt}
\renewcommand{\paragraph}[1]{\par\addvspace{4pt}\noindent\textbf{#1}\hspace{0.5em}\ignorespaces}
\addtocontents{toc}{\protect\setcounter{tocdepth}{2}}
\begingroup
\setcounter{tocdepth}{2}
\setlength{\parskip}{0pt}
\fontsize{9.5}{11.2}\selectfont
\tableofcontents
\endgroup
\clearpage
This supplement provides implementation details (Section~\ref{sec:app_implementation}), multi-seed retrieval summaries (Section~\ref{sec:app_complete_results}), RSG component ablations (Section~\ref{sec:app_rsg_ablation}), the controlled calibration-geometry comparison behind Fig.~7 (Section~\ref{sec:app_geometry}), additional analyses of retained semantic knowledge (Section~\ref{sec:app_semantics}), a neuron-level theoretical account of latent-knowledge elicitation (Section~\ref{app:neuron-theory}), selected qualitative retrieval examples (Section~\ref{sec:app_qualitative}), and further discussion (Section~\ref{sec:app_discussion}).

\section{Implementation Details}
\label{sec:app_implementation}

\paragraph{Data Splits and Galleries.}
Each model is trained and evaluated within its dataset. We use Flickr30K~\cite{young2014flickr} and COCO~\cite{lin2014coco} under the image--caption retrieval protocol of Karpathy and Fei-Fei~\cite{karpathy2015}. Flickr30K has 29,000 training, 1,014 validation, and 1,000 test images. COCO has 113,287 training, 5,000 validation, and 5,000 test images. Each image has five paired captions, with disjoint image groups across splits. Flickr30K test galleries contain 5,000 captions for image-to-text retrieval and 1,000 images for text-to-image retrieval; COCO uses 25,000 captions and 5,000 images. COCO evaluation uses the full test gallery, without 1,000-image folds. All five paired captions are positives for an image query; a caption has one paired image.

\paragraph{Partial-Observation Operators.}
Images are masked after resizing/cropping to the evaluation resolution and before normalization. For an input of size $W\times H$ and nominal missingness $r\in\{0.25,0.50,0.75\}$, a rectangle of size $\operatorname{round}(W\sqrt r)\times\operatorname{round}(H\sqrt r)$ is filled with RGB $(127,127,127)$. \emph{Center} centers the rectangle, and \emph{Random} places it at a uniformly sampled valid position (block masking, not pixel dropout).
\emph{Word deletion} uniformly retains $\max(1,\operatorname{round}((1-r)L))$ of $L$ whitespace-delimited words without replacement, in their original order. Seed 42 and the image or image--caption identifier fix the corruption stream. All methods receive identical partial queries.

\paragraph{Frozen Backbones and Calibration Maps.}
CLIP ViT-B/32, ViT-B/16, and ViT-L/14~\cite{radford2021clip,dosovitskiy2020vit} use $224\times224$ inputs, with embedding dimensions of 512, 512, and 768, respectively. SigLIP2 ViT-B/32~\cite{tschannen2025siglip2} uses $256\times256$ inputs, retaining the selected checkpoint's native resolution to match its pretrained image encoder and original preprocessing. It produces 768-dimensional embeddings and uses its native tokenizer with a maximum length of 64 tokens. The image and text encoders remain frozen, and each backbone has its own complete gallery. Separate visual and textual maps use cached normalized features and $d\!\to\!512\!\to\!d$ linear layers with GELU~\cite{hendrycks2016gaussian}. Zero-initializing the last layer initially preserves the input.
For a unit feature $q$, the map produces a residual $r_\theta(q)$ and a tangent displacement
\begin{equation}
 t_\theta(q)=(I-qq^\top)r_\theta(q),\qquad
 u_\theta(q)=\frac{q+t_\theta(q)}{\|q+t_\theta(q)\|_2}.
\end{equation}
The deployed feature is
\begin{equation}
 F_\beta(q)=\frac{(1-\beta)q+\beta u_\theta(q)}
 {\|(1-\beta)q+\beta u_\theta(q)\|_2},\quad\beta=0.6.
 \label{eq:app_map}
\end{equation}
The two maps act on separate sphere factors, coupled by training. At inference, the observed modality's map produces $F_{0.6}(q)$, which is ranked against the complete, frozen gallery.

\paragraph{Trainable Parameters.}
Each calibration map is a $d\!\to\!512\!\to\!d$ MLP, so the two maps contain 1.05M trainable parameters for CLIP ViT-B/32 ($d=512$) and 1.58M for CLIP ViT-L/14 ($d=768$), that is, 0.69\% and 0.37\% of the 151.28M and 427.62M parameters updated by full fine-tuning. Inference applies only the observed modality's map (0.53M and 0.79M parameters, respectively). Because the encoders remain frozen, gallery features are encoded once and reused without re-encoding, whereas full fine-tuning changes the encoders and requires re-encoding the gallery.

\paragraph{Training Objective and Schedule.}
In the notation of the main paper, $r_\theta=h_m$, $t_\theta=r_m^{\perp}$ and $u_\theta=g_m$ (Eq.~(1)), and $F_\beta(q)=\mathbf z_m^{\beta}$ (Eq.~(6)).
\method{} minimizes the objective of the main paper, $\mathcal{L}_{\mathrm{final}}=\mathcal{L}_{\mathrm{ret}}+\mathcal{L}_{\mathrm{RSG}}+\mathcal{L}_{\mathrm{MRC}}$, with no additional terms.
$\mathcal{L}_{\mathrm{ret}}$ is the symmetric InfoNCE~\cite{radford2021clip} over the similarities $s_{ij}$ of calibrated partial-image and partial-text features $u_\theta(q)$, that is, the full update without mixing.
$\mathcal{L}_{\mathrm{RSG}}=\mathbb{E}_i[\varphi_i+\omega_i]$ groups the semantic-guidance terms into the two quantities of the main paper.
The reference-consistency quantity $\varphi_i$ (Eq.~(3)) contains the anchor distance $D_{a,i}$, the coordinate distances $\tfrac12\sum_m\tilde D_{m,i}$ on the two sphere factors, and the relative-improvement hinges $\sum_m[\tilde D_{m,i}-\rho D_{m,i}]_+$ with $\rho=0.75$.
The structure and identity quantity $\omega_i$ (Eq.~(5)) contains the anchor-relation divergence $\mathrm{KL}(\mathbf p_i\Vert\tilde{\mathbf p}_i)$ over the distributions of Eq.~(4) and the complete-input identity penalty $\sum_m\|g_m(\mathbf z_{c,m,i})-\mathbf z_{c,m,i}\|_2^2$, weighted by $\lambda_{\mathrm{id}}=0.1$.
The terms of $\mathcal{L}_{\mathrm{RSG}}$ follow Section~3.3 of the main paper and use detached complete-reference features and anchors as supervision targets.
$\mathcal{L}_{\mathrm{MRC}}$ (Eq.~(8), Section~3.4 of the main paper) uses the mixed feature $F_{0.6}$ and complete frozen minibatch galleries. Let $m_0$ and $m_\theta$ (denoted $m_1$ in Eq.~(8)) be the observed query's best-positive minus strongest-negative margins before and after calibration. Its per-query penalty is
\begin{equation}
 \ell_{\mathrm{keep}}=\tau^{-1}\mathbf{1}[m_0>0]\,[m_0-m_\theta]_+.
 \label{eq:app_keep}
\end{equation}
Penalties average over all queries, including zeros, and then both directions. Image-query positives include all five paired minibatch captions.
Training uses 50\% center-masked images and word-deleted captions balanced across 25\%, 50\%, and 75\% missingness. Each epoch shuffles images and samples one of five captions per image; text-severity counts differ by at most one. AdamW~\cite{loshchilov2018decoupled} uses learning rate $10^{-4}$, weight decay $10^{-4}$, batch size 512 image groups, and 20 epochs. The temperature $\tau=0.07$ is shared by $s_{ij}$, the anchor-relation distributions of Eq.~(4), and Eq.~(8).
Every SPARC configuration is trained with three training seeds; evaluation corruptions remain fixed by seed 42 so that all runs receive identical partial queries.

\paragraph{Checkpoint Selection and Reporting.}
The epoch-selection score averages validation R@1 for 50\% center masking, 50\% word deletion, and an auxiliary joint query. The joint query normalizes the sum of the two calibrated partial features and retrieves fixed complete-pair references $a_i=\operatorname{normalize}(c_i+\operatorname{normalize}(\frac15\sum_j c_{ij}^{\mathrm{text}}))$. This auxiliary task is used for validation only; deployment observes one modality. With $\beta=0.6$ fixed, the earliest epoch maximizing this score supplies one checkpoint per dataset, backbone, and training seed for all test operators and rates. Test results do not select checkpoints. The supplementary comparison reports mR@1 and mRSum, each averaged with equal weights over the nine incomplete-query conditions within a dataset and backbone. These conditions comprise center masking and random masking for image-to-text retrieval and word deletion for text-to-image retrieval, each at 25\%, 50\%, and 75\% missingness. RSum (0--300) adds R@1, R@5, and R@10 within one direction and condition; mRSum averages these nine sums. Three-seed results are reported as mean $\pm$ sample SD and diagnostic analyses as aggregate point estimates; we do not infer statistical significance from either.

\paragraph{Fairness of Baseline Comparisons.}
All trainable baselines, including CLIP-Adapter~\cite{gao2021adapter} and CLIP-Refine~\cite{yamaguchi2025refine}, use the same in-domain complete image--text pairs and partial-view training conditions as SPARC: 50\% center masking for images and balanced word-deletion rates of 25\%, 50\%, and 75\%. Retaining their original protocols refers to preserving their method-specific objectives and optimization settings while adapting them to these shared partial inputs.
The module-ablation variants of Table~4 (A1--A3 and Joint FD) and the geometry variants of Section~\ref{sec:app_geometry} additionally provide matched-architecture controls: they use the same input features, calibration networks, parameter count, initialization, optimizer settings, batch size, 20-epoch training budget, training seeds, and validation-based checkpoint-selection criterion as SPARC. At evaluation, they receive the identical seed-42 partial queries and the same fixed complete gallery. Their training objectives, update parameterization, and mixing weight are the experimental variables, as specified for each comparison.

\subsection{Flickr30K-C and COCO-C Evaluation Protocol}
\label{sec:app_corruption_protocol}

\paragraph{Benchmark Names and Condition Inventory.}
The original benchmark calls its retrieval datasets Flickr30K-IP/TP and COCO-IP/TP, where IP and TP denote image and text perturbations~\cite{qiu2022robustness}. Flickr30K-C and COCO-C in the main paper are shorthand for the corresponding corruption evaluations. This gives 85 image conditions and 60 text conditions (Table~\ref{tab:app_corruption_inventory}); sentence-level text perturbations have a single setting each.

\begin{table}[!htbp]
\centering
\caption{\textbf{Common-Corruption Condition Inventory.} Counts follow the original multimodal robustness benchmark~\cite{qiu2022robustness}. The 16 text perturbation types have unequal numbers of severity settings.}
\label{tab:app_corruption_inventory}
\small
\setlength{\tabcolsep}{5pt}
\renewcommand{\arraystretch}{1.10}
\begin{tabularx}{\linewidth}{@{}>{\raggedright\arraybackslash}Xrrr@{}}
\toprule
Perturbation group & Types & Settings per type & Conditions\\
\midrule
Image perturbations & 17 & 5 & 85\\
Text: character level & 6 & 5 & 30\\
Text: word level & 5 & 5 & 25\\
Text: sentence level & 5 & 1 & 5\\
\midrule
Text total & 16 & Mixed & 60\\
\bottomrule
\end{tabularx}
\end{table}

Image families cover noise, blur, weather, digital distortion, and stylization. Text perturbations cover character errors, word-level edits, and the five sentence-level settings formal, casual, passive, active, and back translation. The original definitions and released perturbations are linked from the \href{https://mmrobustness.github.io/}{benchmark project page}~\cite{qiu2022robustness}.

\paragraph{Query-Only Evaluation with Complete Galleries.}
We apply this corruption inventory to the query-only retrieval setting of the main paper. Image-to-text evaluation uses image-perturbed queries against the fixed complete caption gallery; text-to-image evaluation uses text-perturbed queries against the fixed complete image gallery. Original image--caption IDs define the positives. Each method retains its validation-selected checkpoint and the complete gallery of its backbone. The clean-input results are reported separately and are excluded from corruption averages.

\paragraph{Condition-Weighted Reporting.}
For each dataset and retrieval direction, we first compute R@K within every corruption condition, using that condition's valid query records and the same complete gallery. We then average the 85 image conditions for I$\to$T or the 60 text conditions for T$\to$I with equal condition weights. Writing $\mathcal C_{\mathrm{img}}$ and $\mathcal C_{\mathrm{txt}}$ for these inventories,
\begin{equation}
\overline{\mathrm{R@K}}_{\mathrm{I\to T}}=
\frac{1}{85}\sum_{c\in\mathcal C_{\mathrm{img}}}\mathrm{R@K}(c),\qquad
\overline{\mathrm{R@K}}_{\mathrm{T\to I}}=
\frac{1}{60}\sum_{c\in\mathcal C_{\mathrm{txt}}}\mathrm{R@K}(c).
\label{eq:app_corruption_average}
\end{equation}
The directional mean RSum is $\overline{\mathrm{RSum}}=\sum_{K\in\{1,5,10\}}\overline{\mathrm{R@K}}$. Conditions, not perturbation types, receive equal weight. These conditions are distinct from the nine incomplete-query conditions used for mR@1 and mRSum.

\FloatBarrier
\section{Multi-Seed Retrieval Results}\label{sec:app_complete_results}

Table~\ref{tab:app_multiseed} summarizes retrieval performance on Flickr30K and COCO with four frozen pretrained backbones. For each of the three SPARC training seeds, we first average the nine condition-specific scores to obtain mR@1 and mRSum, and then summarize the three per-seed averages as mean $\pm$ sample standard deviation. The SPARC means equal the averages of the nine condition-specific SPARC results reported in Tables~1 and~3 of the main paper, and the Base and FT entries equal the corresponding averages for the frozen and fine-tuned models. Base is the frozen backbone without training. Table~\ref{tab:app_b32_per_seed} lists the per-seed SPARC results with CLIP ViT-B/32 in all nine incomplete-query conditions.

\begin{table}[!htbp]
\centering
\caption{\textbf{Multi-Seed Retrieval Summary across Datasets and Backbones.}
Base denotes the original frozen backbone and FT denotes its fine-tuned counterpart.
mR@1 and mRSum average the nine incomplete-query conditions with equal weights.
SPARC entries additionally report the sample standard deviation across three training seeds.
Best mean results within each dataset, backbone, and metric are bold.}
\label{tab:app_multiseed}
\begingroup
\fontsize{9.5}{11.3}\selectfont
\setlength{\tabcolsep}{4pt}
\renewcommand{\arraystretch}{1.15}
\begin{tabular*}{\linewidth}{@{\extracolsep{\fill}}llrr>{\columncolor{oursrow}}rrr>{\columncolor{oursrow}[\tabcolsep][0pt]}r@{}}
\toprule
\multicolumn{2}{c}{} & \multicolumn{3}{c}{mR@1 (\%)} & \multicolumn{3}{c}{mRSum}\\
\cmidrule(lr){3-5}\cmidrule(lr){6-8}
Dataset & Backbone & Base & FT & \textbf{SPARC} & Base & FT & \textbf{SPARC}\\
\midrule
\multirow{4}{*}{\rotatebox[origin=c]{90}{\shortstack{Flickr30K\\\cite{young2014flickr}}}}
& CLIP ViT-B/32~\cite{radford2021clip} & 33.24 & 43.47 & $\mathbf{45.17\,\pm\,0.14}$
& 150.56 & 181.16 & $\mathbf{188.43\,\pm\,0.31}$\\
& CLIP ViT-B/16~\cite{radford2021clip} & 38.15 & 50.38 & $\mathbf{51.84\,\pm\,0.02}$
& 167.44 & 201.12 & $\mathbf{204.85\,\pm\,0.08}$\\
& CLIP ViT-L/14~\cite{radford2021clip} & 42.20 & 53.54 & $\mathbf{57.20\,\pm\,0.32}$
& 180.14 & 207.59 & $\mathbf{218.06\,\pm\,0.83}$\\
& SigLIP2 ViT-B/32~\cite{tschannen2025siglip2} & 34.16 & 41.94 & $\mathbf{43.65\,\pm\,0.22}$
& 150.99 & 176.42 & $\mathbf{182.35\,\pm\,0.58}$\\
\midrule
\multirow{4}{*}{\rotatebox[origin=c]{90}{\shortstack{COCO\\\cite{lin2014coco}}}}
& CLIP ViT-B/32~\cite{radford2021clip} & 14.86 & 23.83 & $\mathbf{25.61\,\pm\,0.09}$
& 84.28 & 120.81 & $\mathbf{127.30\,\pm\,0.03}$\\
& CLIP ViT-B/16~\cite{radford2021clip} & 16.62 & 28.56 & $\mathbf{29.82\,\pm\,0.44}$
& 92.55 & 137.22 & $\mathbf{142.47\,\pm\,0.84}$\\
& CLIP ViT-L/14~\cite{radford2021clip} & 18.10 & 29.71 & $\mathbf{34.59\,\pm\,0.38}$
& 98.10 & 140.73 & $\mathbf{156.69\,\pm\,0.88}$\\
& SigLIP2 ViT-B/32~\cite{tschannen2025siglip2} & 20.28 & 25.99 & $\mathbf{27.55\,\pm\,0.22}$
& 100.71 & 125.80 & $\mathbf{132.13\,\pm\,0.61}$\\
\bottomrule
\end{tabular*}
\endgroup
\end{table}

\begin{table}[!htbp]
\centering
\caption{\textbf{Per-Seed SPARC Results with CLIP ViT-B/32.}
CM and RM denote center and random masking for I$\to$T retrieval, and WD denotes word deletion for T$\to$I retrieval.
For each condition, the mean over the three seeds equals the corresponding SPARC entry in Table~1 of the main paper.}
\label{tab:app_b32_per_seed}
\begingroup
\definecolor{bThirtyTwoHeader}{HTML}{F4F4F4}
\definecolor{bThirtyTwoGrid}{HTML}{000000}
\setlength{\tabcolsep}{3pt}
\setlength{\arrayrulewidth}{0.3pt}
\setlength{\aboverulesep}{0pt}
\setlength{\belowrulesep}{0pt}
\setlength{\cmidrulesep}{0pt}
\renewcommand{\arraystretch}{1.16}
\arrayrulecolor{black}
\resizebox{\linewidth}{!}{%
\begin{tabular}{@{}cl|*{3}{r}|*{3}{r}|*{3}{r}|*{3}{r}|*{3}{r}|*{3}{r}@{}}
\toprule
\rowcolor{bThirtyTwoHeader}
& & \multicolumn{6}{c|}{\textbf{25\%}}
  & \multicolumn{6}{c|}{\textbf{50\%}}
  & \multicolumn{6}{c}{\textbf{75\%}} \\
\cmidrule{3-20}
\rowcolor{bThirtyTwoHeader}
& & \multicolumn{3}{c|}{R@1 (\%)} & \multicolumn{3}{c|}{RSum}
  & \multicolumn{3}{c|}{R@1 (\%)} & \multicolumn{3}{c|}{RSum}
  & \multicolumn{3}{c|}{R@1 (\%)} & \multicolumn{3}{c}{RSum} \\
\rowcolor{bThirtyTwoHeader}
& & \multicolumn{1}{c}{CM} & \multicolumn{1}{c}{RM} & \multicolumn{1}{c|}{WD}
  & \multicolumn{1}{c}{CM} & \multicolumn{1}{c}{RM} & \multicolumn{1}{c|}{WD}
  & \multicolumn{1}{c}{CM} & \multicolumn{1}{c}{RM} & \multicolumn{1}{c|}{WD}
  & \multicolumn{1}{c}{CM} & \multicolumn{1}{c}{RM} & \multicolumn{1}{c|}{WD}
  & \multicolumn{1}{c}{CM} & \multicolumn{1}{c}{RM} & \multicolumn{1}{c|}{WD}
  & \multicolumn{1}{c}{CM} & \multicolumn{1}{c}{RM} & \multicolumn{1}{c}{WD} \\
\rowcolor{bThirtyTwoHeader}
\multirow{-4}{*}{\rotatebox[origin=c]{90}{Dataset}} & Method
  & \multicolumn{1}{c}{I$\to$T} & \multicolumn{1}{c}{I$\to$T} & \multicolumn{1}{c|}{T$\to$I}
  & \multicolumn{1}{c}{I$\to$T} & \multicolumn{1}{c}{I$\to$T} & \multicolumn{1}{c|}{T$\to$I}
  & \multicolumn{1}{c}{I$\to$T} & \multicolumn{1}{c}{I$\to$T} & \multicolumn{1}{c|}{T$\to$I}
  & \multicolumn{1}{c}{I$\to$T} & \multicolumn{1}{c}{I$\to$T} & \multicolumn{1}{c|}{T$\to$I}
  & \multicolumn{1}{c}{I$\to$T} & \multicolumn{1}{c}{I$\to$T} & \multicolumn{1}{c|}{T$\to$I}
  & \multicolumn{1}{c}{I$\to$T} & \multicolumn{1}{c}{I$\to$T} & \multicolumn{1}{c}{T$\to$I} \\
\midrule
\multirow{3}{*}{\rotatebox[origin=c]{90}{\small\itshape\shortstack{Flickr30K\\\cite{young2014flickr}}}}& SPARC (seed 42)
  & 70.60 & 72.00 & 51.94 & 254.70 & 256.30 & 215.70
  & 52.50 & 53.60 & 32.80 & 216.10 & 215.80 & 158.14
  & 30.20 & 29.70 & 12.74 & 145.20 & 150.10 & 82.08 \\
 & SPARC (seed 43)
  & 70.80 & 71.60 & 51.50 & 255.80 & 255.20 & 216.36
  & 52.50 & 53.70 & 32.52 & 215.90 & 216.80 & 157.42
  & 30.00 & 29.70 & 13.30 & 145.40 & 149.20 & 82.38 \\
 & SPARC (seed 44)
  & 71.30 & 72.10 & 51.90 & 254.80 & 256.50 & 216.48
  & 52.50 & 54.10 & 32.84 & 216.90 & 216.90 & 159.76
  & 30.40 & 30.00 & 12.78 & 145.60 & 150.10 & 82.08 \\
\midrule
\multirow{3}{*}{\rotatebox[origin=c]{90}{\small\itshape\shortstack{COCO\\\cite{lin2014coco}}}}& SPARC (seed 42)
  & 44.70 & 46.12 & 27.63 & 194.68 & 196.94 & 143.94
  & 31.16 & 32.14 & 13.84 & 152.04 & 154.98 & 88.43
  & 14.84 & 15.40 & 4.01 & 88.36 & 91.48 & 34.50 \\
 & SPARC (seed 43)
  & 44.52 & 46.28 & 27.63 & 193.46 & 197.42 & 143.92
  & 31.88 & 31.78 & 14.24 & 153.80 & 153.14 & 88.95
  & 15.14 & 15.58 & 4.38 & 88.66 & 91.86 & 34.56 \\
 & SPARC (seed 44)
  & 44.52 & 46.08 & 27.60 & 194.46 & 196.34 & 143.01
  & 31.52 & 31.60 & 14.09 & 153.04 & 153.76 & 88.66
  & 14.90 & 15.46 & 4.42 & 89.08 & 92.06 & 35.47 \\
\bottomrule
\end{tabular}%
}
\endgroup
\end{table}

\FloatBarrier
\section{RSG Component Ablations}
\label{sec:app_rsg_ablation}

\paragraph{Controlled Protocol.}
We keep SC, MRC, InfoNCE, data, training budget, and the checkpoint-selection rule fixed, changing only one RSG component at a time. Removing reference consistency deletes the entire $\varphi_i$ in Eq.~(3), including the anchor distance, coordinate distances, and relative-improvement hinges, while retaining the anchor-relation KL and identity terms. Removing anchor-relation KL deletes only $\mathrm{KL}(\mathbf p_i\Vert\tilde{\mathbf p}_i)$. Removing identity sets $\lambda_{\mathrm{id}}=0$, compared with $0.1$ in full SPARC. All entries are means $\pm$ sample standard deviation over three training seeds.

\begin{table}[!htbp]
\centering
\caption{\textbf{RSG Component Ablations.} Incomplete-query mR@1 (\%) averages the nine incomplete-query conditions with equal weights. Complete-query I$\to$T R@1 (\%) uses CLIP ViT-B/32. All values are means $\pm$ sample SD across three training seeds. B/32 and L/14 denote CLIP~\cite{radford2021clip} backbones. Best means are bold.}
\label{tab:app_rsg_ablation}
\begingroup
\fontsize{8.6}{10.5}\selectfont
\setlength{\tabcolsep}{1.8pt}
\renewcommand{\arraystretch}{1.12}
\settowidth{\cmpmethodwidth}{w/o Complete-Input Identity}
\addtolength{\cmpmethodwidth}{3pt}
\setlength{\cmpnumwidth}{\dimexpr(\linewidth-\cmpmethodwidth-12\tabcolsep)/6\relax}
\begin{tabular}{@{}>{\raggedright\arraybackslash}p{\cmpmethodwidth}*{6}{>{\raggedleft\arraybackslash}p{\cmpnumwidth}}@{}}
\toprule
& \multicolumn{4}{c}{\shortstack{Incomplete queries\\mR@1 (\%)}} & \multicolumn{2}{c}{\shortstack{Complete queries\\I$\to$T R@1 (\%)}}\\
\cmidrule(lr){2-5}\cmidrule(lr){6-7}
Setting & \shortstack{Flickr30K~\cite{young2014flickr}\\B/32} & \shortstack{COCO~\cite{lin2014coco}\\B/32} & \shortstack{Flickr30K~\cite{young2014flickr}\\L/14} & \shortstack{COCO~\cite{lin2014coco}\\L/14} & \shortstack{Flickr30K~\cite{young2014flickr}\\B/32} & \shortstack{COCO~\cite{lin2014coco}\\B/32}\\
\midrule
\rowcolor{oursrow}
\mbox{Full SPARC} & $\mathbf{45.17\pm0.14}$ & $\mathbf{25.61\pm0.09}$ & $\mathbf{57.20\pm0.32}$ & $\mathbf{34.59\pm0.38}$ & $\mathbf{80.40\pm0.20}$ & $\mathbf{54.58\pm0.21}$\\
\mbox{w/o Reference Consistency} & $44.55\pm0.19$ & $25.28\pm0.14$ & $56.71\pm0.30$ & $34.04\pm0.33$ & $80.25\pm0.14$ & $54.46\pm0.25$\\
\mbox{w/o Anchor-Relation KL} & $44.91\pm0.17$ & $25.45\pm0.16$ & $56.95\pm0.33$ & $34.36\pm0.37$ & $80.32\pm0.28$ & $54.53\pm0.24$\\
\mbox{w/o Complete-Input Identity} & $45.10\pm0.20$ & $25.57\pm0.05$ & $57.11\pm0.40$ & $34.52\pm0.36$ & $79.50\pm0.28$ & $53.73\pm0.20$\\
\bottomrule
\end{tabular}
\endgroup
\end{table}

\paragraph{Complementary Effects.}
Each RSG component is evaluated on the quantity it is designed to affect. Removing reference consistency produces the largest mean mR@1 decrease in all four dataset--backbone settings, with losses of 0.62, 0.33, 0.49, and 0.55 points, respectively, each exceeding the corresponding seed SDs. Removing anchor-relation KL lowers mean mR@1 by 0.16--0.26 points; the effect is small relative to the seed SDs but has the same sign in all four settings. This retrieval ablation alone does not establish neighborhood preservation, which requires a corresponding structural metric. Removing complete-input identity barely changes incomplete-query mR@1 (0.04--0.09 points), as expected for a term that acts only on complete inputs, but lowers complete-query I$\to$T R@1 by 0.90 points on Flickr30K and 0.85 points on COCO, three to four times the larger seed SD. These results support complementary roles for semantic-reference guidance, relation regularization, and the identity term in complete-query I$\to$T retrieval. The identity term acts as a soft penalty rather than a hard constraint: complete-query retrieval still changes after calibration (Table~2 of the main paper), most strongly for text queries.

\FloatBarrier
\section{Calibration Geometry and Match Preservation}
\label{sec:app_geometry}

\paragraph{Controlled Variants for Fig.~7.}
Table~\ref{tab:app_geometry_variants} defines the five variants compared in Fig.~7 of the main paper. All variants use the same RSG, network capacity, InfoNCE, data, training budget, training seeds, and checkpoint-selection rule. Only the update parameterization, mixing weight, and ranking constraint differ as specified. Direct uses SC and RSG without MRC, that is, $\beta=1$ and no ranking constraint; it is therefore the same configuration as A2 in Table~4, and its mR@1 and mRSum coincide with the A2 entries. Normalized Residual uses $g_m(\mathbf z)=\operatorname n(\mathbf z+h_m(\mathbf z))$ in place of the tangent-projected update, with all other settings identical to SPARC.

\begin{table}[!htbp]
\centering
\caption{\textbf{Controlled Calibration Variants.} The remaining RSG, network, and training settings are shared. Ranking constraint refers to the soft ranking term of $\mathcal L_{\mathrm{MRC}}$; feature mixing is controlled separately by $\beta$.}
\label{tab:app_geometry_variants}
\small
\setlength{\tabcolsep}{5pt}
\renewcommand{\arraystretch}{1.10}
\begin{tabularx}{\linewidth}{@{}l>{\raggedright\arraybackslash}Xcc@{}}
\toprule
Variant & Update & Mixing Weight $\beta$ & Ranking Constraint\\
\midrule
Direct (= A2) & Tangent Update & 1.0 & Off\\
Normalized Residual & Residual then Normalization & 0.6 & On\\
Mixed & Tangent Update & 0.6 & Off\\
Ranking Regularized & Tangent Update & 1.0 & On\\
\rowcolor{oursrow}
SPARC & Tangent Update & 0.6 & On\\
\bottomrule
\end{tabularx}
\end{table}

\paragraph{Cohorts, Rates, and Aggregation.}
The original frozen-query prediction defines the correct and incorrect cohorts separately within each missingness condition. With the complete gallery fixed, we compute
\begin{equation}
\begin{aligned}
\mathrm{Deg.}&=\frac{\#(\text{originally correct}\to\text{incorrect})}{\#(\text{originally correct})}\times100\%,\\
\mathrm{Rep.}&=\frac{\#(\text{originally incorrect}\to\text{correct})}{\#(\text{originally incorrect})}\times100\%.
\end{aligned}
\label{eq:app_deg_rep}
\end{equation}
For each training seed, rates are computed separately in each condition and then averaged with equal weights over the nine conditions. We report the mean and sample SD of the three per-seed averages in Table~\ref{tab:app_geometry}. These are absolute, cohort-conditional transition rates. Fig.~7 of the main paper displays the same measurements as relative improvements (\%) over Direct: the relative reduction in Deg. and the relative gain in Rep., mR@1, and mRSum.

\begin{table}[!htbp]
\centering
\caption{\textbf{Calibration Geometry and Match Preservation, CLIP ViT-B/32.} Means $\pm$ sample SD over three training seeds. mR@1 and mRSum are higher-is-better; Deg. is lower-is-better and Rep. is higher-is-better. All four metrics average the nine incomplete-query conditions within each seed. Direct coincides with A2 in Table~4, and SPARC with Table~\ref{tab:app_multiseed}. Best means are bold.}
\label{tab:app_geometry}
\begingroup
\fontsize{9.5}{11.3}\selectfont
\setlength{\tabcolsep}{4pt}
\renewcommand{\arraystretch}{1.12}
\begin{tabularx}{\linewidth}{@{}l>{\raggedright\arraybackslash}Xrrrr@{}}
\toprule
Dataset & Variant & mR@1 (\%) $\uparrow$ & mRSum $\uparrow$ & Deg. (\%) $\downarrow$ & Rep. (\%) $\uparrow$\\
\midrule
\multirow{5}{*}{\rotatebox[origin=c]{90}{\shortstack{Flickr30K\\\cite{young2014flickr}}}} & Direct & $44.33\pm0.10$ & $186.24\pm0.34$ & $7.00\pm0.19$ & $22.84\pm0.13$\\
& Normalized Residual & $44.65\pm0.17$ & $187.10\pm0.20$ & $6.20\pm0.09$ & $22.88\pm0.31$\\
& Mixed & $44.73\pm0.07$ & $187.10\pm0.28$ & $5.80\pm0.08$ & $22.84\pm0.13$\\
& Ranking Regularized & $44.99\pm0.13$ & $187.95\pm0.16$ & $5.30\pm0.03$ & $22.88\pm0.21$\\
\rowcolor{oursrow}
& SPARC & $\mathbf{45.17\pm0.14}$ & $\mathbf{188.43\pm0.31}$ & $\mathbf{5.00\pm0.08}$ & $\mathbf{22.99\pm0.18}$\\
\midrule
\multirow{5}{*}{\rotatebox[origin=c]{90}{\shortstack{COCO\\\cite{lin2014coco}}}} & Direct & $25.10\pm0.11$ & $125.87\pm0.15$ & $8.00\pm0.21$ & $14.20\pm0.10$\\
& Normalized Residual & $25.28\pm0.12$ & $126.60\pm0.07$ & $7.10\pm0.16$ & $14.25\pm0.11$\\
& Mixed & $25.32\pm0.09$ & $126.48\pm0.03$ & $6.50\pm0.08$ & $14.20\pm0.10$\\
& Ranking Regularized & $25.47\pm0.09$ & $126.96\pm0.03$ & $5.90\pm0.02$ & $14.24\pm0.11$\\
\rowcolor{oursrow}
& SPARC & $\mathbf{25.61\pm0.09}$ & $\mathbf{127.30\pm0.03}$ & $\mathbf{5.50\pm0.10}$ & $\mathbf{14.33\pm0.09}$\\
\bottomrule
\end{tabularx}
\endgroup
\end{table}

\paragraph{Interpretation.}
Direct versus Mixed isolates feature mixing. Deg. decreases from 7.00\% to 5.80\% on Flickr30K and from 8.00\% to 6.50\% on COCO, relative reductions of 17\% and 19\%, while Rep. is unchanged, so the smaller effective step mainly protects existing matches. Mixing an original and an updated feature resembles the residual ratio of feature adapters such as CLIP-Adapter~\cite{gao2021adapter}; under the tangent parameterization it reduces to a shorter tangent step with a closed-form step factor (Lemma~\ref{lem:nt-step}), which makes its effect on matching margins analyzable (Proposition~\ref{prop:nt-mrc}). Mixed versus SPARC isolates the ranking constraint at $\beta=0.6$: Deg. decreases further to 5.00\% and 5.50\%, relative reductions of 14\% and 15\%, and Rep. rises slightly, from 22.84\% to 22.99\% and from 14.20\% to 14.33\%. Normalized Residual versus SPARC compares update parameterizations with all remaining settings matched. SPARC attains higher mR@1 (45.17\% vs.\ 44.65\% and 25.61\% vs.\ 25.28\%) and lower Deg. (5.00\% vs.\ 6.20\% and 5.50\% vs.\ 7.10\%), relative reductions of 19\% and 23\% with seed SDs of at most 0.16 points. The accuracy difference is modest, whereas the reduction in Deg. is large relative to seed variation, so the tangent parameterization contributes mainly through match preservation. Across all variants, the differences in Rep. are small relative to the seed SDs, so the principal effect of feature mixing, the ranking constraint, and the tangent parameterization is on match preservation (Deg.).

\FloatBarrier
\section{Additional Analysis of Retained Semantic Knowledge}
\label{sec:app_semantics}

\subsection{Diagnostic Feature Locations and Annotation Sources}
\label{sec:app_diagnostic_locations}

\paragraph{Local Target-Region Features.}
For the local category probe, we extract patch hidden states after the final frozen CLIP visual Transformer block and before \texttt{ln\_post} and the visual projection. We exclude CLS. Let $h_{ij}$ be patch $j$ of image $i$, $P_{ij}$ its pixel support after preprocessing, and $B_i$ the target region in the same resized/cropped coordinate system. The region representation is
\begin{equation}
 a_{ij}=\operatorname{area}(P_{ij}\cap B_i),\qquad
 r_i=\operatorname n\!\left(\frac{\sum_j a_{ij}h_{ij}}{\sum_j a_{ij}}\right),
 \quad\sum_j a_{ij}>0.
\label{eq:app_region_pooling}
\end{equation}
Thus overlapping patches contribute in proportion to their target-overlap area. Pooling precedes the final visual layer normalization and projection; the pooled region vector is then $\ell_2$-normalized for the probe. The target class, transformed box, and contributing patch indices are retained with each record.

\paragraph{Global Retrieval Features.}
The image branch uses the final CLS hidden state, followed by \texttt{ln\_post} and the visual projection. The text branch applies \texttt{ln\_final} to the Transformer output and projects the EOT vector with the text projection. Both projected features are $\ell_2$-normalized before cosine matching. This order follows the \href{https://github.com/openai/CLIP/blob/main/clip/model.py}{official OpenAI CLIP implementation}~\cite{radford2021clip}.

\paragraph{SAE Diagnostic Representation.}
For B1 and B2, a separate Top-$K$ SAE~\cite{gao2024sae} is trained for each dataset and backbone on the projected, $\ell_2$-normalized global features of complete training images (512-dimensional for CLIP ViT-B/32 and 768-dimensional for CLIP ViT-L/14). Its pre-TopK activations and Top-$K$ codes are used only for diagnosis; it does not encode patch states or raw CLS/EOT hidden states.

\begin{table}[!htbp]
\centering
\caption{\textbf{Feature Locations and Annotation Records for Fig.~2.} B1 and B2 share the same frozen SAE on projected, normalized global image features. The natural word-deletion extension of C1 uses global EOT retrieval features. Region boxes and patch indices refer to the preprocessed image coordinates.}
\label{tab:app_panel_locations}
\begingroup
\fontsize{9}{10.8}\selectfont
\setlength{\tabcolsep}{4pt}
\renewcommand{\arraystretch}{1.18}
\begin{tabularx}{\linewidth}{@{}>{\raggedright\arraybackslash}p{0.22\linewidth}>{\raggedright\arraybackslash}p{0.36\linewidth}>{\raggedright\arraybackslash}X@{}}
\toprule
Panel & Feature location & Annotations and records\\
\midrule
A: local category readability & Final CLIP~\cite{radford2021clip} visual patch states before \texttt{ln\_post} and projection; overlap-area pooling and normalization & Target category, target box (Flickr30K Entities~\cite{plummer2017entities} or COCO~\cite{lin2014coco}), and patch indices after preprocessing\\
\addlinespace[2pt]
B1: SAE / pre-TopK responses & Projected, normalized global image features input to the fixed Top-$K$ SAE~\cite{gao2024sae}; pre-activations and Top-$K$ codes & Category-associated unit lists and fixed unit-selection rules\\
\addlinespace[2pt]
B2: semantic specificity & Exactly the same features and SAE as B1 & Semantic units, matched control units, and the matching criteria\\
\addlinespace[2pt]
B3: category-direction response & Projected, normalized global image features & Fixed category directions determined from clean training data\\
\addlinespace[2pt]
C1: readable but mismatched & Global retrieval features; final EOT retrieval features for natural text missingness & Category labels, frozen readability thresholds, and true-pair IDs\\
\addlinespace[2pt]
C2: positive/negative similarity decline & The same global query features against the fixed complete gallery & Positive IDs and the hardest-negative selection rule\\
\addlinespace[2pt]
C3: semantic-component removal & Remove the designated component in global retrieval space, then renormalize & Semantic subspace, random controls, and perturbation magnitudes\\
\addlinespace[2pt]
C4: fine-grained cue control & The same global image retrieval features for the fine-grained probe and two-candidate matching & Attribute/relation labels, cue regions, correct/distractor captions, and cue-preservation/erasure verification\\
\bottomrule
\end{tabularx}
\endgroup
\end{table}

\paragraph{Annotation Provenance.}
For Flickr30K, entity phrases, coarse phrase types, and region boxes come from \href{https://github.com/BryanPlummer/flickr30k_entities}{Flickr30K Entities}~\cite{plummer2017entities} and are joined to the actual retrieval split by image ID. For COCO, categories and boxes come from the corresponding COCO instance-annotation release~\cite{lin2014coco} and are linked by image ID to the image--caption pairs. The \href{https://github.com/cocodataset/cocoapi/blob/master/PythonAPI/pycocotools/coco.py}{official COCO API} distinguishes instance annotations from caption annotations. Captions supply candidate attributes and relations for screening; cue presence, preservation, and erasure are then established with cue-region annotations and verification records rather than by caption screening alone. The panel-specific records are listed in Table~\ref{tab:app_panel_locations}.

\subsection{Probe and SAE Configuration}
\label{sec:app_probe_sae_configuration}

Table~\ref{tab:app_probe_sae_configuration} lists the diagnostic configurations. Probe coefficients, readability thresholds, SAE parameters, and semantic-unit selection rules are all fixed before test analysis.

\begin{table}[!htbp]
\centering
\caption{\textbf{Probe and SAE Configurations.} Hyperparameters are selected on validation data.}
\label{tab:app_probe_sae_configuration}
\begingroup
\fontsize{9.5}{11.3}\selectfont
\setlength{\tabcolsep}{5pt}
\renewcommand{\arraystretch}{1.18}
\begin{tabularx}{\linewidth}{@{}>{\raggedright\arraybackslash}p{0.24\linewidth}>{\raggedright\arraybackslash}X@{}}
\toprule
Item & Configuration\\
\midrule
Category / fine-grained probes & Separate linear one-vs-rest logistic regressions with $\ell_2$ regularization; $C\in\{0.01,0.1,1,10\}$ selected by validation AP\\
\addlinespace[2pt]
Readability thresholds & Class-specific thresholds selected on validation data and frozen before testing; AP itself does not require these thresholds\\
\addlinespace[2pt]
SAE inputs & Projected, normalized global image features from clean training images; separate SAE for each dataset and backbone\\
\addlinespace[2pt]
SAE architecture & Top-$K$ SAE~\cite{gao2024sae}; expansion factor 8; 4,096 latent units for ViT-B/32 and 6,144 for ViT-L/14; Top-$K=32$\\
\addlinespace[2pt]
SAE optimization & Adam, learning rate $10^{-3}$, batch size 256, at most 50 epochs; checkpoint selected by validation reconstruction error\\
\addlinespace[2pt]
Data separation & Captions, regions, and corrupted views from one source image remain in the same split; test data are excluded from hyperparameter selection\\
\bottomrule
\end{tabularx}
\endgroup
\end{table}

\subsection{Panel-Specific Sample Counts and Repeated Measurements}
\label{sec:app_panel_samples}

Table~\ref{tab:app_panel_samples} reports the sample counts for the target-preserving diagnostics of Fig.~2. The natural center-masking and word-deletion extension has its own initial cohorts in Table~\ref{tab:app_rbm}; the two sets of counts follow different eligibility rules and are not interchangeable.

\begin{table}[!htbp]
\centering
\caption{\textbf{Panel-Specific Sample Counts for Fig.~2.} Counts for the target-preserving diagnostics. B/32 and L/14 denote CLIP~\cite{radford2021clip} backbones. A dash indicates that the panel is not evaluated in that setting. Units and cohort reuse are stated explicitly.}
\label{tab:app_panel_samples}
\begingroup
\fontsize{9}{10.8}\selectfont
\setlength{\tabcolsep}{4pt}
\renewcommand{\arraystretch}{1.18}
\begin{tabularx}{\linewidth}{@{}lrrrr>{\raggedright\arraybackslash}X@{}}
\toprule
Panel & \shortstack{Flickr30K~\cite{young2014flickr}\\B/32} & \shortstack{Flickr30K~\cite{young2014flickr}\\L/14} & \shortstack{COCO~\cite{lin2014coco}\\B/32} & \shortstack{COCO~\cite{lin2014coco}\\L/14} & Counting unit\\
\midrule
A & 800 & 800 & 4,000 & 4,000 & Images / target regions\\
B1 & 800 & 800 & 4,000 & 4,000 & Images\\
B2 & 800 & 800 & 4,000 & 4,000 & Images, with controls paired to the original samples\\
B3 & 800 & 800 & 4,000 & 4,000 & Images\\
C1 & 620 & 650 & 1,600 & 2,000 & Queries readable and correctly retrieved under complete input\\
C2 & 620 & 650 & 1,600 & 2,000 & The same query cohort as C1\\
C3 & 600 & 630 & 1,550 & 1,930 & Valid queries meeting intervention and control requirements\\
C4 & 400 & --- & --- & --- & 400 pairs from 400 source images\\
\bottomrule
\end{tabularx}
\endgroup
\end{table}

\paragraph{Independent Sources versus Repeated Records.}
For A--B3, the complete, 25\%, 50\%, and 75\% context-masked views are repeated measurements of the same images. Thus 800 images do not become 3,200 independent samples, and 4,000 images do not become 16,000 independent samples. The C4 set contains 200 attribute pairs and 200 relation pairs. Its complete, cue-preserved, and cue-erased views yield 1,200 records, while the number of independent source images remains 400. Additional random-control draws and SAE units likewise do not increase the independent sample count.

\subsection{Paired Image-Cluster Bootstrap}
\label{sec:app_cluster_bootstrap}

\paragraph{Reporting Convention and Resampling Unit.}
Confidence intervals for the Fig.~2 statistics and for Table~\ref{tab:app_rbm} are two-sided 95\% percentile intervals, rather than standard deviations or standard errors. Within each dataset, the resampling unit is the source image ID. We draw 5,000 resamples with replacement from the image IDs; every occurrence of a sampled image carries all its captions, regions, corrupted views, and intervention results. This preserves the pairing across conditions and the dependence among records from the same image.

\paragraph{Recomputing the Statistic in Every Replicate.}
For each resampled collection of image clusters, the complete panel-specific statistic is recomputed. If $\hat\theta^{*(b)}$ denotes replicate $b$, the percentile interval is
\begin{equation}
\mathrm{CI}_{95\%}=\left[Q_{0.025}\!\left(\{\hat\theta^{*(b)}\}_{b=1}^{5000}\right),\;
Q_{0.975}\!\left(\{\hat\theta^{*(b)}\}_{b=1}^{5000}\right)\right].
\label{eq:app_bootstrap_ci}
\end{equation}
The encoder, probe, SAE, semantic/category directions, readability thresholds, original cohort-eligibility flags, and complete gallery remain fixed. Fixing cohort membership does not fix its resampled count: numerators and eligible denominators are summed anew using the sampled image multiplicities. AP ratios recompute both the complete-input and partial-input AP before taking their ratio. Comparisons between conditions use paired differences within the same bootstrap replicate. Table~\ref{tab:app_bootstrap_panels} lists the statistic for each panel.

\begin{table}[!htbp]
\centering
\caption{\textbf{Statistics Recomputed in Each Bootstrap Replicate for Fig.~2.}}
\label{tab:app_bootstrap_panels}
\begingroup
\fontsize{9}{10.8}\selectfont
\setlength{\tabcolsep}{4pt}
\renewcommand{\arraystretch}{1.18}
\begin{tabularx}{\linewidth}{@{}l>{\raggedright\arraybackslash}p{0.40\linewidth}>{\raggedright\arraybackslash}X@{}}
\toprule
Panel & Statistic recomputed in every replicate & Pairing or aggregation\\
\midrule
A & $100\times\mathrm{AP}_{\mathrm{partial}}/\mathrm{AP}_{\mathrm{complete}}$ & Both AP values are recomputed before taking their ratio\\
\addlinespace[2pt]
B1 & Attenuation of SAE and pre-TopK responses & Fixed semantic units are aggregated within each image before image clusters are resampled\\
\addlinespace[2pt]
B2 & Semantic-response loss minus matched-control loss & Semantic and control records stay paired; the training-set SD used for standardization is fixed\\
\addlinespace[2pt]
B3 & Category-direction cosine difference between complete and partial features & Paired by image and fixed category direction\\
\addlinespace[2pt]
C1 & Readable-but-mismatched count divided by the initial eligible-cohort count & Original eligibility flags fixed; numerator and denominator reaggregated in each replicate\\
\addlinespace[2pt]
C2 & Positive and negative similarity declines, and their difference & Query--candidate pairing preserved; gallery and candidate-selection rule fixed\\
\addlinespace[2pt]
C3 & Margin loss from semantic removal minus random-control loss & Random controls are aggregated within each query before source-image clusters are resampled\\
\addlinespace[2pt]
C4 & Fine-grained AP, candidate accuracy, and paired differences between views & Complete, cue-kept, and cue-erased records resampled together\\
\bottomrule
\end{tabularx}
\endgroup
\end{table}

\paragraph{Implementation.}
Image IDs are sampled before their records are expanded. Because \texttt{scipy.stats.bootstrap} with \texttt{paired=True} does not infer image-to-caption grouping from flattened arrays~\cite{scipy_bootstrap}, we pass image-cluster membership explicitly and set \texttt{n\_resamples=5000}, \texttt{confidence\_level=0.95}, and \texttt{method='percentile'}.

\paragraph{Scope of Uncertainty.}
These intervals characterize sampling uncertainty conditional on the fixed diagnostic models, eligibility rules, and gallery. They exclude variability from training initialization, which is reported separately as the seed SD, and changes to the gallery.

\subsection{Global Readability and Matching under Natural Missingness}
\label{sec:app_rbm}

\paragraph{Initial Cohort and Denominator.}
We examine frozen CLIP ViT-B/32 under 50\% center masking for image queries and 50\% word deletion for text queries. Category readability is evaluated by the fixed category readout used for the global-feature analysis. For complete and incomplete query features $q_c$ and $q_p$, define the initial cohort
\begin{equation}
\mathcal C=\{q:\text{category-readable on }q_c\text{ and Top-1 retrieval correct on }q_c\},\qquad n_0=|\mathcal C|.
\label{eq:app_rbm_cohort}
\end{equation}
The readable-but-mismatched (RBM) rate is
\begin{equation}
p_{\mathrm{RBM}}=
\frac{\#\{q\in\mathcal C:\text{category-readable on }q_p\text{ and Top-1 retrieval incorrect on }q_p\}}
{n_0}\times100\%.
\label{eq:app_rbm_rate}
\end{equation}
The denominator is the entire initial cohort, including queries that cease to be readable after corruption. For example, the first row of Table~\ref{tab:app_rbm} gives $155/700=22.14\%$, whereas $155/490$ would measure the matching-error rate conditional on remaining readable.

\begin{table}[!htbp]
\centering
\caption{\textbf{Readable-but-Mismatched Cohort Summaries, Frozen CLIP ViT-B/32.} Rates use the initial-cohort denominator $n_0$. The 95\% intervals are paired image-cluster bootstrap intervals (Section~\ref{sec:app_cluster_bootstrap}) computed from the query-level labels. CM denotes center masking and WD denotes word deletion.}
\label{tab:app_rbm}
\begingroup
\fontsize{9}{10.8}\selectfont
\setlength{\tabcolsep}{4pt}
\renewcommand{\arraystretch}{1.14}
\begin{tabularx}{\linewidth}{@{}l>{\raggedright\arraybackslash}Xrrrrl@{}}
\toprule
Dataset & Query and condition & \shortstack{Initial cohort\\$n_0$} & \shortstack{Still\\readable} & \shortstack{Readable but\\mismatched} & \shortstack{RBM rate\\(\%)} & 95\% CI\\
\midrule
\multirow{2}{*}{Flickr30K~\cite{young2014flickr}} & I$\to$T, 50\% CM & 700 & 490 & 155 & \textbf{22.14} & [19.14\%, 25.23\%]\\
& T$\to$I, 50\% WD & 2,500 & 1,800 & 662 & \textbf{26.48} & [24.61\%, 28.39\%]\\
\midrule
\multirow{2}{*}{COCO~\cite{lin2014coco}} & I$\to$T, 50\% CM & 1,250 & 900 & 359 & \textbf{28.72} & [26.27\%, 31.21\%]\\
& T$\to$I, 50\% WD & 6,000 & 3,900 & 1,869 & \textbf{31.15} & [29.93\%, 32.38\%]\\
\bottomrule
\end{tabularx}
\endgroup
\end{table}

\paragraph{Interpretation and Scope.}
In these four cohorts, 22.14--31.15\% of queries that are category-readable and correctly retrieved under complete inputs remain readable but lose the correct Top-1 match after 50\% center masking or word deletion. It shows that accessible category evidence need not translate into effective candidate discrimination under native cosine retrieval, supporting separate examination of semantic readability and matching utility. This category-level comparison alone does not establish that all instance-specific details survive or identify SPARC's recovery mechanism.

\subsection{Semantic Recovery and Blocking for Fig.~5}
\label{sec:app_figure5_protocol}

\paragraph{Recovery Cohorts and Matched Comparisons.}
The left panel begins with queries whose frozen-CLIP Top-1 retrieval is incorrect under incomplete input. SPARC partitions this cohort into recovered queries (correct after calibration) and unrecovered queries (still incorrect). The two groups are matched as specified in Table~\ref{tab:app_figure5_matching}. After matching, fixed semantic readouts show category-margin gains of 1.25--2.83 logits for recovered queries, exceeding the gains of their matched unrecovered partners by 0.86--1.82 logits.

\begin{table}[!htbp]
\centering
\caption{\textbf{Matched-Cohort Protocol for Fig.~5 (Left).} Match covariates are measured before calibration. Cohort counts, pair IDs, matching failures, and postmatch balance are kept in the matching records rather than inferred from aggregate retrieval scores.}
\label{tab:app_figure5_matching}
\begingroup
\fontsize{9.5}{11.3}\selectfont
\setlength{\tabcolsep}{5pt}
\renewcommand{\arraystretch}{1.18}
\begin{tabularx}{\linewidth}{@{}>{\raggedright\arraybackslash}p{0.23\linewidth}>{\raggedright\arraybackslash}X@{}}
\toprule
Item & Definition\\
\midrule
Original cohort & Frozen CLIP retrieves an incorrect Top-1 candidate for the incomplete query\\
Recovered group & SPARC's calibrated query retrieves a correct Top-1 candidate\\
Unrecovered group & The calibrated query still retrieves an incorrect Top-1 candidate\\
Strata & Matching is separate by dataset, backbone, query modality, perturbation type, and severity\\
Match covariates & Frozen-feature retrieval margin and category-readout margin\\
Algorithm & 1:1 nearest-neighbor matching without replacement; fixed query ordering and tie-breaking rules\\
Caliper & Each standardized covariate differs by at most 0.2 SD; prespecified 0.1 and 0.3 SD sensitivity analyses\\
Records & Prematch group sizes, successful pair count, unmatched-query reasons, pair IDs, and postmatch covariate balance\\
\bottomrule
\end{tabularx}
\endgroup
\end{table}

Standardization statistics and the distance, ordering, and tie rules are fixed before matching. This comparison associates semantic-readout gains with retrieval recovery after controlling the frozen margins; matching alone does not identify a unique causal mechanism~\cite{ravichander2021probing,elazar2021amnesic}.

\paragraph{Blocking the Effective Tangent Step.}
Semantic directions are determined only from clean training/validation data and fixed before test analysis. Let $q$ be the unit original incomplete-query feature and $F=F_\beta(q)$ its final unit mixed feature. For $q^\top F>0$, define the effective tangent step
\begin{equation}
u=\frac{F}{q^\top F}-q,\qquad q^\top u=0,\qquad F=\operatorname n(q+u).
\label{eq:app_effective_tangent_block}
\end{equation}
Project the fixed semantic directions onto $q$'s tangent space using $I-qq^\top$, and orthogonalize them to obtain a matrix $U_q$ with orthonormal columns. The blocked query is
\begin{equation}
q_{\mathrm{block}}=\operatorname n\!\left(q+(I-U_qU_q^\top)u\right).
\label{eq:app_blocked_query}
\end{equation}
This removes the designated component of the effective tangent step, including feature mixing. It does not treat the chord $F-q$ as an already tangent displacement. The positive-inner-product condition and the retained subspace rank are recorded with intervention eligibility.

\paragraph{Random Controls.}
The random-subspace control has the same rank as the semantic subspace and lies in the same tangent space. The random-tangent control uses a random tangent direction. The perturbation magnitude/angular-displacement matching rule is fixed across semantic and random interventions, with the realized displacement retained in the control records. The query, encoder, original calibrated feature, positive IDs, and complete gallery remain unchanged across interventions. These are directional controls on feature use, following the general distinction between probe readability and behavioral relevance~\cite{elazar2021amnesic}.

\paragraph{Two Separately Reported Endpoints.}
The recovered cohort $\mathcal R$ is fixed before any blocking, using the frozen-error and calibrated-correct criteria above. The Top-1 recovery loss is
\begin{equation}
L_{\mathrm{recovery}}=\frac{\#\{q\in\mathcal R:\text{Top-1 retrieval with }q_{\mathrm{block}}\text{ is incorrect}\}}
{|\mathcal R|}\times100\%.
\label{eq:app_top1_recovery_loss}
\end{equation}
The matching-margin endpoint is instead
\begin{equation}
L_{\mathrm{margin}}=\operatorname{mean}_{q\in\mathcal Q_{\mathrm{eval}}}
\left[m(F_\beta(q))-m(q_{\mathrm{block}})\right],
\label{eq:app_block_margin_loss}
\end{equation}
where $\mathcal Q_{\mathrm{eval}}$ is the intervention cohort and the margin is the best-positive minus hardest-negative margin against the fixed complete gallery. When both endpoints are measured on recovered queries, $\mathcal Q_{\mathrm{eval}}=\mathcal R$. Semantic-versus-control comparisons are computed separately for each endpoint. Fig.~5 (right) reports excess Top-1 recovery loss, reaching 28.67\% and 31.36\% relative to random-span and random-tangent controls on COCO with ViT-B/32. The margin endpoint, measured on the same blocking intervention in all four dataset--backbone settings, gives positive excess margin losses of $[0.32,0.41]\times10^{-3}$ relative to random-tangent controls and $[0.24,0.33]\times10^{-3}$ relative to same-rank random subspaces (for COCO B/32, $0.37\times10^{-3}$ and $0.33\times10^{-3}$).

\subsection{From Retained Target Cues to Native Retrieval Features}
\label{sec:app_native}
The main-paper analyses distinguish semantic availability, feature expression, and matching utility. With annotated target cues preserved while their context is masked, local category-probe AP retains 95.7--97.2\% of its complete-input reference at 75\% context masking (Fig.~2A), whereas SAE and category-direction responses in the global embedding weaken as masking increases (Fig.~2B1--B3). These results motivate studying how the remaining semantic information participates in matching.

\begin{figure}[t]
  \centering
  \includegraphics[width=\linewidth]{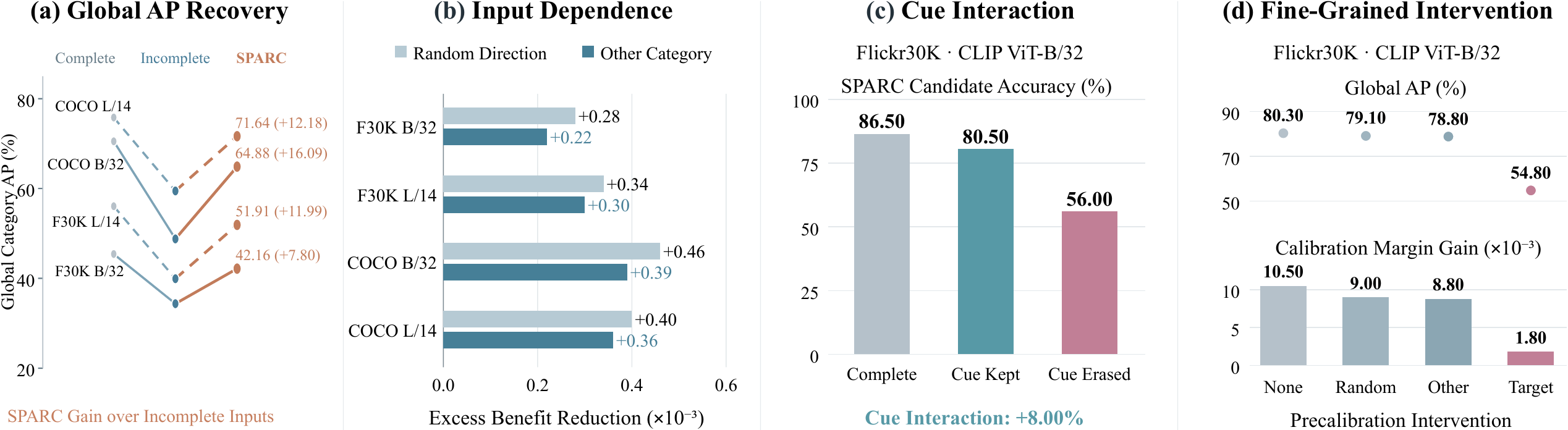}
  \caption{\textbf{Semantic Readability and Functional Use of Retained Semantics in the Retrieval Features.}
  \textbf{(a)} Global category AP for complete inputs, incomplete inputs, and SPARC under 50\% center masking without target protection (the condition of Table~\ref{tab:app_rbm}). Solid and dashed lines denote ViT-B/32 and ViT-L/14; orange labels give SPARC AP and its gain over incomplete inputs in percentage points.
  \textbf{(b)} Excess reduction in calibration benefit, in fixed-gallery matching-margin units of $10^{-3}$, when the incomplete input is weakened along its target-semantic direction before calibration, relative to equal-angle random-direction and other-category controls.
  \textbf{(c)} SPARC candidate accuracy on the Flickr30K CLIP ViT-B/32 two-caption task for complete, cue-kept, and cue-erased inputs. The corresponding frozen-CLIP accuracies (86.0\%, 71.0\%, and 54.5\%) and fine-grained probe AP are shown in Fig.~2C4 of the main paper. The cue interaction is the difference between the SPARC gains on cue-kept and cue-erased inputs, $9.50-1.50=8.00$ percentage points.
  \textbf{(d)} Fine-grained probe AP on the global features (top) and the calibration-induced candidate-margin benefit $B$ in units of $10^{-3}$ (bottom; Eq.~\eqref{eq:app_fine_calibration_benefit}) after pre-calibration interventions on cue-kept inputs: none, matched random direction, unrelated semantic direction (Other), and target semantic direction.
  Candidate accuracy is distinct from full-gallery R@1. All values are aggregate point estimates.}
  \label{fig:app_semantic_use}
\end{figure}

Readout location matters for this interpretation. Local probes measure target-support features, whereas native retrieval uses projected, normalized global vectors. A local readout alone therefore does not establish semantic accessibility in the representation used for cosine matching. Figure~\ref{fig:app_semantic_use} complements the main-paper diagnostics with four analyses on the projected retrieval features: global category readability after calibration (a), the dependence of calibration benefits on surviving input semantics (b), and, on the controlled Flickr30K fine-grained task, the cue-availability interaction (c) and pre-calibration fine-grained interventions (d). The CLIP encoders remain frozen, and the measurements concern the output features and the learned calibration map.

\subsection{Improved Category Readability after Calibration}
Figure~\ref{fig:app_semantic_use}(a) applies fixed clean-trained category readouts to the projected features used for retrieval. Under 50\% center masking, SPARC increases AP from 34.36\% to 42.16\% with Flickr30K B/32 and from 39.92\% to 51.91\% with Flickr30K L/14. On COCO, the corresponding changes are 48.79\% to 64.88\% and 59.46\% to 71.64\%. The gains are 7.80, 11.99, 16.09, and 12.18 percentage points, respectively.
The consistent improvement extends the semantic-availability analysis to the actual retrieval representation. Calibration makes category information more accessible to the same readout while keeping the pretrained encoders fixed. The remaining gap to the complete-input reference indicates that partial observation continues to affect the representation. Category AP measures this specific readout capability; the matching-level analyses below examine how semantic directions contribute to calibration benefits.

\subsection{Dependence on Surviving Input Semantics}
We weaken the incomplete input along its target-semantic direction before applying the unchanged calibration map. Equal-angle random directions and directions associated with another category provide controls. Calibration benefit is the improvement in a fixed-gallery matching margin produced by the map. A positive effect means that target-directed weakening reduces this benefit more than the corresponding control.

Figure~\ref{fig:app_semantic_use}(b) shows positive mean effects in all four dataset--backbone settings. Target-versus-random differences lie in $[0.28,0.46]\times10^{-3}$, and target-versus-other-category differences lie in $[0.22,0.39]\times10^{-3}$. COCO B/32 has the largest mean in both comparisons, at $0.46\times10^{-3}$ and $0.39\times10^{-3}$. Together with the blocking analyses of Section~\ref{sec:app_figure5_protocol}, which act on the learned update rather than on the input, these observations connect calibration gains to semantic directions in both the available input and the learned update.

\paragraph{Interpretation and Scope.}
Panels (a) and (b) connect improved category readability with semantic directions in the partial input. Their positive mean effects are consistent with more effective functional use of retained pretrained semantics. Since SPARC acts after frozen encoders, the analysis concerns calibrated features and matching behavior rather than a change to internal CLIP activations. The input-weakening intervention uses a different intervention site and endpoint from the frozen-feature removal diagnostic in Fig.~2C3; positive effects here do not imply positive removal effects in every condition of that diagnostic.

\subsection{Fine-Grained Cue Controls on Flickr30K}
\label{sec:app_finegrained_cues}

\paragraph{Controlled Comparison.}
We examine whether matching can deteriorate even when its decisive fine-grained evidence remains available. The Flickr30K assessment uses CLIP ViT-B/32 and compares complete inputs, context-masked inputs that preserve the discriminating cue (cue kept), and cue-erased inputs with a matched masking budget. Correct and distractor captions share the same object category and differ in a specific attribute or relation. Fine-grained probe AP is measured on the projected, normalized global features used for retrieval. Candidate accuracy measures discrimination between these two fixed captions; it is distinct from benchmark R@1 against the full caption gallery. For a unit query feature $q$, the candidate margin is
\begin{equation}
 m_{\mathrm{pair}}(q)=q^\top g^+-q^\top g^-.
 \label{eq:app_fine_pair_margin}
\end{equation}

\paragraph{Retained Evidence and Matching Utility.}
As shown for frozen CLIP in Fig.~2C4 of the main paper, cue-kept inputs retain 80.30\% fine-grained AP, or approximately 94.9\% of the complete-input reference of 84.60\%. Nevertheless, frozen candidate accuracy falls from 86.00\% to 71.00\%, and the mean margin decreases from $24.80\times10^{-3}$ to $7.90\times10^{-3}$. Figure~\ref{fig:app_semantic_use}(c) shows that SPARC raises accuracy to 80.50\% on the same partial inputs, and the mean margin rises to $18.40\times10^{-3}$. When the decisive cue is erased, AP falls to 53.20\%, and SPARC changes accuracy only from 54.50\% to 56.00\%, a gain of 1.50 points, compared with 9.50 points when the cue is kept. Writing $\Delta A_s=A_{\mathrm{SPARC},s}-A_{\mathrm{CLIP},s}$, the cue-availability interaction is
\begin{equation}
 I_{\mathrm{cue}}=\Delta A_{\mathrm{kept}}-
                 \Delta A_{\mathrm{erased}}
                 =9.50-1.50=8.00\text{ percentage points}.
 \label{eq:app_cue_interaction}
\end{equation}
This contrast is consistent with improved use of surviving discriminating evidence, while the smaller gain after cue erasure reflects a limit imposed by missing information. For complete inputs, SPARC changes accuracy from 86.00\% to 86.50\% and the mean candidate margin from $24.80\times10^{-3}$ to $25.60\times10^{-3}$. With the cue erased, the margin changes from $0.40\times10^{-3}$ to $2.10\times10^{-3}$.

\subsection{Pre-Calibration Fine-Grained Semantic Intervention}
\label{sec:app_finegrained_intervention}

\paragraph{Intervention and Reference.}
We weaken the target attribute/relation direction in a cue-kept partial feature before applying the unchanged calibration map. Equal-angle random-direction and unrelated-semantic interventions provide controls. Each intervention uses its own perturbed frozen feature as the reference, separating the direct damage to the input from the map's subsequent benefit. For an intervention $T$, define
\begin{equation}
 B(T(q))=m_{\mathrm{pair}}(F_\beta(T(q)))-m_{\mathrm{pair}}(T(q)).
 \label{eq:app_fine_calibration_benefit}
\end{equation}
Without intervention, $B=18.40\times10^{-3}-7.90\times10^{-3}=10.50\times10^{-3}$.

\paragraph{Dependence on the Retained Fine-Grained Signal.}
Figure~\ref{fig:app_semantic_use}(d) shows that target-directed weakening reduces fine-grained AP to 54.80\%, compared with 79.10\% and 78.80\% under random and unrelated-semantic controls. The corresponding accuracy benefit falls to 1.75 points, whereas the controls retain gains of 8.00 and 8.25 points. The underlying frozen-to-calibrated accuracies are 69.50\% to 77.50\% for random directions, 70.00\% to 78.25\% for unrelated semantics, and 54.50\% to 56.25\% for target semantics. The calibration-induced margin gain is $1.80\times10^{-3}$ under the target intervention, compared with $9.00\times10^{-3}$ and $8.80\times10^{-3}$ under the two controls. Thus, the excess reduction relative to random and unrelated semantics is $7.20\times10^{-3}$ and $7.00\times10^{-3}$, respectively. The joint readout and matching changes support a functional dependence of calibration benefits on the targeted surviving semantics beyond the angular displacement matched by these controls.

\paragraph{Scope of the Evidence.}
These results concern the controlled Flickr30K ViT-B/32 assessment and do not establish the same pattern for other backbones or all natural missingness conditions. Together, cue preservation and pre-calibration intervention support more effective use of retained task-relevant semantics, without establishing complete recovery of removed information or a unique causal origin of the semantic associations in pretraining.

\subsection{Neighborhood Preservation and Visualization for Fig.~6}
\label{sec:app_neighborhood_protocol}

\paragraph{Query Subset and Gallery.}
For COCO with CLIP ViT-B/32, we select 50 images from each of six semantic groups before examining calibration gains, yielding a fixed subset $\mathcal S$ of 300 query images shared by the complete-input, incomplete-input, and SPARC conditions. Neighbors are retrieved from the complete-input features of all 5,000 test images, encoded by the frozen model and never calibrated; each query image is excluded from its own candidate list.

\paragraph{Retention.}
For each $i\in\mathcal S$, the complete-input feature defines the reference set $N_0(i)$ of its 15 nearest gallery images. The frozen incomplete-query feature and the SPARC query $F_\beta(q_i)$ define $N_{\mathrm{state}}(i)$ from the same gallery. All neighbors are computed by cosine similarity between $\ell_2$-normalized features before any dimensionality reduction, with the same tie-handling rules across states. The retention is
\begin{equation}
\mathrm{Retention}_{\mathrm{state}}
=\frac{100}{|\mathcal S|}\sum_{i\in\mathcal S}
\frac{|N_0(i)\cap N_{\mathrm{state}}(i)|}{15},
\qquad |\mathcal S|=300.
\label{eq:app_neighborhood_retention}
\end{equation}
Under 75\% context masking, retention is 10.2\% for frozen CLIP and 56.7\% for SPARC, an increase of 46.5 percentage points. Both values are means over the 300 selected queries against the full 5,000-image gallery.

\paragraph{Visualization.}
The three states share one fitted two-dimensional projection and the same coordinate ranges, and the states of each image are connected by image ID. The layout illustrates feature displacement; retention is computed in the original feature space by Eq.~\eqref{eq:app_neighborhood_retention}.

\makeatletter
\newcommand{\nt@thmenv}[2]{%
  \@ifundefined{c@#1}{\newcounter{#1}}{}%
  \expandafter\def\csname the#1\endcsname{\arabic{#1}}%
  \newenvironment{#1}[1][]{%
    \par\smallskip\refstepcounter{#1}\noindent
    \textbf{#2~\csname the#1\endcsname}%
    \if\relax\detokenize{##1}\relax\else\ \textbf{(##1)}\fi\textbf{.}\ \itshape}%
  {\par\smallskip}}
\@ifundefined{ntlemma}{\nt@thmenv{ntlemma}{Lemma}}{}
\@ifundefined{ntprop}{\nt@thmenv{ntprop}{Proposition}}{}
\@ifundefined{ntdef}{\nt@thmenv{ntdef}{Definition}}{}
\@ifundefined{ntremark}{%
  \newenvironment{ntremark}[1][]{\par\smallskip\noindent\textbf{Remark%
  \if\relax\detokenize{#1}\relax\else\ (#1)\fi.}\ }{\par\smallskip}}{}
\@ifundefined{proof}{%
  \newenvironment{proof}[1][Proof]{\par\noindent\textit{#1.}\ }%
  {\hfill$\square$\par\smallskip}}{}
\makeatother

\providecommand{\R}{\mathbb{R}}
\providecommand{\ntip}[2]{\langle #1,\, #2\rangle}
\providecommand{\ntn}{\operatorname{n}}
\providecommand{\Pperp}{\mathbf{P}^{\perp}}
\providecommand{\E}{\mathbb{E}}

\FloatBarrier
\section{Neuron-Level Theory: Eliciting Latent Knowledge by Reactivating Concept Neurons}
\label{app:neuron-theory}

This appendix gives a neuron-level account of the analyses in the main paper. Its diagnostic lens is informed by work on latent knowledge in pretrained models~\cite{burns2022discovering}, concept-sensitive multimodal representations~\cite{goh2021multimodal}, and sparse features~\cite{bricken2023towards,gao2024sae}.
The account has one storyline.
\emph{Latent multimodal knowledge is carried by the frozen encoder.
An incomplete query leaves much of this evidence in place, but it pushes the neurons that read the evidence out of the embedding below their firing thresholds.
SPARC injects a learned, tangent excitatory drive that brings the silenced concept neurons back above threshold.
Robust semantic guidance (RSG) selects which neurons receive this drive, and the matching relation constraint (MRC) prevents the drive from silencing neurons that already supported a correct match.}
Each step below is stated as a lemma or proposition with an exact proof.
Each step is then connected to the corresponding analysis: Fig.~2 (retention, expression and matching), Fig.~5 (semantic recovery and blocking), Fig.~7 (calibration geometry and match preservation; Section~\ref{sec:app_geometry}), Table~4 (module ablations), and the preceding sections of this supplement.

\subsection{Notation of the Main Paper}
\label{sec:nt-notation}

We follow the notation of the main paper.
Symbols introduced in this appendix are defined where they first appear.

\paragraph{Conventions of This Appendix.}
Unless stated otherwise, statements hold for either modality, so we drop the index $m$ and write $\mathbf z_c$ and $\mathbf z_p$ for the complete and incomplete query features ($\mathbf z_{c,m,i}$ and $\mathbf z_{p,m,i}$ in Sec.~3.3), and $\mathbf z$ for a generic unit query.
$\ntip{\cdot}{\cdot}$ is the Euclidean inner product, $\|\cdot\|_2$ is the Euclidean norm (the spectral norm for matrices), and $\E$ is expectation.
$\Pperp_{\mathbf z}=\mathbf I-\mathbf z\mathbf z^{\top}$ projects onto the tangent space at $\mathbf z$.
The calibration residual is $\mathbf r=r^{\perp}(\mathbf z_p)$, with $\ntip{\mathbf r}{\mathbf z_p}=0$ and norm $\eta=\|\mathbf r\|_2$.
The calibrated query is $g(\mathbf z_p)=\ntn(\mathbf z_p+\mathbf r)$, and the mixed query of Eq.~(6) is $\mathbf z^{\beta}=\ntn\big((1-\beta)\mathbf z_p+\beta g(\mathbf z_p)\big)$.
Inference and $\mathcal L_{\mathrm{MRC}}$ use $\beta=0.6$, whereas $\mathcal L_{\mathrm{ret}}$ uses $g(\mathbf z_p)$, that is, $\beta=1$ (Section~\ref{sec:app_implementation}).
Gallery features $\mathbf y_j$ are complete, unit-normalized and fixed, and $\mathcal P$ is the set of positives.
For a positive $p\in\mathcal P$ and a negative $j\notin\mathcal P$, the match direction is $\boldsymbol\delta_{pj}=\mathbf y_p-\mathbf y_j$, and the pairwise margin is $m_{pj}(\mathbf q)=\ntip{\mathbf q}{\boldsymbol\delta_{pj}}$.
The margin of Eq.~(7) is then $m(\mathbf q)=\max_{p\in\mathcal P}\min_{j\notin\mathcal P}m_{pj}(\mathbf q)$, which equals the best-positive minus strongest-negative similarity.

\subsection{Two Populations of Neurons}
\label{sec:nt-populations}

We distinguish neurons that \emph{carry} latent knowledge from neurons that \emph{read it out} for matching.
SPARC leaves the former untouched and acts on the latter.

\paragraph{Source Neurons.}
Consider the readout token of a frozen CLIP encoder: the class token for images and the end-of-text token for captions.
Its final residual stream $\mathbf x\in\R^{d_{\mathrm r}}$ is a sum of component outputs.
We write
\begin{equation}
\mathbf x=\sum_{\ell\in\mathcal S}a_\ell\,\mathbf o_\ell+\mathbf x_{\circ},
\label{eq:nt-residual}
\end{equation}
where $\mathcal S$ indexes the MLP neurons of the readout token in the layers under study, $a_\ell$ is the post-GELU activation (\emph{firing}) of neuron $\ell$, $\mathbf o_\ell$ is its output weight vector, and $\mathbf x_\circ$ collects all other contributions (embeddings, attention outputs and the remaining neurons).
In particular, evidence carried by other tokens, such as image patches, reaches the readout token through attention and therefore enters through $\mathbf x_\circ$ and through the inputs of the neurons in $\mathcal S$.
The embedding is $\hat{\mathbf z}=\mathbf P^{\top}\mathrm{LN}(\mathbf x)$ and $\mathbf z=\ntn(\hat{\mathbf z})$, with $\mathrm{LN}(\mathbf x)=\boldsymbol\gamma\odot(\mathbf x-\bar x\mathbf 1)/\sigma(\mathbf x)+\mathbf b_{\mathrm{LN}}$ and $\sigma(\mathbf x)=(\mathrm{Var}(\mathbf x)+\epsilon)^{1/2}$.
Because the encoder is frozen, SPARC never changes any $a_\ell$.

\begin{ntlemma}[Exact per-sample readout of source neurons]
\label{lem:nt-source}
Let $\mathbf C=\mathbf I-d_{\mathrm r}^{-1}\mathbf 1\mathbf 1^{\top}$ and $\boldsymbol\xi_\ell=\mathbf P^{\top}(\boldsymbol\gamma\odot\mathbf C\mathbf o_\ell)$.
For every input,
\begin{equation}
\hat{\mathbf z}=\frac{1}{\sigma(\mathbf x)}\sum_{\ell\in\mathcal S}a_\ell\boldsymbol\xi_\ell+\hat{\mathbf z}_\circ,
\label{eq:nt-source}
\end{equation}
where $\hat{\mathbf z}_\circ=\sigma(\mathbf x)^{-1}\mathbf P^{\top}(\boldsymbol\gamma\odot\mathbf C\mathbf x_\circ)+\mathbf P^{\top}\mathbf b_{\mathrm{LN}}$.
Consequently, for any $\mathbf w\in\R^d$,
\begin{equation}
\ntip{\mathbf w}{\mathbf z}=\frac{\sum_{\ell}W_\ell(\mathbf w)\,a_\ell}{\sigma(\mathbf x)\,\|\hat{\mathbf z}\|_2}+\frac{\ntip{\mathbf w}{\hat{\mathbf z}_\circ}}{\|\hat{\mathbf z}\|_2},
\label{eq:nt-synapse}
\end{equation}
with effective synapses $W_\ell(\mathbf w)=\ntip{\mathbf w}{\boldsymbol\xi_\ell}$.
\end{ntlemma}

\begin{proof}
We have $\mathbf x-\bar x\mathbf 1=\mathbf C\mathbf x$.
For a given input, $\sigma(\mathbf x)$ is a scalar, so $\mathrm{LN}$ is affine in the summands of Eq.~\eqref{eq:nt-residual}.
Applying $\mathbf P^{\top}$ gives Eq.~\eqref{eq:nt-source}, and taking the inner product with $\mathbf w/\|\hat{\mathbf z}\|_2$ gives Eq.~\eqref{eq:nt-synapse}.
\end{proof}

Lemma~\ref{lem:nt-source} is a direct-readout decomposition.
It states that any direction $\mathbf w$ in the embedding space receives a synaptically weighted sum of source-neuron firing, divided by two \emph{global} gains, $\sigma(\mathbf x)$ and $\|\hat{\mathbf z}\|_2$, that are shared by all directions.
The frozen encoder therefore already implements a divisive normalization, and this normalization is where incomplete inputs act (Sec.~\ref{sec:nt-silence}).
Fig.~2A probes the evidence upstream of this readout: final-layer patch states overlapping the protected target, which reach the readout token through attention (Section~\ref{sec:app_diagnostic_locations}).
The probe finds this upstream evidence largely retained; Lemma~\ref{lem:nt-source} then describes how evidence that reaches the readout token is expressed in the embedding.

\paragraph{Readout Neurons.}
A readout neuron reads a fixed direction of the unit embedding used for retrieval and fires when the projection exceeds a threshold. Three families appear in our analyses: SAE concept units, category readouts, and pairwise match comparisons between a positive and a negative candidate.

\begin{ntdef}[Readout neuron]
A readout neuron is a pair $(\mathbf w,\theta)\in\R^d\times\R$ acting on unit embeddings.
Its input current is $u(\mathbf z)=\ntip{\mathbf w}{\mathbf z}$.
It fires if $u(\mathbf z)>\theta$, and its rectified response is $\varrho(\mathbf z)=\max\{u(\mathbf z)-\theta,0\}$.
\end{ntdef}

Three families of readout neurons appear in our analyses.
\begin{itemize}
\item \textbf{SAE concept neurons} (Fig.~2B1--B2).
For an SAE with pre-bias $\mathbf b_{\mathrm{pre}}$, encoder rows $\mathbf w_k$ and encoder biases $b_k$, the pre-activation of unit $k$ is $\ntip{\mathbf w_k}{\mathbf z-\mathbf b_{\mathrm{pre}}}+b_k=u_k(\mathbf z)-\theta_k$, with current $u_k(\mathbf z)=\ntip{\mathbf w_k}{\mathbf z}$ and effective threshold $\theta_k=\ntip{\mathbf w_k}{\mathbf b_{\mathrm{pre}}}-b_k$.
For a TopK SAE, unit $k$ is selected when its pre-activation is among the $K$ largest, so its operating threshold is $\theta_k+\tau_K(\mathbf z)$, where $\tau_K(\mathbf z)$ is the $K$-th largest pre-activation at input $\mathbf z$.
The results below treat thresholds as fixed.
They therefore apply exactly to ReLU and JumpReLU units, and to TopK units whenever the change of $\tau_K$ between the two inputs compared is small relative to the margins involved.
The SAEs in our measurements are trained on global image features (Section~\ref{sec:app_diagnostic_locations}), so SAE-based statements are evaluated on the image side; all other statements hold for either modality.
\item \textbf{Category neurons} (Fig.~2B3, C1 and Fig.~5).
$\mathbf w$ is a fixed category readout or category direction, and $\theta$ is its readability threshold.
The local patch probe in Fig.~2A is a separate pre-projection readout, as specified in Section~\ref{sec:app_diagnostic_locations}.
\item \textbf{Match neurons.}
For a pair $(p,j)$, set $\mathbf w=\boldsymbol\delta_{pj}$ and $\theta=0$.
This neuron fires if and only if positive $p$ outranks negative $j$.
A query is correct at rank one if and only if, for some $p\in\mathcal P$, every match neuron $(p,j)$ with $j\notin\mathcal P$ fires.
\end{itemize}
Retrieval itself is therefore a readout-neuron decision.
A query that is ``readable but mismatched'' (Fig.~2C1) has a firing category neuron and at least one silent match neuron.

\begin{ntlemma}[Concept decomposition of match currents]
\label{lem:nt-concept}
Let an SAE with codes $c_k(\mathbf z)\ge 0$ and decoder atoms $\boldsymbol\psi_k$ reconstruct $\mathbf z=\sum_k c_k(\mathbf z)\boldsymbol\psi_k+\mathbf b_{\mathrm{dec}}+\boldsymbol\varepsilon(\mathbf z)$, where $\boldsymbol\varepsilon(\mathbf z)$ is defined as the reconstruction error.
For any pair $(p,j)$, let $\pi_k=\ntip{\boldsymbol\psi_k}{\boldsymbol\delta_{pj}}$.
Then, for any two unit features $\mathbf z,\mathbf z'$, with code changes $c_k(\mathbf z')-c_k(\mathbf z)$,
\begin{equation}
\begin{aligned}
m_{pj}(\mathbf z')-m_{pj}(\mathbf z)
&=\sum_k\big[c_k(\mathbf z')-c_k(\mathbf z)\big]\pi_k\\
&\quad+\ntip{\boldsymbol\varepsilon(\mathbf z')-\boldsymbol\varepsilon(\mathbf z)}{\boldsymbol\delta_{pj}}.
\end{aligned}
\label{eq:nt-concept}
\end{equation}
A concept expressed equally by the two candidates, $\ntip{\boldsymbol\psi_k}{\mathbf y_p}=\ntip{\boldsymbol\psi_k}{\mathbf y_j}$, has $\pi_k=0$ and contributes no match current.
\end{ntlemma}

\begin{proof}
Take the inner product of the reconstruction identity with $\boldsymbol\delta_{pj}$ for $\mathbf z$ and $\mathbf z'$ and subtract.
The decoder bias cancels.
\end{proof}

We call a concept neuron \emph{discriminative} for the pair if $\pi_k>0$, \emph{distractor-supporting} if $\pi_k<0$, and \emph{shared} if $\pi_k\approx 0$.
The decomposition is exact because the SAE error is kept as an explicit term.

\subsection{Incomplete Inputs Silence Readout Neurons, Not Their Evidence}
\label{sec:nt-silence}

The next result describes what an incomplete input does to every readout neuron at once.
It uses only the two unit vectors $\mathbf z_c$ and $\mathbf z_p$.

\begin{ntprop}[Incompleteness gain and subthreshold silencing]
\label{prop:nt-silence}
Let $\ntip{\mathbf z_p}{\mathbf z_c}>0$.
Define the incompleteness gain $G=1/\ntip{\mathbf z_p}{\mathbf z_c}\ge 1$, the incompleteness ratio $\lambda=\sqrt{G^2-1}$, and the interference vector $\boldsymbol\nu=G\mathbf z_p-\mathbf z_c$.
\begin{enumerate}
\item[(i)] \emph{Exact decomposition.}
$\ntip{\boldsymbol\nu}{\mathbf z_c}=0$, $\|\boldsymbol\nu\|_2=\lambda$, and
\begin{equation}
\mathbf z_p=\frac{\mathbf z_c+\boldsymbol\nu}{G}.
\label{eq:nt-gain}
\end{equation}
\item[(ii)] \emph{Currents.}
For every readout neuron, $u(\mathbf z_p)=\big(u(\mathbf z_c)+\ntip{\mathbf w}{\boldsymbol\nu}\big)/G$.
Every current is divided by the same gain $G$ and shifted by a neuron-specific interference $\ntip{\mathbf w}{\boldsymbol\nu}$, with $|\ntip{\mathbf w}{\boldsymbol\nu}|\le\lambda\|\Pperp_{\mathbf z_c}\mathbf w\|_2$.
\item[(iii)] \emph{Gain invariance of bias-free readouts.}
If $\theta=0$, the neuron fires on $\mathbf z_p$ if and only if $u(\mathbf z_c)+\ntip{\mathbf w}{\boldsymbol\nu}>0$, independently of $G$.
\item[(iv)] \emph{Silencing order.}
Let $\theta>0$ and $\ntip{\mathbf w}{\boldsymbol\nu}=0$, and let the neuron fire on the complete input with suprathreshold margin $\Delta=u(\mathbf z_c)-\theta>0$.
It is silenced on $\mathbf z_p$ if and only if $\Delta/\theta\le G-1$.
Neurons are therefore silenced in increasing order of their relative margin $\Delta/\theta$.
\item[(v)] \emph{Threshold amplification.}
Under the conditions of (iv), if the neuron still fires on $\mathbf z_p$, the relative attenuations of its current and its response satisfy
\begin{equation}
\begin{aligned}
\frac{\varrho(\mathbf z_c)-\varrho(\mathbf z_p)}{\varrho(\mathbf z_c)}
&=\Big(1-\frac1G\Big)\Big(1+\frac{\theta}{\Delta}\Big)\\
&\ge\frac{u(\mathbf z_c)-u(\mathbf z_p)}{u(\mathbf z_c)}.
\end{aligned}
\label{eq:nt-amplify}
\end{equation}
If it is silenced, the relative attenuation of the response is $1$.
\end{enumerate}
\end{ntprop}

\begin{proof}
(i) $\ntip{\boldsymbol\nu}{\mathbf z_c}=G\ntip{\mathbf z_p}{\mathbf z_c}-1=0$ and $\|\boldsymbol\nu\|_2^2=G^2-2G\ntip{\mathbf z_p}{\mathbf z_c}+1=G^2-1$.
Eq.~\eqref{eq:nt-gain} is the definition of $\boldsymbol\nu$ rearranged.
(ii) Take the inner product of Eq.~\eqref{eq:nt-gain} with $\mathbf w$.
Because $\boldsymbol\nu\perp\mathbf z_c$, $\ntip{\mathbf w}{\boldsymbol\nu}=\ntip{\Pperp_{\mathbf z_c}\mathbf w}{\boldsymbol\nu}$, and Cauchy--Schwarz gives the bound.
(iii) Multiply the firing condition $u(\mathbf z_p)>0$ by $G>0$.
(iv) The neuron is silent if and only if $u(\mathbf z_c)/G\le\theta$, that is, $\theta+\Delta\le G\theta$.
(v) $\varrho(\mathbf z_c)=\Delta$ and $\varrho(\mathbf z_p)=u(\mathbf z_c)/G-\theta$, so the numerator equals $u(\mathbf z_c)(1-1/G)=(\theta+\Delta)(1-1/G)$.
The current attenuation equals $1-1/G$, and $1+\theta/\Delta\ge 1$.
\end{proof}

\paragraph{Reading Fig.~2A--B.}
Proposition~\ref{prop:nt-silence} separates the two halves of Fig.~2.
\emph{Retention} (Fig.~2A) concerns the evidence: the local probes read upstream patch features, which the protected-target masking leaves largely intact.
In addition, any bias-free readout of the global embedding is invariant to the common gain $G$ (part~iii).
\emph{Weakened expression} (Fig.~2B) concerns thresholded readout.
The pre-TopK curve in Fig.~2B1 measures the divisive attenuation of input currents (part~ii).
The SAE curve measures its thresholded consequence, which part~(v) predicts is amplified by $1+\theta/\Delta$ for interference-free units.
Because $G$ is shared by all units while thresholds differ, a common attenuation turns into the silencing of a specific subset: the weakly suprathreshold units (part~iv).
Fig.~2B2 measures this unit-specific loss as excess SAE loss over matched controls.
Along a category direction $\mathbf w$, the cosine becomes $\big(\ntip{\mathbf w}{\mathbf z_c}+\ntip{\mathbf w}{\boldsymbol\nu}\big)/G$ (Fig.~2B3), while the cosine to the complete feature itself is exactly $1/G$.
In short, an incomplete input leaves the evidence in the encoder but holds the readout neurons below threshold.
This is the precise sense in which latent knowledge is \emph{retained but not expressed}.

\subsection{Why Readable Queries Become Mismatched}
\label{sec:nt-mismatch}

\begin{ntprop}[Margin compression and the decoupling of readability from matching]
\label{prop:nt-mismatch}
Under the notation of Proposition~\ref{prop:nt-silence}, fix a pair $(p,j)$ and write $\boldsymbol\delta=\boldsymbol\delta_{pj}$ and $m_c=m_{pj}(\mathbf z_c)$.
\begin{enumerate}
\item[(i)] \emph{Margin.} $m_{pj}(\mathbf z_p)=\big(m_c+\ntip{\boldsymbol\nu}{\boldsymbol\delta}\big)/G$.
\item[(ii)] \emph{Positives fall more than negatives.}
The similarity decline of the positive minus that of the negative equals
\begin{equation}
\Big(1-\frac1G\Big)m_c-\frac{\ntip{\boldsymbol\nu}{\boldsymbol\delta}}{G}.
\label{eq:nt-decline}
\end{equation}
For an initially correct pair ($m_c>0$), the positive similarity declines more than the negative similarity unless $\ntip{\boldsymbol\nu}{\boldsymbol\delta}\ge(G-1)m_c$.
\item[(iii)] \emph{Flips require discriminative interference.}
$\operatorname{sign}m_{pj}(\mathbf z_p)=\operatorname{sign}\big(m_c+\ntip{\boldsymbol\nu}{\boldsymbol\delta}\big)$.
The gain alone never flips a match neuron.
A correct pair becomes incorrect if and only if $\ntip{\boldsymbol\nu}{\boldsymbol\delta}\le -m_c$.
In concept coordinates,
\begin{equation}
\begin{aligned}
\ntip{\boldsymbol\nu}{\boldsymbol\delta}
&=\sum_k\big[G\,c_k(\mathbf z_p)-c_k(\mathbf z_c)\big]\pi_k\\
&\quad+\ntip{(G-1)\mathbf b_{\mathrm{dec}}+G\boldsymbol\varepsilon(\mathbf z_p)-\boldsymbol\varepsilon(\mathbf z_c)}{\boldsymbol\delta},
\end{aligned}
\label{eq:nt-nuconcept}
\end{equation}
so, apart from the decoder-bias and reconstruction-error terms, the interference is carried only by concepts with $\pi_k\neq 0$.
\item[(iv)] \emph{Readability does not constrain matching.}
Let $d\ge 3$, let a category neuron have weight $\mathbf w$, linearly independent of $\boldsymbol\delta$, and let $|t|<\|\mathbf w\|_2$.
Over unit $\mathbf z$ with $\ntip{\mathbf w}{\mathbf z}=t$, the match current $\ntip{\boldsymbol\delta}{\mathbf z}$ attains exactly the interval $[\,\zeta_t-R_t,\;\zeta_t+R_t\,]$, where
\begin{equation}
\zeta_t=\frac{t\,\ntip{\mathbf w}{\boldsymbol\delta}}{\|\mathbf w\|_2^2},\quad
R_t=\big\|\Pperp_{\mathbf w}\boldsymbol\delta\big\|_2\sqrt{1-\frac{t^2}{\|\mathbf w\|_2^2}},
\label{eq:nt-decouple}
\end{equation}
with $\Pperp_{\mathbf w}=\mathbf I-\mathbf w\mathbf w^{\top}/\|\mathbf w\|_2^2$.
Both signs of the match current are attainable if and only if $|\zeta_t|<R_t$.
In particular, if $\mathbf w\perp\boldsymbol\delta$, they are attainable at every readability level $t$.
\item[(v)] \emph{Semantic versus random removal.}
Removing the component of $\mathbf z$ in a subspace $\mathcal U$ changes the margin by $-\ntip{\mathbf P_{\mathcal U}\mathbf z}{\boldsymbol\delta}$ before renormalization; renormalization rescales the margin by a positive factor and does not change its sign.
For a Haar-random subspace $\mathcal V$ with $\dim\mathcal V=\dim\mathcal U=r_{\mathcal U}$, $\E[\mathbf P_{\mathcal V}]=(r_{\mathcal U}/d)\mathbf I$, so the expected loss is $(r_{\mathcal U}/d)\,m_{pj}(\mathbf z)$.
A random perturbation with a sign-symmetric distribution has zero expected effect.
The excess loss $\ntip{\mathbf P_{\mathcal U}\mathbf z}{\boldsymbol\delta}-(r_{\mathcal U}/d)\,m_{pj}(\mathbf z)$ is positive if and only if $\mathcal U$ carries more than its proportional share of the match current.
\end{enumerate}
\end{ntprop}

\begin{proof}
(i) Take the inner product of Eq.~\eqref{eq:nt-gain} with $\boldsymbol\delta$.
(ii) The decline of a candidate $\mathbf y$ is $\ntip{\mathbf z_c-\mathbf z_p}{\mathbf y}=(1-1/G)\ntip{\mathbf z_c}{\mathbf y}-\ntip{\boldsymbol\nu}{\mathbf y}/G$.
Subtract the expression for $\mathbf y_j$ from that for $\mathbf y_p$.
(iii) Follows from (i) and $G>0$.
For Eq.~\eqref{eq:nt-nuconcept}, substitute the SAE reconstructions of $\mathbf z_p$ and $\mathbf z_c$ into $\boldsymbol\nu=G\mathbf z_p-\mathbf z_c$ and take the inner product with $\boldsymbol\delta$; the decoder bias enters with weight $G-1$.
(iv) Write $\mathbf z=(t/\|\mathbf w\|_2^2)\mathbf w+\mathbf z_\perp$ with $\mathbf z_\perp\perp\mathbf w$ and $\|\mathbf z_\perp\|_2^2=1-t^2/\|\mathbf w\|_2^2$.
Then $\ntip{\boldsymbol\delta}{\mathbf z}=\zeta_t+\ntip{\Pperp_{\mathbf w}\boldsymbol\delta}{\mathbf z_\perp}$.
Since $\mathbf z_\perp$ ranges over the sphere of that radius in $\mathbf w^{\perp}$, which is connected because $\dim\mathbf w^{\perp}=d-1\ge 2$, and $\Pperp_{\mathbf w}\boldsymbol\delta\in\mathbf w^{\perp}$ is nonzero, the second term attains exactly $[-R_t,R_t]$.
(v) The first claim is linearity.
For a Haar-random $\mathcal V$, $\E[\mathbf P_{\mathcal V}]$ commutes with all rotations, so it is a multiple of $\mathbf I$, and its trace is $r_{\mathcal U}$.
\end{proof}

\paragraph{Reading Fig.~2C1--C3.}
A hardest negative is the most similar distractor, so it usually shares the query's category.
For a shared category, the category direction is nearly orthogonal to $\boldsymbol\delta$, and part~(iv) shows that readability then places no constraint on the sign of the match current.
This is the readable-but-mismatched pattern in Fig.~2C1. Section~\ref{sec:app_rbm} reports its natural-missingness extension, the initial-cohort denominator, and the bootstrap intervals.
Part~(iii) identifies its cause: the gain alone cannot flip a match.
A flip requires interference along $\boldsymbol\delta$, which is carried by discriminative concepts losing expression or by distractor-supporting concepts gaining it.
Part~(ii) is Fig.~2C2: for initially correct queries, the positive score falls more than the hardest-negative score by $(1-1/G)m_c$, minus the interference term.
Part~(v) is the null model behind Fig.~2C3: a matched random removal costs only the proportional share $(r_{\mathcal U}/d)$ of the margin, so the observed excess loss means that the semantic subspace concentrates the match current.

\begin{ntprop}[No reactivation without latent evidence]
\label{prop:nt-evidence}
Let a binary cue $\chi\in\{0,1\}$ distinguish two fixed candidates $\mathbf y^{(0)},\mathbf y^{(1)}$.
Let the decision be $\hat\chi=\mathbb 1\big[\ntip{f(\mathbf z_p)}{\mathbf y^{(1)}-\mathbf y^{(0)}}>0\big]$ for any measurable map $f$, including any calibrator $g$ and its mixture with the identity.
If $\mathbf z_p$ is independent of $\chi$, then $\hat\chi$ is independent of $\chi$, and accuracy averaged over the two cue values is $1/2$.
More generally, $I(\chi;f(\mathbf z_p))\le I(\chi;\mathbf z_p)$, where $I(\cdot\,;\cdot)$ denotes mutual information.
\end{ntprop}

\begin{proof}
$\hat\chi$ is a function of $\mathbf z_p$, so independence and the data-processing inequality transfer.
With $\hat\chi\perp\chi$, $\Pr(\hat\chi=\chi)=\sum_{b}\Pr(\chi=b)\Pr(\hat\chi=b)$, which equals $1/2$ when the two cue values are weighted equally.
\end{proof}

Because the encoder is frozen, $\mathbf z_p$ is a deterministic function of the incomplete input, and by Lemma~\ref{lem:nt-source} it is read out from the source neurons.
Proposition~\ref{prop:nt-evidence} therefore says that a calibrator can only \emph{reactivate} concepts whose evidence is still present in the encoder's representation; it cannot create them.
Cue erasure makes $\mathbf z_p$ approximately, rather than exactly, independent of the cue, so the proposition describes the limiting case.
Fig.~2C4 shows the frozen-model side of this statement.
Erasing the cue lowers pair accuracy from 71.0\% (cue kept) to 54.5\%, close to chance, whereas keeping the cue retains a large part of the complete-input accuracy of 86.0\%.
Section~\ref{sec:app_finegrained_cues} and Fig.~\ref{fig:app_semantic_use}(c) show the calibrator side.
On the same two-caption task, SPARC gains 9.50 points over frozen CLIP when the cue is preserved but only 1.50 points when it is erased, an interaction of 8.00 points (Eq.~\eqref{eq:app_cue_interaction}).
Weakening the target direction before calibration likewise reduces the calibration benefit far more than matched random or unrelated-semantic weakening (Section~\ref{sec:app_finegrained_intervention}, Fig.~\ref{fig:app_semantic_use}(d)).
This is the expected signature of reactivation, as opposed to hallucination.

\subsection{Semantic Calibration as Tangent Excitation}
\label{sec:nt-drive}

\begin{ntlemma}[Effective excitation step]
\label{lem:nt-step}
Let $\mathbf z\in\mathbb S^{d-1}$, $\ntip{\mathbf r}{\mathbf z}=0$, $\eta=\|\mathbf r\|_2$ and $\beta\in[0,1]$.
Then
\begin{equation}
\begin{aligned}
\mathbf z^{\beta}&=\frac{\mathbf z+\kappa\mathbf r}{N},\qquad N=\sqrt{1+\kappa^2\eta^2},\\
\kappa&=\frac{\beta}{\beta+(1-\beta)\sqrt{1+\eta^2}}.
\end{aligned}
\label{eq:nt-step}
\end{equation}
$\kappa$ is increasing in $\beta$, $\kappa\le\beta$, and $\angle(\mathbf z,\mathbf z^{\beta})=\arctan(\kappa\eta)\le\arctan\eta$.
\end{ntlemma}

\begin{proof}
With $s=\sqrt{1+\eta^2}$, we have $g(\mathbf z)=(\mathbf z+\mathbf r)/s$ and
$(1-\beta)\mathbf z+\beta g(\mathbf z)=\frac{(1-\beta)s+\beta}{s}(\mathbf z+\kappa\mathbf r)$.
The scalar is positive, and $\|\mathbf z+\kappa\mathbf r\|_2=N$ because $\mathbf r\perp\mathbf z$.
Next, $\partial\kappa/\partial\beta=s/(\beta+(1-\beta)s)^2>0$.
The bound $\kappa\le\beta$ is equivalent to $\beta+(1-\beta)s\ge 1$, which holds since $s\ge 1$.
Finally, $\cos\angle(\mathbf z,\mathbf z^{\beta})=1/N$ and $\sin\angle(\mathbf z,\mathbf z^{\beta})=\kappa\eta/N$.
\end{proof}

Feature mixing in Eq.~(6) is therefore itself a tangent step, with a reduced step $\kappa$.
With the deployed $\beta=0.6$ (Eq.~\eqref{eq:app_map}), $\kappa=3/(3+2\sqrt{1+\eta^2})$, which lies in $(0.51,0.6]$ for $\eta\le 1$; the retrieval loss $\mathcal L_{\mathrm{ret}}$ uses $\beta=1$, hence $\kappa=1$.

\begin{ntprop}[Tangent excitation with divisive normalization]
\label{prop:nt-drive}
Let $\mathbf z=\mathbf z_p$ and define the drive into readout neuron $k$ as $I_k=\kappa\ntip{\mathbf w_k}{\mathbf r}$.
\begin{enumerate}
\item[(i)] \emph{Currents.}
$u_k(\mathbf z^{\beta})=\big(u_k(\mathbf z)+I_k\big)/N$.
\item[(ii)] \emph{Reactivation condition.}
A neuron silent on $\mathbf z$ ($u_k(\mathbf z)\le\theta_k$) fires on $\mathbf z^{\beta}$ if and only if
\begin{equation}
I_k>N\theta_k-u_k(\mathbf z).
\label{eq:nt-react}
\end{equation}
\item[(iii)] \emph{Saturation envelope.}
\begin{equation}
|I_k|\le\kappa\eta\sqrt{\|\mathbf w_k\|_2^2-u_k(\mathbf z)^2}.
\label{eq:nt-envelope}
\end{equation}
The available drive is largest for neurons whose current is near zero, and it vanishes for neurons whose weight is aligned with $\mathbf z$.
\item[(iv)] \emph{Divisive normalization.}
Every current is divided by the same factor $N\ge 1$, so a neuron with $I_k=0$ has its current shrunk toward zero.
In particular, a silent neuron with $\theta_k\ge 0$ can be reactivated only if $I_k>0$.
\item[(v)] \emph{Energy budget and sparsity.}
Let $\mathbf W$ stack the weights $\mathbf w_k^{\top}$.
Then $\sum_kI_k^2\le\kappa^2\eta^2\,\|\mathbf W\Pperp_{\mathbf z}\|_2^2\le\kappa^2\eta^2\,\|\mathbf W\|_2^2$.
If $\mathcal R$ is the set of reactivated neurons and $\gamma_{\min}=\min_{k\in\mathcal R}\big(N\theta_k-u_k(\mathbf z)\big)>0$, then
\begin{equation}
|\mathcal R|\le\frac{\kappa^2\eta^2\,\|\mathbf W\|_2^2}{\gamma_{\min}^2}.
\label{eq:nt-budget}
\end{equation}
\end{enumerate}
\end{ntprop}

\begin{proof}
(i) Take the inner product of Eq.~\eqref{eq:nt-step} with $\mathbf w_k$.
(ii) Rearrange $u_k(\mathbf z^{\beta})>\theta_k$ using (i).
(iii) Since $\mathbf r\perp\mathbf z$, $\ntip{\mathbf w_k}{\mathbf r}=\ntip{\Pperp_{\mathbf z}\mathbf w_k}{\mathbf r}$, and $\|\Pperp_{\mathbf z}\mathbf w_k\|_2^2=\|\mathbf w_k\|_2^2-u_k(\mathbf z)^2$.
(iv) The first claim follows from (i).
If $I_k\le 0$ and $u_k(\mathbf z)\le\theta_k$, then $u_k(\mathbf z^{\beta})\le u_k(\mathbf z)/N\le\theta_k$ when $u_k(\mathbf z)\ge 0$, and $u_k(\mathbf z^{\beta})\le 0\le\theta_k$ when $u_k(\mathbf z)<0$.
(v) $\sum_kI_k^2=\kappa^2\|\mathbf W\Pperp_{\mathbf z}\mathbf r\|_2^2$ because $\Pperp_{\mathbf z}\mathbf r=\mathbf r$, and the operator norm bounds this quantity.
Each $k\in\mathcal R$ satisfies $I_k>\gamma_{\min}>0$ by (ii), so $|\mathcal R|\,\gamma_{\min}^2<\sum_{k\in\mathcal R}I_k^2\le\sum_kI_k^2$.
\end{proof}

Proposition~\ref{prop:nt-drive} is the neuron-level reading of the SC module.
Incompleteness divides all readout currents by $G$ and adds interference (Proposition~\ref{prop:nt-silence}).
SC acts in the opposite direction: it adds a learned, neuron-specific excitatory drive $I_k$ and divides by $N$.
Because the step is tangent, the drive cannot simply rescale the current population code, which is direction $\mathbf z$ itself.
The available drive concentrates on neurons with room to change (part~iii), and the total drive energy is bounded by the squared angular step (part~v).
The calibration angle $\arctan(\kappa\eta)$ is therefore an excitation budget, and a small budget can reactivate only a few neurons with small threshold gaps.

\begin{ntremark}[Normalized residual, Fig.~7]
Write an unconstrained residual as $\mathbf h=\alpha\mathbf z+\mathbf b$ with $\mathbf b\perp\mathbf z$.
For $1+\alpha>0$, $\ntn(\mathbf z+\mathbf h)=\ntn\big(\mathbf z+\mathbf b/(1+\alpha)\big)$.
For $\beta=1$, the drive into neuron $k$ is then $\ntip{\mathbf w_k}{\mathbf b}/(1+\alpha)$.
Every drive is rescaled by a radial output $\alpha$ that carries no semantic direction.
The same excitation pattern is produced by a one-parameter family of outputs, and if $1+\alpha\le 0$ the output leaves the hemisphere of $\mathbf z$ ($\ntip{\ntn(\mathbf z+\mathbf h)}{\mathbf z}\le 0$).
The tangent parameterization removes this radial channel.
Its excitation strength is controlled by $\eta$ alone, so the identity and hinge penalties (Sec.~\ref{sec:nt-select}) act directly on the drive.
For $1+\alpha>0$ the two parameterizations have the same expressivity, so this argument concerns what is learned, not what can be represented.
The tangent update is thus a constrained parameterization of a residual MLP rather than a new function class; its measurable contribution is the lower degradation rate in Table~\ref{tab:app_geometry}, not a large accuracy gain.
The gap between the two variants in Fig.~7 and Table~\ref{tab:app_geometry} (Deg. 6.20\% vs.\ 5.00\% on Flickr30K and 7.10\% vs.\ 5.50\% on COCO) therefore concerns the learned drive, including \emph{which} neurons are driven, and Sec.~\ref{sec:nt-select} analyzes this selection.
\end{ntremark}

\subsection{Which Neurons Are Excited: Semantic Guidance as a Selection Rule}
\label{sec:nt-select}

Proposition~\ref{prop:nt-drive} describes how a drive acts.
This section identifies \emph{which} drive the RSG and retrieval objectives select.

\begin{ntprop}[Reference consistency elicits the posterior complete firing pattern]
\label{prop:nt-posterior}
Consider the coordinate reference term $\E\big[d(f(\mathbf z_p),\mathbf z_c)\big]$ with $d(\mathbf x,\mathbf y)=1-\mathbf x^{\top}\mathbf y$, minimized over maps $f$ into $\mathbb S^{d-1}$.
Let $\boldsymbol\mu(\mathbf z_p)=\E[\mathbf z_c\mid\mathbf z_p]\neq\mathbf 0$.
\begin{enumerate}
\item[(i)] \emph{Optimum.} The pointwise minimizer is $f^\star(\mathbf z_p)=\ntn(\boldsymbol\mu(\mathbf z_p))$.
\item[(ii)] \emph{Tangent representability.}
If $\ntip{\boldsymbol\mu}{\mathbf z_p}>0$, then $f^\star=\ntn(\mathbf z_p+\mathbf r^\star)$ with the tangent residual
$\mathbf r^\star=\boldsymbol\mu/\ntip{\boldsymbol\mu}{\mathbf z_p}-\mathbf z_p$.
\item[(iii)] \emph{Posterior drive.} For $\beta=1$, the drive and the resulting current of neuron $k$ are
\begin{equation}
\begin{aligned}
I_k^\star&=\frac{\E[u_k(\mathbf z_c)\mid\mathbf z_p]}{\ntip{\boldsymbol\mu}{\mathbf z_p}}-u_k(\mathbf z_p),\\
u_k\big(f^\star(\mathbf z_p)\big)&=\frac{\E[u_k(\mathbf z_c)\mid\mathbf z_p]}{\|\boldsymbol\mu\|_2}.
\end{aligned}
\label{eq:nt-posterior}
\end{equation}
\item[(iv)] \emph{Deterministic reference.}
If $\mathbf z_c$ is determined by $\mathbf z_p$, then with $G,\boldsymbol\nu,\lambda$ from Proposition~\ref{prop:nt-silence},
\begin{equation}
I_k^\star=\frac{\lambda^2u_k(\mathbf z_c)-\ntip{\mathbf w_k}{\boldsymbol\nu}}{G},\quad
u_k\big(f^\star(\mathbf z_p)\big)=u_k(\mathbf z_c).
\label{eq:nt-restore}
\end{equation}
\item[(v)] \emph{First gradient.}
At the zero initialization of SC ($\mathbf r=\mathbf 0$), the functional gradient of $d\big(\ntn(\mathbf z_p+\mathbf r),\mathbf z_c\big)$ with respect to $\mathbf r$ is $-\Pperp_{\mathbf z_p}\mathbf z_c$.
The first descent step therefore drives neuron $k$ in proportion to $\ntip{\Pperp_{\mathbf z_p}\mathbf w_k}{\mathbf z_c}=\big(\lambda^2u_k(\mathbf z_c)-\ntip{\mathbf w_k}{\boldsymbol\nu}\big)/G^2$.
\end{enumerate}
\end{ntprop}

\begin{proof}
(i) By linearity of conditional expectation, $\E[d(f,\mathbf z_c)\mid\mathbf z_p]=1-\ntip{f(\mathbf z_p)}{\boldsymbol\mu}$, which is minimized over unit vectors by $f=\ntn(\boldsymbol\mu)$.
(ii) $\ntip{\mathbf r^\star}{\mathbf z_p}=1-1=0$, and $\mathbf z_p+\mathbf r^\star=\boldsymbol\mu/\ntip{\boldsymbol\mu}{\mathbf z_p}$ is a positive multiple of $\boldsymbol\mu$.
(iii) Take inner products with $\mathbf w_k$ and use $\ntip{\mathbf w_k}{\boldsymbol\mu}=\E[u_k(\mathbf z_c)\mid\mathbf z_p]$.
(iv) Here $\boldsymbol\mu=\mathbf z_c$ and $\ntip{\mathbf z_c}{\mathbf z_p}=1/G$, so $\mathbf r^\star=G\mathbf z_c-\mathbf z_p$.
Substituting Eq.~\eqref{eq:nt-gain} gives $I_k^\star=Gu_k(\mathbf z_c)-(u_k(\mathbf z_c)+\ntip{\mathbf w_k}{\boldsymbol\nu})/G$, and $G-1/G=\lambda^2/G$.
(v) The Jacobian of $\ntn$ at a unit vector $\mathbf z_p$ is $\Pperp_{\mathbf z_p}$.
Moreover, $\Pperp_{\mathbf z_p}\mathbf z_c=\mathbf z_c-\mathbf z_p/G$.
Taking the inner product with $\mathbf w_k$ and substituting Eq.~\eqref{eq:nt-gain} gives the stated form.
\end{proof}

Proposition~\ref{prop:nt-posterior} is the formal core of ``eliciting latent knowledge''.
The reference terms do not ask SC to reproduce an arbitrary target.
They drive each concept neuron toward the current it is \emph{expected to have on the complete input, given the incomplete evidence}.
Eqs.~\eqref{eq:nt-restore} and the first-gradient form make the selection explicit.
Neurons that were active on the complete input receive drive proportional to $\lambda^2u_k(\mathbf z_c)$, which grows with the incompleteness ratio.
Neurons that interference pushed up ($\ntip{\mathbf w_k}{\boldsymbol\nu}>0$) receive inhibitory drive.
Silenced concept neurons are thus restored and spurious ones suppressed.
The normalization by $\|\boldsymbol\mu\|_2\le 1$ in Eq.~\eqref{eq:nt-posterior} compensates for posterior uncertainty.
When the complete content is fully determined by the incomplete evidence, $\|\boldsymbol\mu\|_2=1$ and the complete firing pattern is restored exactly.
Together with Proposition~\ref{prop:nt-evidence}, this shows that the posterior can only point toward concepts whose evidence is present in $\mathbf z_p$.

\paragraph{The Remaining RSG Terms as Constraints on the Drive.}
\emph{Relative-improvement hinge.}
The term $[\tilde D_{m,i}-\rho D_{m,i}]_+$ in Eq.~(3) vanishes only if $\ntip{g(\mathbf z_p)}{\mathbf z_c}\ge 1-\rho\,(1-\ntip{\mathbf z_p}{\mathbf z_c})$.
This sets a floor on the reference-aligned drive, so that at least a fraction $1-\rho$ of the reference discrepancy is removed ($\rho=0.75$).
\emph{Complete-input identity.}
On a complete input, $\|g(\mathbf z_c)-\mathbf z_c\|_2^2=2\big(1-(1+\eta_c^2)^{-1/2}\big)$, a strictly increasing function of the drive magnitude $\eta_c=\|r^{\perp}(\mathbf z_c)\|_2$ on the complete input.
The identity term in Eq.~(5), weighted by $\lambda_{\mathrm{id}}=0.1$ (Section~\ref{sec:app_implementation}), is therefore a soft penalty on drive at complete inputs: it discourages large complete-input displacements but does not eliminate drive, and complete-query retrieval still changes after calibration, most strongly for text queries (Table~2 of the main paper).
\emph{Cross-modal anchor.}
Write $\bar{\mathbf t}_{c,i}=\ntn\big(\tfrac15\sum_{l=1}^{5}\mathbf t_{c,il}\big)$ for the normalized caption mean, so that $\mathbf a_i=\ntn(\mathbf v_{c,i}+\bar{\mathbf t}_{c,i})$.
By linearity, the anchor current is
$u_k(\mathbf a_i)=\big(u_k(\mathbf v_{c,i})+\tfrac{1}{5}\sum_{l=1}^{5}u_k(\mathbf t_{c,il})/\|\tfrac15\sum_{l}\mathbf t_{c,il}\|_2\big)/\|\mathbf v_{c,i}+\bar{\mathbf t}_{c,i}\|_2$.
Within the caption term, a concept named in only one caption contributes one fifth of its caption current, so the anchor target favors concepts supported consistently across the image and its captions.
\emph{Anchor-relation KL.}
The distributions $p_{ij}$ in Eq.~(4) depend only on inner products among anchors, that is, on the similarity structure of the population code.
The KL term preserves this structure, consistent with the neighborhood preservation of full SPARC in Fig.~6 (15-neighbor retention of 10.2\% for frozen CLIP and 56.7\% for SPARC at 75\% context masking, evaluated on the selected 300-query subset against the fixed 5,000-image gallery; Section~\ref{sec:app_neighborhood_protocol}).

\begin{ntprop}[Contrastive retrieval selects discriminative neurons]
\label{prop:nt-infonce}
In $\mathcal L_{\mathrm{ret}}$, let $\tilde{\mathbf q}=g(\mathbf z_p)$ be a calibrated query and let the keys $\mathbf k_j$ be the calibrated partial features of the other modality in the minibatch, with positive key $\mathbf k_+$, logits $\ntip{\tilde{\mathbf q}}{\mathbf k_j}/\tau$ and softmax weights $p_j$.
Treating the keys as fixed for the query-side term, the gradient of the query-side InfoNCE term is
$\nabla_{\tilde{\mathbf q}}\mathcal L=-\tau^{-1}\bar{\boldsymbol\delta}$, with $\bar{\boldsymbol\delta}=\mathbf k_+-\sum_jp_j\mathbf k_j$.
At the zero initialization of SC, the first descent step drives neuron $k$ in proportion to $\ntip{\Pperp_{\mathbf z_p}\mathbf w_k}{\bar{\boldsymbol\delta}}$.
If $\mathbf w_k=\boldsymbol\psi_k$ (tied encoder and decoder), this is the soft discriminative weight $\bar\pi_k=\ntip{\boldsymbol\psi_k}{\bar{\boldsymbol\delta}}$ minus the saturation term $u_k(\mathbf z_p)\ntip{\mathbf z_p}{\bar{\boldsymbol\delta}}$.
\end{ntprop}

\begin{proof}
The gradient of $-\log\operatorname{softmax}$ is standard.
The chain rule through $\ntn(\mathbf z_p+\mathbf r)$ at $\mathbf r=\mathbf 0$ contributes $\Pperp_{\mathbf z_p}$.
Then $\ntip{\Pperp_{\mathbf z_p}\mathbf w_k}{\bar{\boldsymbol\delta}}=\ntip{\mathbf w_k}{\bar{\boldsymbol\delta}}-u_k(\mathbf z_p)\ntip{\mathbf z_p}{\bar{\boldsymbol\delta}}$.
\end{proof}

Because the keys are themselves calibrated partial features, $\bar{\boldsymbol\delta}$ is a minibatch estimate of a match direction rather than the fixed-gallery direction $\boldsymbol\delta_{pj}$ of Section~\ref{sec:nt-notation}; $\mathcal L_{\mathrm{MRC}}$, computed against complete frozen galleries, supplies the fixed-gallery direction (Proposition~\ref{prop:nt-mrc}).

\paragraph{Reading Table~4.}
The two selection rules are complementary.
Reference consistency answers \emph{which neurons were silenced} (Proposition~\ref{prop:nt-posterior}).
InfoNCE answers \emph{which neurons separate the candidates} (Proposition~\ref{prop:nt-infonce}).
\begin{itemize}
\item A1 trains SC with $\mathcal L_{\mathrm{ret}}$ only, so its drive follows batch discrimination among incomplete pairs, without reference to the complete firing pattern.
Adding RSG (A2) supplies the posterior selection and improves all four settings, for example from 40.24\% to 44.33\% mR@1 on Flickr30K with ViT-B/32.
\item FD regresses toward complete references.
Insofar as it does so, Proposition~\ref{prop:nt-posterior}(i) applies to it as well, which is consistent with FD exceeding A1 in all four settings even without SC.
Supervision by complete references, shared by FD, Joint FD, and SPARC, therefore accounts for much of the gain over A1.
\item Joint FD and SPARC share the complete references, SC, and MRC, and differ only in the hinge, anchor-relation and identity constraints described above.
Their gap of 0.40--0.57 points in mR@1 is modest; because supervision is matched, it reflects how these constraints shape the drive rather than additional supervision.
\item The controlled component ablations in Table~\ref{tab:app_rsg_ablation} isolate reference consistency, anchor-relation KL, and complete-input identity. Removing reference consistency gives the largest mean loss on incomplete queries, whereas removing identity most strongly affects complete-query I$\to$T retrieval.
\end{itemize}

\subsection{From Reactivation to Rank Recovery}
\label{sec:nt-recovery}

\begin{ntprop}[Recovery condition and the normalization tax]
\label{prop:nt-recovery}
Let $\mathbf z=\mathbf z_p$ and use Lemma~\ref{lem:nt-step}.
\begin{enumerate}
\item[(i)] \emph{Margins.} For every pair, $m_{pj}(\mathbf z^{\beta})=\big(m_{pj}(\mathbf z)+\kappa\ntip{\mathbf r}{\boldsymbol\delta_{pj}}\big)/N$.
\item[(ii)] \emph{Rank neutrality of normalization.}
The query is correct at rank one after calibration if and only if there is $p\in\mathcal P$ such that, for all $j\notin\mathcal P$,
\begin{equation}
\kappa\ntip{\mathbf r}{\boldsymbol\delta_{pj}}>-m_{pj}(\mathbf z).
\label{eq:nt-recover}
\end{equation}
\item[(iii)] \emph{Neuron attribution.}
For any pair, $m_{pj}(\mathbf z^{\beta})-m_{pj}(\mathbf z)=\sum_k\Delta c_k\,\pi_k+\ntip{\Delta\boldsymbol\varepsilon}{\boldsymbol\delta_{pj}}$, where $\Delta c_k=c_k(\mathbf z^{\beta})-c_k(\mathbf z)$.
A previously incorrect pair is repaired if and only if this sum exceeds $-m_{pj}(\mathbf z)$.
\item[(iv)] \emph{Normalization tax.}
Let $m(\mathbf z)>0$ with best positive $p_0$.
If, for all $j\notin\mathcal P$,
\begin{equation}
\kappa\ntip{\mathbf r}{\boldsymbol\delta_{p_0j}}\ge(N-1)\,m(\mathbf z),
\label{eq:nt-tax}
\end{equation}
then $m(\mathbf z^{\beta})\ge m(\mathbf z)$.
For a single pair, the condition is also necessary.
The tax satisfies $(N-1)m(\mathbf z)\le\tfrac12\kappa^2\eta^2m(\mathbf z)$.
\end{enumerate}
\end{ntprop}

\begin{proof}
(i) Take the inner product of Eq.~\eqref{eq:nt-step} with $\boldsymbol\delta_{pj}$.
(ii) Since $N>0$, signs are unchanged by the division.
(iii) Apply Lemma~\ref{lem:nt-concept} with $\mathbf z'=\mathbf z^{\beta}$.
(iv) For all $j$, $m_{p_0j}(\mathbf z)\ge m(\mathbf z)$.
Hence $m(\mathbf z^{\beta})\ge\min_jm_{p_0j}(\mathbf z^{\beta})\ge\big(m(\mathbf z)+(N-1)m(\mathbf z)\big)/N=m(\mathbf z)$.
For a single pair, $m_{pj}(\mathbf z^{\beta})\ge m_{pj}(\mathbf z)$ is equivalent to Eq.~\eqref{eq:nt-tax} by (i).
Finally, $\sqrt{1+x}-1\le x/2$.
\end{proof}

Proposition~\ref{prop:nt-recovery} links reactivation to retrieval.
By part~(ii), rank recovery depends only on the drive projected onto the match directions, and the normalization $N$ never flips a decision.
By part~(iii), the projected drive decomposes exactly over concept neurons.
A recovery therefore requires that reactivated or strengthened neurons ($\Delta c_k>0$) be discriminative ($\pi_k>0$) by enough to overcome the negative margin.
Reactivating shared neurons ($\pi_k\approx 0$) increases $\eta$ without changing any match current.
By part~(iv), such drive is not neutral for queries that were already correct: it pays the tax $(N-1)m(\mathbf z)$ without any return.
MRC penalizes margin reductions, and this tax is one of their two sources; the other is drive with $\ntip{\mathbf r}{\boldsymbol\delta_{pj}}<0$.

\paragraph{Reading Fig.~5 (Left).}
Section~\ref{sec:app_figure5_protocol} describes the frozen-error cohort, matching strata, calipers, and balance records.
For a fixed readout pair $(\mathbf w_a,\mathbf w_b)$, the category-margin gain is
\begin{equation}
\begin{aligned}
&\ntip{\mathbf w_a-\mathbf w_b}{\mathbf z^{\beta}-\mathbf z}\\
&=\frac{\kappa\ntip{\mathbf w_a-\mathbf w_b}{\mathbf r}-(N-1)\ntip{\mathbf w_a-\mathbf w_b}{\mathbf z}}{N},
\end{aligned}
\label{eq:nt-catgain}
\end{equation}
which is the excitation of the semantic readout neurons minus their normalization tax.
Matching the recovered and still-incorrect groups on frozen retrieval and category margins equalizes $\ntip{\mathbf w_a-\mathbf w_b}{\mathbf z}$ across groups, up to the matching tolerance.
The remaining difference of 0.86--1.82 logits therefore reflects the drive term and the normalization $N$, that is, stronger excitation of semantic readout neurons in recovered queries.
This association becomes a causal statement in Sec.~\ref{sec:nt-lesion}.

\begin{ntprop}[Mixing attenuates all drive; MRC inhibits misdirected drive]
\label{prop:nt-mrc}
Fix $\mathbf z$ and $\mathbf r$ and vary $\kappa\ge 0$.
\begin{enumerate}
\item[(i)] \emph{Monotone degradation.}
Let $p_0$ be the best positive of an initially correct query.
The set of $\kappa$ for which $p_0$ remains ranked above all negatives is an interval $[0,\kappa_{\max})$, possibly with $\kappa_{\max}=\infty$.
For a single positive (T$\to$I), correct-to-incorrect degradation at rank one is therefore non-decreasing in $\kappa$, and hence in $\beta$.
\item[(ii)] \emph{Repair threshold.}
For a single positive and an initially incorrect query, repair requires $\kappa>\max_{j:\,m_{pj}(\mathbf z)\le 0}\big(-m_{pj}(\mathbf z)/\ntip{\mathbf r}{\boldsymbol\delta_{pj}}\big)$, with every such $\ntip{\mathbf r}{\boldsymbol\delta_{pj}}>0$.
A uniformly smaller $\kappa$ therefore cannot repair queries whose threshold exceeds it, that is, queries whose negative margins are large relative to their discriminative drive.
\item[(iii)] \emph{Rank-constraint gradient.}
For $\beta=1$ ($\kappa=1$), let $(p,j)$ be the best positive and hardest negative, assumed unique so that $m=m_{pj}$ locally, and let the MRC hinge be active.
Since $\mathbf r=\Pperp_{\mathbf z}\mathbf h$ lies in $T_{\mathbf z}$, take the gradient within $T_{\mathbf z}$.
Up to the positive averaging constant in Eq.~(8),
\begin{equation}
-\nabla_{\mathbf r}\frac{[m_0-m_{pj}(\mathbf z^{1})]_+}{\tau}
=\frac{1}{\tau N}\Big(\Pperp_{\mathbf z}\boldsymbol\delta_{pj}-\frac{m_{pj}(\mathbf z^{1})}{N}\,\mathbf r\Big).
\label{eq:nt-mrcgrad}
\end{equation}
The descent step adds drive along the tangent part of $\boldsymbol\delta_{pj}$, raising the currents of neurons whose tangent weights align with the match direction, that is, discriminative neurons in the sense of Lemma~\ref{lem:nt-concept}.
When $m_{pj}(\mathbf z^{1})>0$, the second term shrinks $\mathbf r$ in proportion to the current margin, which lowers the tax.
It acts only on queries whose margin was positive and decreased.
\item[(iv)] \emph{Stability bound.}
For any unit queries $\mathbf q,\mathbf q'$ and the same gallery, $|m(\mathbf q')-m(\mathbf q)|\le 2\|\mathbf q'-\mathbf q\|_2$.
Since $\|\mathbf z^{\beta}-\mathbf z\|_2=\sqrt{2-2/N}$,
\[
m(\mathbf z^{\beta})\ge m(\mathbf z)-2\sqrt{2-2/N},
\]
so an initially correct query with $m(\mathbf z)>2\sqrt{2-2/N}$ remains correct after calibration.
In addition, for every query with $m_0>0$, the per-query MRC penalty $\ell$ of Eq.~\eqref{eq:app_keep} gives $m_1\ge m_0-\tau\ell$; a zero penalty preserves the margin, and a penalty below $m_0/\tau$ preserves a positive margin, on the gallery used to compute it.
\end{enumerate}
\end{ntprop}

\begin{proof}
(i) $p_0$ outranks negative $j$ if and only if $m_{p_0j}(\mathbf z)+\kappa\ntip{\mathbf r}{\boldsymbol\delta_{p_0j}}>0$, by Proposition~\ref{prop:nt-recovery}(i)--(ii).
Each condition holds at $\kappa=0$ and defines a half-line containing $0$.
Their intersection is an interval containing $0$.
With one positive, $p_0$ is the only positive, and monotonicity in $\beta$ follows from Lemma~\ref{lem:nt-step}.
(ii) Each violated condition $m_{pj}+\kappa\ntip{\mathbf r}{\boldsymbol\delta_{pj}}>0$ with $m_{pj}\le 0$ requires $\ntip{\mathbf r}{\boldsymbol\delta_{pj}}>0$ and $\kappa>-m_{pj}/\ntip{\mathbf r}{\boldsymbol\delta_{pj}}$.
(iii) With $N=\sqrt{1+\|\mathbf r\|_2^2}$, the ambient gradient is $\nabla_{\mathbf r}m_{pj}(\mathbf z^{1})=\boldsymbol\delta_{pj}/N-(m_{pj}(\mathbf z)+\ntip{\mathbf r}{\boldsymbol\delta_{pj}})\mathbf r/N^3$, and the second numerator equals $N\,m_{pj}(\mathbf z^{1})$.
Projecting onto $T_{\mathbf z}$ with $\Pperp_{\mathbf z}$ and using $\Pperp_{\mathbf z}\mathbf r=\mathbf r$ gives Eq.~\eqref{eq:nt-mrcgrad}.
(iv) For any unit $\mathbf y$, $|\ntip{\mathbf q'-\mathbf q}{\mathbf y}|\le\|\mathbf q'-\mathbf q\|_2$, and a maximum over a fixed set of $1$-Lipschitz functions is $1$-Lipschitz.
Applying this to the positive and the negative maxima in Eq.~(7) gives the bound.
By Lemma~\ref{lem:nt-step}, $\ntip{\mathbf z^{\beta}}{\mathbf z}=1/N$, so $\|\mathbf z^{\beta}-\mathbf z\|_2^2=2-2/N$.
The last claim restates $\ell=\tau^{-1}[m_0-m_1]_+$.
\end{proof}

For $\beta<1$, $\kappa$ and $N$ depend on $\mathbf r$ only through $\eta$, so the tangent gradient is $(\kappa/N)\Pperp_{\mathbf z}\boldsymbol\delta_{pj}+c\,\mathbf r$ for a scalar $c$: as in part~(iii), the drive is added along the tangent part of $\boldsymbol\delta_{pj}$, now with weight $\kappa/N$.
With multiple positives (I$\to$T), part~(i) applies to the loss of the original best match.
Part~(iv) is per query and per gallery.
Because $\mathcal L_{\mathrm{MRC}}$ averages over queries and is computed on training minibatch galleries, it does not guarantee that every held-out query keeps its ranking on the full test gallery; degradation therefore remains an empirical quantity, which Fig.~7 and Table~\ref{tab:app_geometry} measure.

\paragraph{Reading Fig.~7 and Table~4 (A3).}
Proposition~\ref{prop:nt-mrc} separates two ways of protecting matches.
Feature mixing shrinks every drive by the same factor $\kappa\le\beta$.
For a fixed drive and a single positive, this cannot increase degradation (part~i), and it can give up repairs that require a larger step (part~ii).
The rank constraint instead acts on at-risk queries alone and redirects their drive toward discriminative neurons (part~iii).
Table~\ref{tab:app_geometry} is consistent with this distinction.
Mixed lowers Deg. relative to Direct, from 7.00\% to 5.80\% on Flickr30K and from 8.00\% to 6.50\% on COCO, with Rep. unchanged; the repair cost allowed by part~(ii) is therefore negligible at $\beta=0.6$.
Ranking Regularized and SPARC lower Deg. further, to 5.30\% and 5.00\% on Flickr30K and to 5.90\% and 5.50\% on COCO, while Rep. stays at or slightly above the Direct level.
Because the variants are trained separately, the drives differ across variants, so this reading is a consistency check rather than a derivation.
The same mechanism explains why adding MRC to SC (A3 vs.\ A1) helps in every setting, for example from 26.81\% to 33.84\% mR@1 on COCO with ViT-L/14.
A1's drive is not rank-aware, whereas MRC removes its misdirected and tax-only components.

\subsection{Lesioning the Drive: Semantic Blocking}
\label{sec:nt-lesion}

\begin{ntprop}[Blocking is a lesion of concept-neuron excitation]
\label{prop:nt-lesion}
Let $\mathcal U$ be a subspace of the tangent space $T_{\mathbf z}=\mathbf z^{\perp}$ with projection $\mathbf P_{\mathcal U}$.
Blocking removes the $\mathcal U$-component of the effective update $\kappa\mathbf r$, which gives the blocked query $\mathbf z^{\beta}_{\mathcal U}=\ntn(\mathbf z+\kappa\mathbf r_{\mathcal U})$ with $\mathbf r_{\mathcal U}=\mathbf r-\mathbf P_{\mathcal U}\mathbf r$.
\begin{enumerate}
\item[(i)] \emph{Exact lesion.}
If $\Pperp_{\mathbf z}\mathbf w_k\in\mathcal U$ for all $k$ in a set $\mathcal K$, the blocked drive is $I_k(\mathbf r_{\mathcal U})=0$ for every $k\in\mathcal K$.
These neurons receive only the divisive factor $1/N_{\mathcal U}$, where $N_{\mathcal U}=\sqrt{1+\kappa^2\|\mathbf r_{\mathcal U}\|_2^2}$.
\item[(ii)] \emph{Lost rescues.}
A recovered single-positive query loses its rescue if and only if $\kappa\ntip{\mathbf r_{\mathcal U}}{\boldsymbol\delta_{pj}}\le-m_{pj}(\mathbf z)$ for some $j$.
\item[(iii)] \emph{Random controls.}
For a Haar-random subspace $\mathcal V\subset T_{\mathbf z}$ with $\dim\mathcal V=\dim\mathcal U=r_{\mathcal U}$, $\E[\mathbf P_{\mathcal V}]=\frac{r_{\mathcal U}}{d-1}\Pperp_{\mathbf z}$.
The expected removed match drive is therefore $\frac{r_{\mathcal U}}{d-1}\kappa\ntip{\mathbf r}{\boldsymbol\delta_{pj}}$.
A sign-symmetric random tangent perturbation of matched norm has zero expected effect on the match drive.
\end{enumerate}
\end{ntprop}

\begin{proof}
(i) $I_k(\mathbf r_{\mathcal U})=\kappa\ntip{\mathbf w_k}{\mathbf r_{\mathcal U}}=\kappa\ntip{\Pperp_{\mathbf z}\mathbf w_k}{\mathbf r_{\mathcal U}}$ since $\mathbf r_{\mathcal U}\in T_{\mathbf z}$.
This is zero because $\mathbf r_{\mathcal U}\perp\mathcal U$.
The divisive factor follows from Proposition~\ref{prop:nt-drive}(i).
(ii) Apply Proposition~\ref{prop:nt-recovery}(ii) with $\mathbf r_{\mathcal U}$ in place of $\mathbf r$.
(iii) Invariance under rotations of $T_{\mathbf z}$ makes $\E[\mathbf P_{\mathcal V}]$ a multiple of $\Pperp_{\mathbf z}$, and the trace equals $r_{\mathcal U}$.
\end{proof}

\paragraph{Reading Fig.~5 (Right).}
The effective-step implementation and the separate Top-1 and margin endpoints are defined in Section~\ref{sec:app_figure5_protocol}.
The blocking intervention removes the component of the calibration update in the span of the semantic readout directions.
By part~(i), this is an exact lesion: it sets the excitation of the corresponding readout neurons to zero, and they receive only divisive inhibition.
By part~(iii), dimension-matched random spans remove only the proportional share $r_{\mathcal U}/(d-1)$ of the match drive in expectation, which is small when $r_{\mathcal U}\ll d$.
Magnitude-matched random tangent perturbations remove none of it in expectation.
The positive excess loss of recovered matches, 28.67\% and 31.36\% relative to random-span and random-tangent controls on COCO with ViT-B/32, therefore shows that the recoveries are carried disproportionately by excitation of semantic concept neurons.
With the input, the encoder and the gallery held fixed, this is the causal step of the storyline.
The fixed-gallery matching-margin endpoint of the same lesion is positive in all four dataset--backbone settings under both random-tangent and same-rank controls (Section~\ref{sec:app_figure5_protocol}).

\subsection{The Causal Chain and Neuron-Level Predictions}
\label{sec:nt-chain}

The results above form one chain from latent knowledge to retrieval:
\begin{enumerate}
\item \textbf{Knowledge persists in the encoder.}
Incomplete inputs leave much of the target evidence in the encoder's patch representations (Fig.~2A), from which the readout token is formed (Lemma~\ref{lem:nt-source}), and only this evidence can later be used (Proposition~\ref{prop:nt-evidence}; Fig.~2C4, Section~\ref{sec:app_finegrained_cues}).
\item \textbf{Readout neurons are silenced.}
A shared gain $G$ and neuron-specific interference push weakly suprathreshold concept neurons below threshold (Proposition~\ref{prop:nt-silence}; Fig.~2B).
\item \textbf{Silenced discriminative neurons cause mismatches.}
Category neurons and match neurons read nearly orthogonal directions, so queries stay readable while discriminative interference flips matches (Proposition~\ref{prop:nt-mismatch}; Fig.~2C1--C3).
\item \textbf{SC delivers a bounded tangent drive.}
The drive is neuron-specific, divisively normalized, saturation-aware and energy-bounded (Lemma~\ref{lem:nt-step}, Proposition~\ref{prop:nt-drive}; Fig.~7).
\item \textbf{RSG and InfoNCE select the neurons.}
The drive is directed toward the posterior-expected complete firing pattern and toward discriminative neurons, with a soft penalty on drive at complete inputs (Propositions~\ref{prop:nt-posterior}--\ref{prop:nt-infonce}; Table~4, Fig.~6).
\item \textbf{Reactivation recovers ranks, and MRC protects them.}
Recovery requires discriminative reactivation, and MRC inhibits misdirected and tax-only drive (Propositions~\ref{prop:nt-recovery}--\ref{prop:nt-mrc}; Fig.~5 left, Fig.~7, Table~4).
\item \textbf{Lesioning the drive undoes recovery.}
Blocking the excitation of semantic neurons removes recoveries beyond matched random controls (Proposition~\ref{prop:nt-lesion}; Fig.~5 right, Section~\ref{sec:app_figure5_protocol}).
\end{enumerate}

\paragraph{Predictions for Unit-Level Measurements.}
The theory makes direct predictions for SAE concept neurons, each of which can be checked without retraining.
\begin{itemize}
\item \emph{Envelope.}
Every unit's drive $I_k=N u_k(\mathbf z^{\beta})-u_k(\mathbf z_p)$ satisfies Eq.~\eqref{eq:nt-envelope}.
Reactivated units are expected to lie near the envelope at small $|u_k(\mathbf z_p)|$.
\item \emph{Selectivity.}
Across units, $I_k$ is expected to correlate positively with the complete-input current $u_k(\mathbf z_c)$ and negatively with the interference $\ntip{\mathbf w_k}{\boldsymbol\nu}$ (Eq.~\eqref{eq:nt-restore}).
\item \emph{Attribution.}
In recovered queries, the exact sum in Proposition~\ref{prop:nt-recovery}(iii) is expected to be dominated by reactivated units with $\pi_k>0$.
In degraded queries, it is expected to be dominated by excited units with $\pi_k<0$, with fewer such units under MRC.
\item \emph{Unit lesions.}
Choosing $\mathcal U=\operatorname{span}\{\Pperp_{\mathbf z}\mathbf w_k:k\in\mathcal R^{+}\}$, where $\mathcal R^{+}$ is the set of reactivated discriminative units, is expected to remove recoveries, whereas random sets matched in size, firing rate and drive magnitude are not (Proposition~\ref{prop:nt-lesion}).
\item \emph{Evidence dependence.}
On cue-erased pairs, the discriminative drive $\sum_k\Delta c_k\pi_k$, averaged over the two cue values, is expected to be centered near zero (Proposition~\ref{prop:nt-evidence}).
\end{itemize}

\subsection{Scope of the Analysis}
\label{sec:nt-scope}

All identities above are exact.
Lemma~\ref{lem:nt-source} evaluates the normalization statistics on the same input, and Lemma~\ref{lem:nt-concept} keeps the SAE reconstruction error as an explicit term.
For TopK SAEs, thresholds are the operating thresholds at the given input.
Proposition~\ref{prop:nt-posterior} characterizes the population minimizer of the reference term over all unit-valued maps, and its first-gradient form describes training at the zero initialization; the trained heads approximate these targets with finite capacity and data.
Propositions~\ref{prop:nt-drive}--\ref{prop:nt-lesion} hold for any learned residual and are evaluated on the trained heads.
Comparisons across separately trained variants (Fig.~7, Table~4) are read as consistency checks, because the learned drives differ across variants.
The analysis concerns the query side.
Gallery features are fixed, and the encoders, and hence all source-neuron activations, are unchanged by SPARC.

\FloatBarrier
\section{Qualitative Retrieval Examples}\label{sec:app_qualitative}

\begin{figure}[H]
  \centering
  \includegraphics[width=0.80\linewidth]{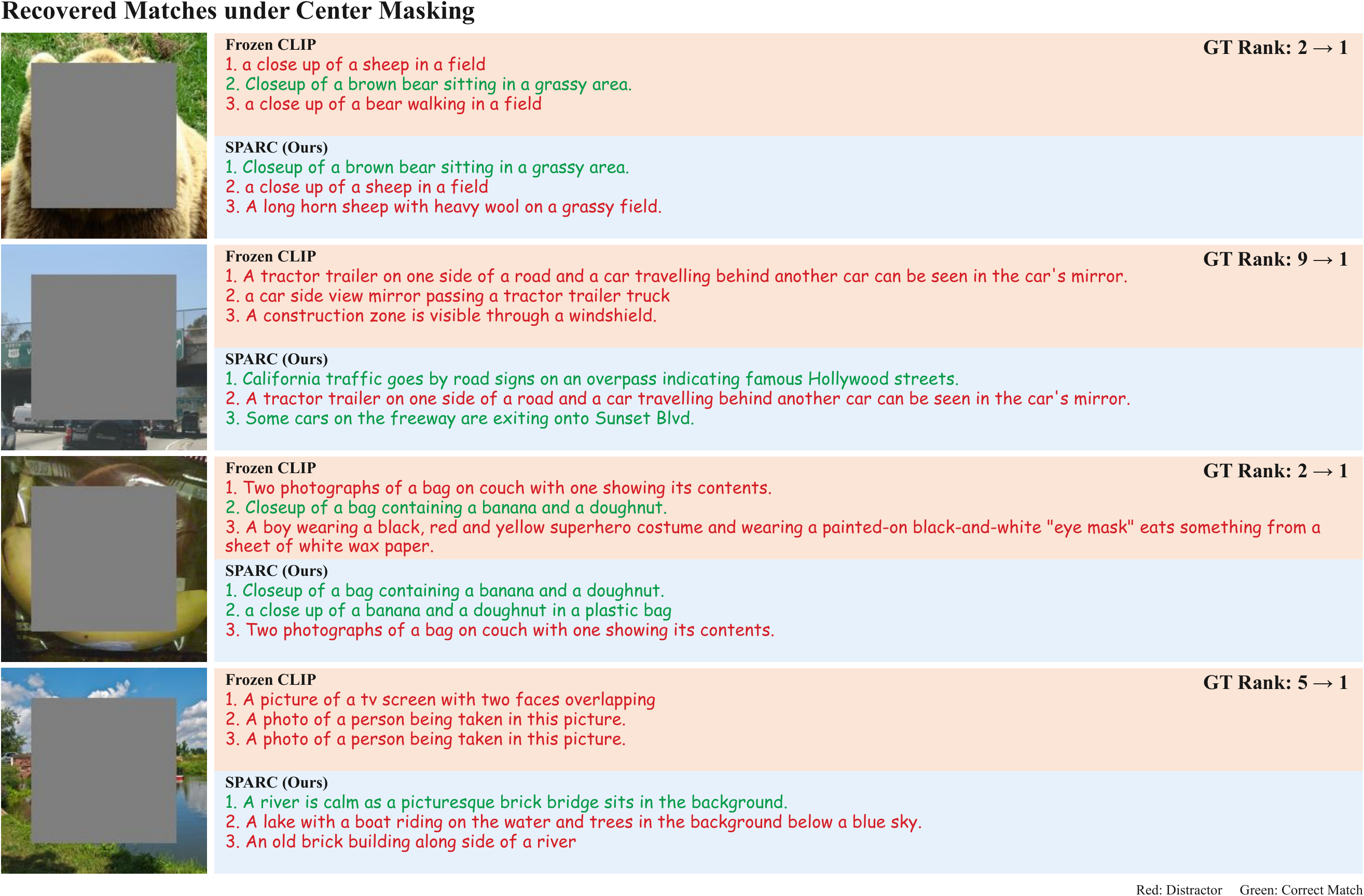}\par
  \includegraphics[width=0.80\linewidth]{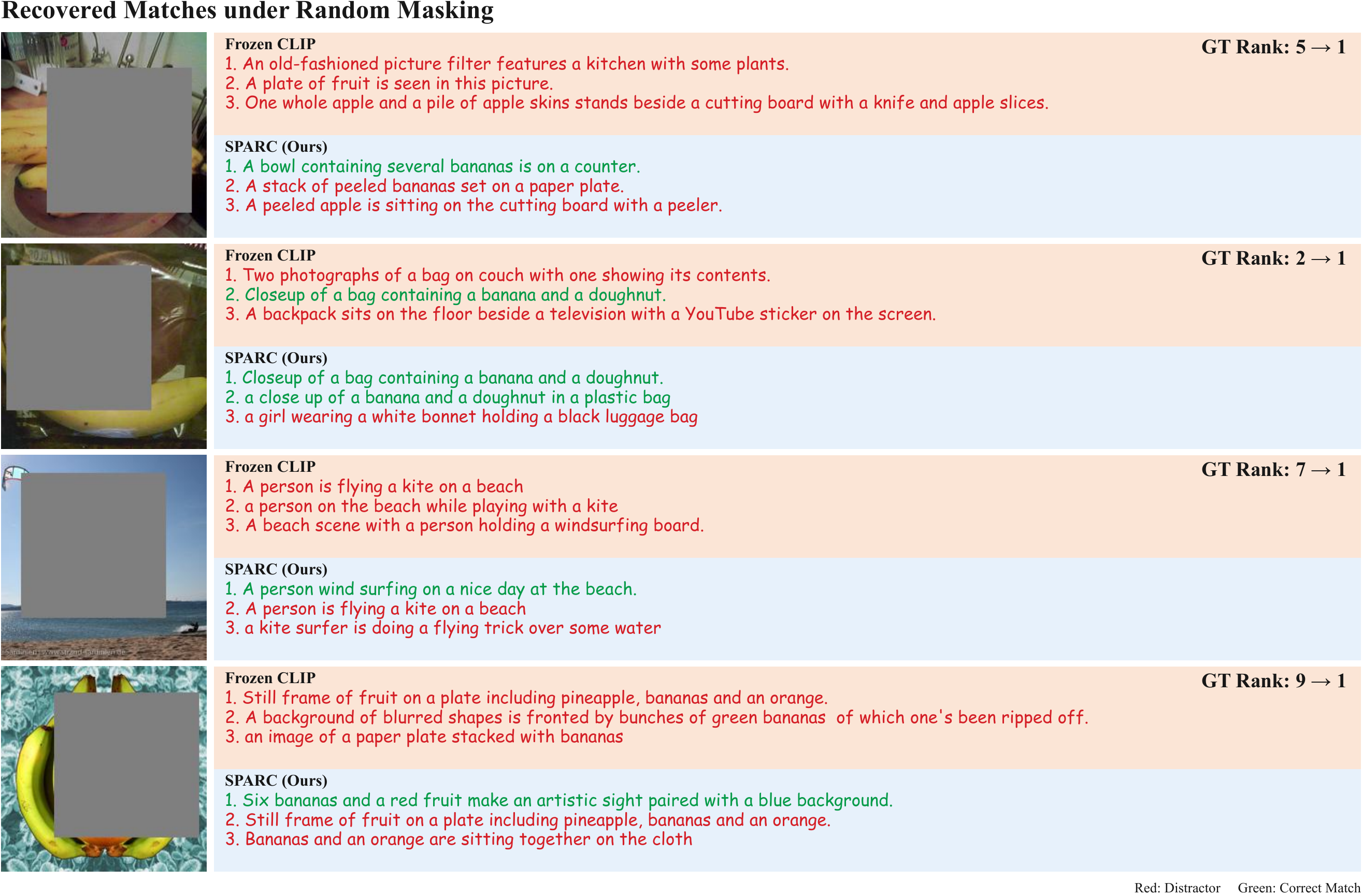}
  \caption{\textbf{Selected Recovered Matches under Image Masking.}
  Upper: four center-masked queries. Lower: four random-position-masked queries.
  Each panel compares the first-ranked captions from Frozen CLIP and SPARC for the same partial image: the former is non-paired, while the latter is benchmark-paired, marked by GT. Headers give the best true-caption rank before and after calibration. The seed-42 corruption patterns are reused. These are selected recoveries; a non-paired caption may still be semantically compatible with the image.}
  \label{fig:app_recovered_images}
  \label{fig:app_recovered_center}
  \label{fig:app_recovered_random}
\end{figure}

\begin{figure}[H]
  \centering
  \includegraphics[width=0.95\linewidth]{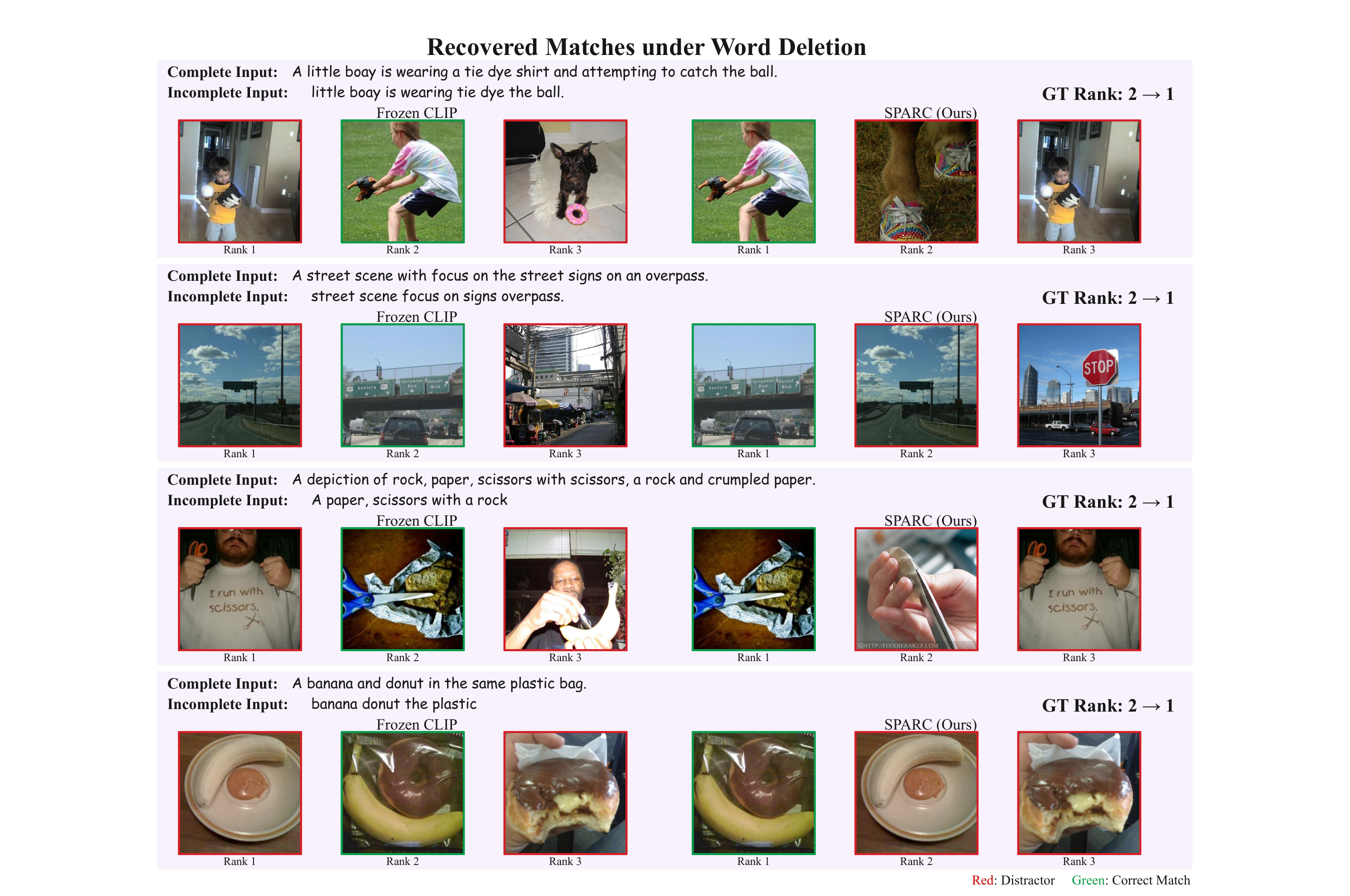}
  \caption{\textbf{Selected Recovered Matches under Word Deletion.}
  Four validation captions use their original seed-42 deletion patterns. Each panel compares the first-ranked images for the identical remaining words: Frozen CLIP retrieves a non-paired image, while SPARC retrieves the benchmark-paired image, marked by GT. Headers report the true-image rank before and after calibration. Only selected successful recoveries are displayed.}
  \label{fig:app_recovered_word}
\end{figure}

We show twelve selected COCO validation recoveries (Figures~\ref{fig:app_recovered_images} and~\ref{fig:app_recovered_word}): four each for center masking, random-position masking, and word deletion, all using CLIP ViT-B/32 at 50\% missingness. Frozen CLIP and SPARC receive identical partial queries, use the same validation-selected checkpoint throughout, and share complete galleries encoded by the original frozen CLIP. The image-query gallery contains 25,000 captions, and the text-query gallery contains 5,000 images. These validation examples show selected successful recoveries; the test-set retrieval summary is reported in Table~\ref{tab:app_multiseed}.

\FloatBarrier
\section{Further Discussion}
\label{sec:app_discussion}

\appquestion{Q1. Why can category information remain decodable while cross-modal retrieval fails, and how do we analyze it?}
Category recognition and instance-level matching require different distinctions. Recognizing a dog does not necessarily distinguish a dog running in water from one resting on grass. Moreover, probe decodability does not establish that the prediction uses the available cues~\cite{ravichander2021probing,elazar2021amnesic}; native cosine matching must be assessed separately. To distinguish information removal from underutilization of surviving semantics, our controlled analysis protects annotated target cues while masking their surrounding context, reducing confounding from directly deleting the target. Retained category readout alongside changes in semantic responses and matching relations motivates studying how surviving information is expressed and used, whereas natural image masking and word deletion separately evaluate benchmark retrieval degradation. Preserving a category cue does not establish that all instance-specific attributes and relations remain available.

We therefore combine measurements that address complementary questions instead of reporting recall alone. Category probes test decodability; SAEs describe response changes in category-associated components; matching margins quantify separation between genuine matches and distractors; and directional interventions test the functional contributions of specified semantic directions. Together with gallery-neighborhood relations, these measurements connect information availability, response strength, and decision-level use, showing how retained semantics can coexist with incorrect matching while avoiding a mechanism claim based solely on a change in retrieval recall (Section~\ref{sec:app_semantics}, and Proposition~\ref{prop:nt-mismatch} in Section~\ref{app:neuron-theory}).

\appquestion{Q2. How does SPARC differ from a conventional feature adapter, and why retain useful rankings?}
SPARC uses a residual MLP comparable to feature adapters such as CLIP-Adapter~\cite{gao2021adapter}; it differs in three specific respects. First, the update is tangent to the unit sphere, which removes the radial output channel; with all other settings matched, this lowers the correct-to-incorrect degradation rate relative to a normalized residual (Section~\ref{sec:app_geometry}). Second, complete image--text references supervise the map during training only, while inference uses the incomplete query alone. Third, a margin-retention constraint computed against a frozen complete gallery targets queries that are already correct. Some partial queries already support correct matches, and an aggregate retrieval objective does not directly constrain the ranking change of each initially correct query. The retention term therefore softly penalizes decreases in initially positive matching margins on the training-batch gallery, limiting disruption to these useful relations and reducing degradation beyond feature mixing alone (Table~\ref{tab:app_geometry}). Because optimization averages over queries and uses finite minibatches, it does not guarantee that every held-out query retains its original ranking (Proposition~\ref{prop:nt-mrc}). Category readouts, response analyses, and semantic-direction interventions (Section~\ref{sec:app_semantics}) investigate how retained information participates in matching. The contribution combines this problem setting, these design choices, and the accompanying functional evidence.

\appquestion{Q3. Why use separate visual and textual calibration maps, and what does deployment require?}
Image occlusion and missing words affect different cues, motivating modality-specific adjustments. Normalized visual and textual features occupy separate unit spheres~\cite{wang2020understanding}. We write their joint space as a product of spheres only as notation: the two maps act independently at inference and are coupled solely through the training objectives, namely InfoNCE and the cross-modal anchors. Tangent projection and normalization align the calibration parameterization with cosine retrieval, and their empirical role is assessed through the geometry comparison in Section~\ref{sec:app_geometry}. Deployment requires neither a complete query nor an additional paired modality. SPARC receives only the available partial-query feature and applies its modality's calibration map; the complete gallery consists of fixed features from the original frozen encoders. Complete image--text pairs are used only during training.

\appquestion{Q4. Why does eliciting latent knowledge outperform feature distillation?}
Feature distillation (FD) maps incomplete features toward complete targets, whereas SPARC elicits the semantic evidence retained in the incomplete query, using complete features only as training-time semantic references. Concretely, SPARC evaluates each calibrated feature relative to its own incomplete input (the relative-improvement hinge), preserves the semantic structure among samples through cross-modal anchors (the anchor-relation KL), and regularizes the map on complete inputs (the identity term); none of these is specified by distillation. Table~4 of the main paper confirms the advantage of this design. SPARC outperforms FD by 3.14--6.49 mR@1 points, and it outperforms FD even when FD is equipped with SPARC's calibration structure and training budget (Joint FD), in all four dataset--backbone settings. Consistent with eliciting retained knowledge, SPARC's gains are more than six times larger when the discriminating cue survives in the query (Section~\ref{sec:app_finegrained_cues}).

\appquestion{Q5. Does calibration transfer to unseen conditions, and how do its gains on incomplete queries relate to those on complete queries?}
We train the visual map only on 50\% center-masked images, so random-position masking and the 25\% and 75\% image rates are unseen during training. Nevertheless, SPARC outperforms all compared methods in every incomplete-query condition with CLIP ViT-B/32 (Table~1 of the main paper) and outperforms the frozen and fine-tuned counterparts in every condition with the other three backbones (Table~3 of the main paper). The corruption conditions of Flickr30K-C and COCO-C~\cite{qiu2022robustness} are likewise unseen during training: 85 image conditions (17 perturbation types at five severities) and 60 text conditions (16 perturbation types with type-specific severity settings; Section~\ref{sec:app_corruption_protocol}), each applied to the original test queries. With CLIP ViT-B/32, SPARC achieves the highest mean R@1 and RSum on both benchmarks (Table~2 of the main paper). Text training covers all three word-deletion rates, so the word-deletion conditions are not out-of-distribution in rate.

SPARC is trained for incomplete queries and also benefits complete queries (Table~2 of the main paper), while its advantage grows as queries become incomplete. On the image side, the visual map changes complete-query I$\to$T R@1 by 1.10 and 4.48 points on Flickr30K and COCO, but improves masked queries by 10.36--16.92 points, including masking types and rates unseen during training (Table~1 of the main paper). On the text side, the relative gain over frozen CLIP exceeds that on complete queries at every word-deletion rate, for example 25.6\% versus 14.4\% on Flickr30K and 41.1\% versus 27.2\% on COCO at 25\% deletion. In addition, SPARC's margin over the second-best method is larger on the broader text perturbations of Flickr30K-C and COCO-C (3.12 and 2.57 T$\to$I R@1 points) than on complete queries (1.12 and 1.57 points; Table~2 of the main paper).

\paragraph{Limitation.}
SPARC shows that latent multimodal knowledge retained in frozen vision--language encoders can be elicited through lightweight query-side calibration to improve robust image--text retrieval under incomplete queries. However, the current setting considers one-sided incompleteness, where only the query is partially observed and the gallery remains complete. In future work, we plan to extend latent-knowledge elicitation to two-sided incompleteness, where both queries and gallery items are partially observed.

\end{document}